\documentclass[11pt]{article}

\usepackage[final]{acl}

\usepackage{times}
\usepackage{latexsym}

\usepackage{amsmath,amsfonts,bm}

\def\eqref#1{equation~\ref{#1}}
\def\1{\bm{1}}

\DeclareMathAlphabet{\mathsfit}{\encodingdefault}{\sfdefault}{m}{sl}
\SetMathAlphabet{\mathsfit}{bold}{\encodingdefault}{\sfdefault}{bx}{n}

\usepackage{inconsolata}

\usepackage[utf8]{inputenc} % allow utf-8 input
\usepackage[T2A,T1]{fontenc}    % use 8-bit T1 fonts
\usepackage{hyperref}       % hyperlinks
\usepackage{url}            % simple URL typesetting
\usepackage{booktabs}       % professional-quality tables
\usepackage{amsfonts}       % blackboard math symbols
\usepackage{nicefrac}       % compact symbols for 1/2, etc.
\usepackage{microtype}      % microtypography
\usepackage{xcolor}         % colors
\usepackage{amsmath}        % math
\usepackage{multirow}
\usepackage{multicol}
\usepackage{graphicx}
\usepackage{floatflt}
\usepackage{enumitem}
\usepackage{makecell}
\usepackage{wrapfig}
\usepackage{fontawesome}
\usepackage{pifont}
\usepackage[russian,english]{babel}
\usepackage{CJKutf8}
\usepackage{soul}
\definecolor{lightyellow}{rgb}{0.7, 1.0, 0.7}

\sethlcolor{lightyellow}

\newcommand{\cmark}{\ding{51}}%
\newcommand{\xmark}{\ding{55}}%

\newcommand*{\revision}{} % TODO: change to this one for main submission

 \newcommand*{\remove}[1]{\leavevmode\ignorespaces} % TODO: change to this one for main submission

\title{Sparse Feature Coactivation Reveals Causal Semantic Modules\\in Large Language Models}

\author{
Ruixuan Deng\textsuperscript{$1$}\thanks{Indicates equal contribution.}
\hspace{20pt}
Xiaoyang Hu\textsuperscript{$2$}\footnotemark[1]
\hspace{20pt}
Miles Gilberti\textsuperscript{$3$}\footnotemark[1]
\hspace{20pt}
Shane Storks\textsuperscript{$3$}\footnotemark[1]
\hspace{20pt}\\[7.5pt]
\textbf{Aman Taxali}\textsuperscript{$3$}
\hspace{20pt}
\textbf{Mike Angstadt}\textsuperscript{$3$}
\hspace{20pt}
\textbf{Chandra Sripada}\textsuperscript{$3$}
\hspace{20pt}
\textbf{Joyce Chai}\textsuperscript{$3$} \\[7.5pt]
$^1$Georgia Institute of Technology
\hspace{10pt}
$^2$Brown University
\hspace{10pt}
$^3$University of Michigan \\
\texttt{rdeng62@gatech.edu},\hspace{25pt}
\texttt{xiaoyang\_hu@brown.edu},\\
\texttt{\{milgil, sstorks, ataxali, mangstad, sripada, chaijy\}@umich.edu}
}

\begin{document}

\maketitle

\begin{abstract}

We identify semantically coherent, context-consistent network components in large language models (LLMs) using coactivation of sparse autoencoder (SAE) features collected from just a handful of prompts. Focusing on concept-relation prediction tasks, we show that ablating these components for concepts (e.g., countries and words) and relations (e.g., capital city and translation language) changes model outputs in predictable ways, while amplifying these components induces counterfactual responses. Notably, composing relation and concept components yields compound counterfactual outputs. Further analysis reveals that while most concept components emerge from the very first layer, more abstract relation components are concentrated in later layers. Lastly, we show that extracted components more comprehensively capture concepts and relations than individual features while maintaining specificity.
% Overall, our findings suggest a modular organization of knowledge accessed through compositional operations, and advance methods for efficient, targeted LLM manipulation.
{Overall, our findings suggest a modular organization of knowledge and advance methods for efficient, targeted LLM manipulation.\footnote{Code available at \url{https://github.com/shanestorks/SAE-Semantic-Modules}.}} % NH: I think "accessed through compositional operations" is a bit vague, and perhaps we should deemphasize the compositional / composable part given new results
\end{abstract}

\begin{figure*}
    \centering
    \includegraphics[width=0.95\linewidth]{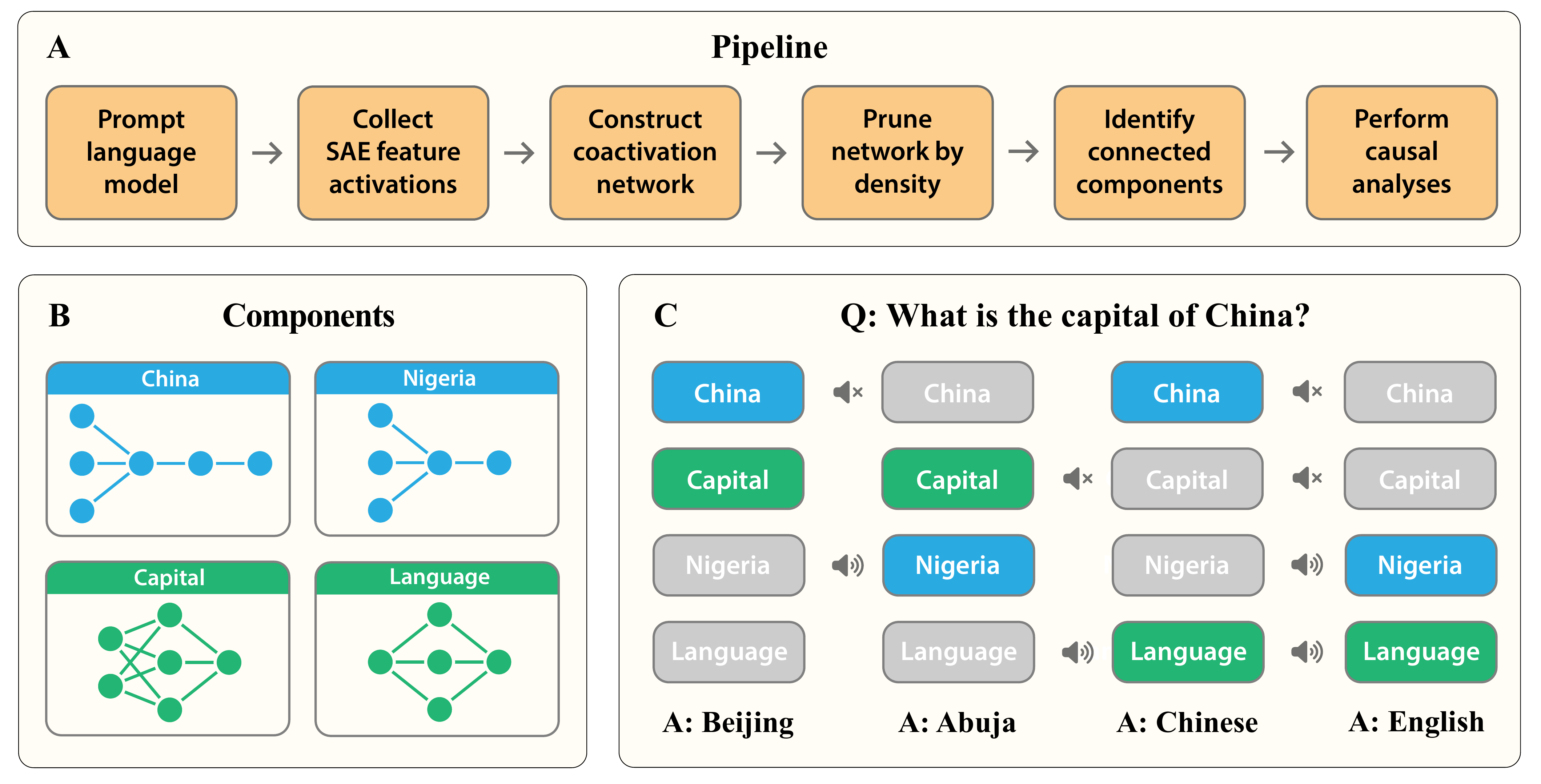}
    \vspace{-5pt}
    \caption{(A) We construct inter-layer feature networks from SAE coactivation patterns, prune high-density features, and extract task-relevant components. (B) Components are often consistent across contexts. (C) Selective ablation and amplification of components steers the model toward counterfactual outputs, overriding the original prompt.}

    % \vspace{-5pt}
    
    \label{fig:overview}
\end{figure*}

\section{Introduction}\label{sec:intro}

Contemporary neural large language models (LLMs) exhibit remarkable proficiency in representing and reasoning over relational knowledge, motivating efforts toward mechanistic interpretation. Surprisingly, despite the immense complexity of LLMs, prior work has modeled this capability with simple linear transformations~\citep{hernandez2024linearity} and vector arithmetic~\citep{merullo2024language}. However, these accounts are limited to layerwise interpretations, lacking fine-grained interpretation of specific features and their interaction.

Sparse autoencoders (SAEs; \citealp{huben2024sparse}) recently emerged as a powerful tool for extracting interpretable, often monosemantic features from the internal representations of LLMs.
SAEs achieve this by learning to reconstruct dense, entangled activations at each neural network layer through sparse codes, where individual features are incentivized to activate rarely and selectively.
Despite this progress, it remains unclear how features from different layers organize and coordinate to produce responses (e.g., how an LLM may access and compose its encoded knowledge to respond ``\textit{Beijing}'' to a prompt for China's capital city).
% For example, when prompted for China’s capital, an LLM activates numerous SAE features spanning many layers—some encoding aspects of China, capital cities, or geography, others seemingly irrelevant—but how these features organize and coordinate across layers to yield ``\textit{Beijing}'' is not well understood. 
%Cross-layer transcoders extend SAEs to approximate layerwise computations and globally interpret information flow in LLMs as computation graphs \citep{dunefsky2024transcoders,ameisen2025circuit}; these graphs consist of hundreds of densely connected nodes, requiring manual feature grouping for interpretability. As such, there remains a need for a global, sparse, feature-level interpretation of LLMs' relational knowledge and reasoning.
{Cross-layer transcoders \citep{dunefsky2024transcoders} extend SAEs by learning to approximate MLP computations and have been used to construct computation graphs that trace information flow across LLM layers \cite{ameisen2025circuit}. However, these graphs can contain hundreds of densely connected nodes, making them difficult to interpret without manual grouping of features, motivating a more automated approach.} % NH: I think we should separate transcoders and circuit tracing.

% I think this discussion could even go in the intro. We can say something like:
% Compared to previous work, we automatically identify feature groups. But why is this valuable?
% It tells us more about the prominence (how many features) and topology (features' layer distribution and connectivity) of different concepts
% More practically, if we want to do concept deletion, deleting one single feature does not work (as shown in section x), but deleting a network of related features does work (as shown in section y).
% what do you think @channel (edited) 

In this work, we employ SAE feature coactivation to construct inter-layer feature networks for individual prompts about various factual and linguistic relational reasoning tasks.
%By filtering out denser features that activate across diverse contexts, we identify semantically coherent connected components within these networks that represent specific concepts and abstract relations across different prompts, and can predictably drive model outputs when manipulated.
{By filtering out features that activate across diverse contexts, we identify semantically coherent connected components representing specific concepts and relations. These components are consistent across prompts, and manipulating them predictably alters model outputs. }% NH: I think better to split into two sentences
As exemplified in Figure~\ref{fig:overview}, a prompt about China's capital activates components related to China and capital cities, eliciting the correct response of ``\textit{Beijing}'' from the model.
However, ablating features in the China component while amplifying those in the Nigeria component effectively steers its response to ``\textit{Abuja}'', whereas ablating the capital component while amplifying the language component steers it to ``\textit{Chinese}.''
Furthermore, simultaneously ablating the China and capital components while amplifying the Nigeria and language components changes the model output to ``\textit{English}.''
%In addition to contributing a sparse, modular structure of directions that causally account for LLM relational reasoning processes, these components readily enable fine-grained interpretation of the topology of stored information.
{These components also readily visualize the topology of stored concepts and relations across layers.} % NH: so I feel like there is no need to introduce this "direction" framing for the first time here and the fact we can manipulate model output using these components were made very clear before this sentence.
% In addition to contributing a sparse, modular structure of directions that causally account for LLM relational reasoning processes, these components readily enable fine-grained interpretation.
% For example, they reveal the prominence and topology of specific concepts and relations; by ablating individual component features, we discover that country and word components tend to emerge from the very first network layer.
% In contrast, more abstract relational components for facts about countries, translation target languages, and lexical transformations (e.g., past tense, capitalization, first letter) are concentrated in later layers.
% This aligns with our understanding of artificial neural networks and biological systems, where higher-level abstractions emerge in later processing stages \cite{dicarlo2012does,lecun2015deep}.
% While the nodes in some relational components also become more causally impactful on model outputs in later layers (e.g., country facts, past tense, and first letter), others exhibit the opposite trend (e.g., translation languages and capitalization).
% Lastly, our analysis demonstrates that our components offer a more comprehensive account of LLM relational reasoning than individual SAE features, and that components for each concept-relation pair are mostly highly specific to the functions we associate them with.
% \jycc{the above paragragh is pretty dense, use itemize to summarize contributions and findings. }
Our experiments yield the following findings:
\begin{itemize}
    \item Selective ablation and amplification of components consistently overrides prompted instructions, steering model outputs to predictable counterfactual answers in up to 98\% of cases when steering based on individual concepts or relations, and up to 90\% of cases when based on compound concept-relation pairs.
    \item Conceptual components representing information about specific entities and words tend to emerge from the very first network layer, while more abstract relational components (e.g., for properties of countries or translation languages) concentrate in later layers.\footnote{This aligns with our understanding of artificial neural networks and biological systems, where higher-level abstractions emerge in later processing stages \cite{dicarlo2012does,lecun2015deep}.} 
    % Further, while the nodes in some relational components also become more causally impactful on model outputs in later layers (e.g., geographic knowledge and first letter of words), others exhibit the opposite trend (e.g., translation languages and capitalization of words).
    % \item While the nodes in some relational components also become more causally impactful on model outputs in later layers (e.g., country facts, past tense, and first letter), others exhibit the opposite trend (e.g., translation languages and capitalization).
    \item Our components offer a more comprehensive causal account of LLM relational reasoning than individual SAE features, and are mostly highly specific to the concepts and relations we associate them with (i.e., ablating them often preserves performance in other contexts).
    
\end{itemize}

Together, these results offer insights into the organization of relational knowledge in LLMs and establish sparse feature coactivation as an effective, lightweight approach for mechanistic interpretability and targeted model manipulation.

\section{Methods}
\label{sec:methods}

We focus our analyses on Gemma 2 2B \citep{gemmateam2024gemma2improvingopen}\footnote{Accessed via \texttt{google/gemma-2-2b} in Hugging Face Hub \citep{wolf-etal-2020-transformers}. LLM activations collected with 1 NVIDIA A100 GPU (40GB VRAM) and 12GB RAM.}, with some results for Gemma 2 9B in Appendix~\ref{apx:9b-results}. As shown in Figure~\ref{fig:overview}, we perform several steps to construct an inter-layer feature network based on coactivation patterns, extract task-relevant components, and perform targeted causal interventions to assess their functional roles.

% The detailed steps are as follows

% \begin{enumerate}[leftmargin=13pt]

% \item \textbf{Activation collection}: 
\paragraph{Activation collection.}

We run each input prompt through the LLM integrated with pre-trained SAEs.\footnote{We use SAE Lens to apply the \texttt{width\_16k/canonical} variant from the \texttt{gemma-scope-2b-pt-res-canonical} release \citep{lieberum2024gemmascopeopensparse,bloom2024saetrainingcodebase}.} Each SAE maps the residual stream activation at layer $\ell$, $x_\ell\in \mathbb{R}^{d_{\text{model}}}$, to a sparse representation $\phi_\ell \in \mathbb{R}^{d_{\text{sae}}}$ where $d_{\text{sae}}=16384$. This produces an activation tensor $\Phi_\ell \in \mathbb{R}^{T \times d_{\text{sae}}}$ for each layer $\ell$, where $T$ is the number of non-BOS tokens in the prompt.

\paragraph{Feature selection.} 

To ensure computational tractability while preserving key information, we select a set $S_\ell$ of top-activated features for each layer. A feature index $i$ is included in $S_\ell$ if it appears in the top $k=5$ activations at \textit{any} token position $t \in \{1, \dots, T\}$:

\vspace{-20pt}

\begin{equation*}
    S_\ell = \bigcup_{t=1}^{T} \{ i \mid \Phi_\ell[t, i] \in \text{top-}k(\Phi_\ell[t, :]) \}
    \label{eq:topk_selection}
\end{equation*}

\vspace{-12pt}
    
\paragraph{Graph construction.}

We construct a directed graph $G = (V, E)$ where each node $(\ell,i)\in V$ corresponds to a selected feature $i \in S_\ell$. Edges $E$ connect nodes in adjacent layers according to the temporal correlation of their activation patterns across tokens in the prompt. Specifically, for feature pairs in adjacent layers $i\in S_\ell$ and $j\in S_{\ell+1}$, we compute the Pearson correlation coefficient $\rho$.
% \vspace{-6pt}
% \begin{equation*}
%     \rho\left(\Phi_\ell[:, i], \Phi_{\ell+1}[:, j]\right) = \frac{\text{cov}(\Phi_\ell[:, i], \Phi_{\ell+1}[:, j])}{\sigma(\Phi_\ell[:, i]) \sigma(\Phi_{\ell+1}[:, j])}
%     \label{eq:correlation}
% \end{equation*}
A directed edge $e = ((\ell, i), (\ell+1, j))$ is added to $E$ if $\rho(\Phi_\ell[:, i], \Phi_{\ell+1}[:, j]) > \tau_{\text{corr}} = 0.9$. 
% Edge weights are assigned as $w(e) = \rho$.

\paragraph{Pruning and component identification.} 
    
Some SAE features activate frequently across unrelated contexts, making them overly generic and hard to interpret. To reduce noise, we prune the graph using activation density scores from Neuronpedia \citep{neuronpedia}.
% \footnote{\url{https://neuronpedia.org}} 
For each node $(\ell, i) \in V$, we retrieve its activation density $d_{\ell,i}$, i.e., the fraction of tokens in a large corpus where its corresponding feature activates. We retain only \textit{sparse features}, i.e., those with $d_{\ell,i} \le \tau_{\text{density}} = 0.01$,\footnote{This follows Neuronpedia’s standard, which classifies features below this density as sparse and interpretable.} creating a pruned graph $G_{\text{sparse}}$. We also remove any isolated nodes from $G_{\text{sparse}}$.
We then use a straightforward BFS-based method from NetworkX\footnote{\url{https://networkx.org/}} to identify weakly connected components within $G_{\text{sparse}}$. 
% Figures~\ref{fig:china_component_capital_currency} shows example components.

\paragraph{Causal validation.} 
To evaluate the functional significance of each component, we perform targeted interventions using \texttt{TransformerLens}\ \citep{nanda2022transformerlens}. Specifically, we  \textbf{ablate} (i.e., set to zero) the activations of SAE features in a given component during the LLM’s forward pass,\footnote{When manipulating SAE features like this, we replace the LLM's internal activations at each layer with the corresponding SAE decoder's output after features are manipulated, then proceed with the forward pass.}, then measure the resulting shift in the probability distribution over next-token predictions. We quantify this shift using KL divergence between the original and perturbed distributions. A component is considered causal if its ablation causes a relatively high shift to other components and leads to systematic changes in model behavior, explored in Section~\ref{sec:c identification} through various multi-relational prediction tasks. 

We further validate components in Section~\ref{sec:component_steering}, where we attempt to \textbf{steer} model behavior. Specifically, we ablate components associated with prompted information, then \textbf{amplify} one or more other components unrelated to the prompt, i.e., raise the activation for each SAE feature in the component by a proportion of its maximum activation observed during activation collection. We refer to this proportion as a \textbf{steering strength} $\alpha$.
% This proportion, henceforth \textbf{steering strength}, is a tuned hyperparameter.
Steering is \textit{successful} when it causes the model to generate a counterfactual response consistent with amplified component(s) rather than components associated with prompted information.

% \textbf{ablate} (i.e., set to zero) or \textbf{amplify}

% Additional details and results are provided in Section~\ref{sec:results}.

\begin{figure}
    \centering
    \includegraphics[width=0.97\linewidth,trim={0 0 0 3.5em},clip]{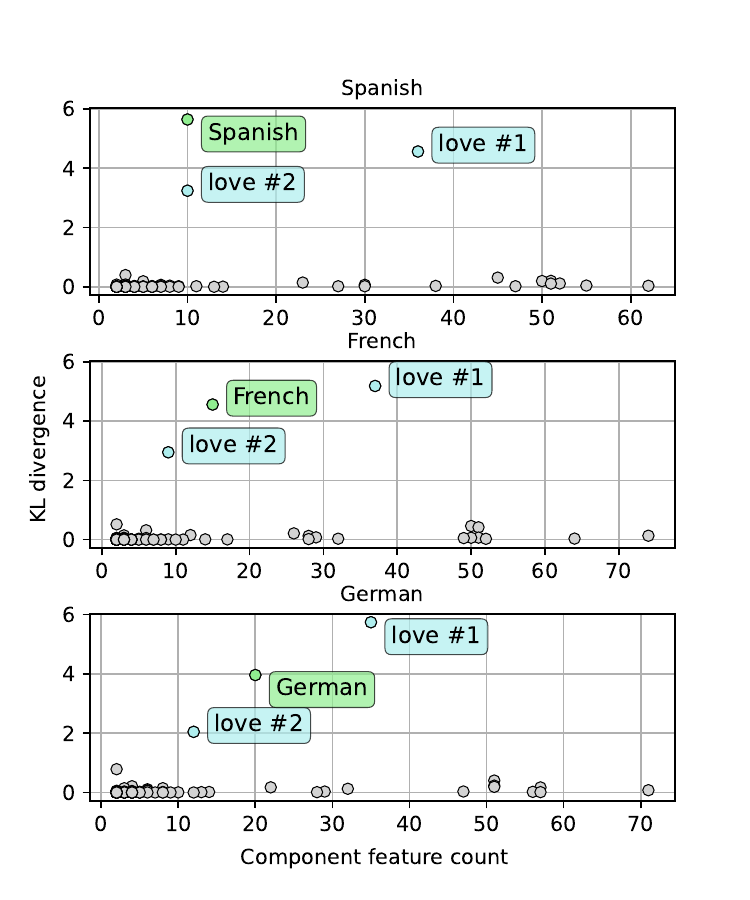}

    \vspace{-10pt}
    
    \caption{We extract components from LLM queries to translate \textit{love} into Spanish, French, and German. Each component is plotted with its feature count on the $x$-axis and the total KL divergence between pre- and post-ablation output token distributions on the $y$-axis. For these and most other concept-relation pairs, only a few components exert significant causal effects. For \textit{love}, the top three components for each language consistently correspond to either the queried word or language.}
    % \jycc{Y axis the scale is different from Y axis in Figure 4? is it possible to make it consistent? }

    \vspace{-10pt}
    
    \label{fig:component_importance} 
\end{figure}

\section{Experimental Results}\label{sec:results}

% \jycc{since tasks and name of the tasks are very important throughout the whole paper, also concepts/relations, it's important to give an introduction of task name (which used later, e.g., verb transformation) and in each task, what is considered concept or relation. Can use itemize to make it easier to read.}
We consider 3 multi-relational prediction tasks where the model is queried to generate a response (e.g., \textit{Beijing}) that requires composing a given \textbf{concept} (e.g., \textit{China}) and \textbf{relation} (e.g., \textit{capital city}):\footnote{Latter 2 tasks adapted in part from data used in \citet{todd2024function}. See Appendix~\ref{apx:evaluation schemes} for full specifications of tasks.}
\begin{enumerate}[noitemsep,nosep]
    \item \textbf{Country facts:} capital city, currency, and language (relations) of 10 countries (concepts)
    \item \textbf{Word translation:} translation of 11 English words (concepts) into Spanish, French, and German (relations)
    \item \textbf{Verb transformation:} synonym, antonym, past tense, capitalization, and first letter (relations) of 8 English verbs (concepts)
\end{enumerate}

\vspace{\parskip}

Gemma 2 2B achieves 100\% accuracy on each task.
% (2) taxonomic class, order, and family of 26 animals,
To collect model activations, we used simple prompt templates with in-context examples, e.g., for each country-capital pair, we used: ``\textit{The capital city of Peru is Lima. The capital city of South Korea is Seoul. The capital city of Saudi Arabia is Riyadh. The capital city of }\{\textit{country}\}\textit{ is}''. Similar templates were used across tasks (Appendix~\ref{apx:prompt templates}).

The remainder of this section is organized as follows. We first identify semantic components corresponding to each concept and relation (Section~\ref{sec:component_identification}). We elicit counterfactual model outputs by ablating and amplifying these components individually and in combination (Section~\ref{sec:component_steering}). Lastly, we perform several analyses to examine the organization and impact of component features across model layers (Section~\ref{sec:component_organization}), whether our approach enables more reliable manipulation of concepts and relations than individual SAE features (Section~\ref{sec:population steering analysis}), and whether ablating components for one task degrades model performance on other tasks (Section~\ref{sec:crosstask}).

\begin{table*}[h!]
    \centering
    \setlength{\tabcolsep}{3.2pt}
    \small
    \begin{tabular}{llllllllllllllllll}
        \toprule
        \multicolumn{6}{c}{\textbf{Capital, $\frac{\text{China}}{\text{Nigeria}}$}} & \multicolumn{6}{c}{\textbf{Spanish, $\frac{\text{love}}{\text{red}}$}} & \multicolumn{6}{c}{\textbf{Capitalize, $ \frac{\text{break}}{\text{like}}$}} \\
        \cmidrule[0.6pt](lr){1-6} \cmidrule[0.6pt](lr){7-12} \cmidrule[0.6pt](lr){13-18}
        \multicolumn{2}{c}{\textit{Original}} & \multicolumn{2}{c}{\textit{Ctry. Abl.}} & \multicolumn{2}{c}{\textit{Fact Abl.}} & \multicolumn{2}{c}{\textit{Original}} & \multicolumn{2}{c}{\textit{Word Abl.}} & \multicolumn{2}{c}{\textit{Lang. Abl.}} & \multicolumn{2}{c}{\textit{Original}} & \multicolumn{2}{c}{\textit{Verb Abl.}} & \multicolumn{2}{c}{\textit{Trans. Abl.}} \\
        \cmidrule[0.6pt](lr){1-2}\cmidrule[0.6pt](lr){3-4}\cmidrule[0.6pt](lr){5-6}\cmidrule[0.6pt](lr){7-8}\cmidrule[0.6pt](lr){9-10}\cmidrule[0.6pt](lr){11-12}\cmidrule[0.6pt](lr){13-14}\cmidrule[0.6pt](lr){15-16}\cmidrule[0.6pt](lr){17-18}
        Beijing & \scriptsize{97.} & Madrid & \scriptsize{39.} & the & \scriptsize{12.} & Amor & \scriptsize{41.} & {"} & \scriptsize{5.9} & Love & \scriptsize{42.} & Break & \scriptsize{22.} & Capital & \scriptsize{10.} & break & \scriptsize{20.} \\
        Be & \scriptsize{.38} & Warsaw & \scriptsize{10.} & China & \scriptsize{6.7} & amor & \scriptsize{19.} & El & \scriptsize{4.9} & love & \scriptsize{14.} & break & \scriptsize{13.} & The & \scriptsize{7.6} & Break & \scriptsize{7.1} \\
        Peking & \scriptsize{.35} & Rome & \scriptsize{9.5} & Beijing & \scriptsize{6.4} & Amo & \scriptsize{7.0} & La & \scriptsize{4.4} & I & \scriptsize{13.} & BREAK & \scriptsize{8.3} & {} & \scriptsize{4.2} & {} & \scriptsize{5.1} \\

        Shanghai & \scriptsize{.23} & Paris & \scriptsize{6.1} & Shanghai & \scriptsize{5.0} & {amo} & \scriptsize{5.1} & {Malo} & \scriptsize{3.2} & {"} & \scriptsize{5.3} & {Capital} & \scriptsize{5.7} & {I} & \scriptsize{3.6} & {capital} & \scriptsize{4.3} \\
        Xi & \scriptsize{.11} & Berlin & \scriptsize{5.0} & a & \scriptsize{2.7} & {"} & \scriptsize{2.9} & {la} & \scriptsize{2.5} & {LOVE} & \scriptsize{2.7} & {The} & \scriptsize{5.0} & {Yes} & \scriptsize{2.3} & {The} & \scriptsize{3.3} \\

        \cmidrule[0.6pt](lr){1-2}\cmidrule[0.6pt](lr){3-4}\cmidrule[0.6pt](lr){5-6}\cmidrule[0.6pt](lr){7-8}\cmidrule[0.6pt](lr){9-10}\cmidrule[0.6pt](lr){11-12}\cmidrule[0.6pt](lr){13-14}\cmidrule[0.6pt](lr){15-16}\cmidrule[0.6pt](lr){17-18}
        Abuja & \scriptsize{85.} & New & \scriptsize{7.7} & Lagos & \scriptsize{17.} & rojo & \scriptsize{38.} & La & \scriptsize{3.5} & Red & \scriptsize{27.} & Like & \scriptsize{21.} & Capital & \scriptsize{5.6} & like & \scriptsize{12.} \\
        Lagos & \scriptsize{11.} & Islamabad & \scriptsize{7.3} & the & \scriptsize{12.} & Rojo & \scriptsize{34.} & {} & \scriptsize{3.4} & red & \scriptsize{15.} & like & \scriptsize{12.} & The & \scriptsize{4.7} & {} & \scriptsize{6.1} \\
        ... & \scriptsize{.38} & Kathmandu & \scriptsize{6.0} & Nigeria & \scriptsize{7.1} & rojo & \scriptsize{2.1} & {"} & \scriptsize{3.2} & " & \scriptsize{8.1} & Capital & \scriptsize{8.6} & Yes & \scriptsize{4.5} & capital & \scriptsize{5.6} \\

        Nigeria & \scriptsize{.29} & Delhi & \scriptsize{5.7} & called & \scriptsize{5.1} & {Ro} & \scriptsize{2.0} & {El} & \scriptsize{2.8} & {The} & \scriptsize{5.1} & {LIKE} & \scriptsize{4.3} & {No} & \scriptsize{4.1} & {the} & \scriptsize{4.6} \\
        … & \scriptsize{.27} & Tehran & \scriptsize{4.9} & Abuja & \scriptsize{3.4} & {Roja} & \scriptsize{1.9} & {Bol} & \scriptsize{2.4} & {...} & \scriptsize{4.1} & {The} & \scriptsize{4.0} & {} & \scriptsize{3.3} & {Like} & \scriptsize{3.8} \\
        \bottomrule
    \end{tabular}
    \caption{Top 5 output tokens and their likelihoods for various concept-relation pairs, before and after ablating concept or relation components. Underscore prefixes of tokens omitted for space. More examples in Appendix~\ref{apx: additional examples ablation}.}

    % \vspace{-10pt}
    
    \label{tab:ablation_top_tokens}
\end{table*}

\subsection{Component Identification}\label{sec:c identification}
\label{sec:component_identification}

For each concept-relation pair, we obtained sparse components using the methods from Section~\ref{sec:methods}. As shown in Figure~\ref{fig:component_importance} for translation,\footnote{Additional examples in Appendix~\ref{apx:component_importance}.} 2-3 of the components typically exerted a markedly higher causal effect on the model output than the others.

\paragraph{Semantic coherence.}
\revision{Given several impactful components, we then identify \textit{semantically coherent} representative components for each concept and relation, i.e., components that yield predictable counterfactual outputs when manipulated (like in Figure~\ref{fig:overview}). We explored a few approaches.} First, we inspected their associated feature descriptions from Neuronpedia.\footnote{These descriptions from Neuronpedia were generated by prompting GPT-4o mini~\citep{openai2024gpt4ocard} to summarize the texts that activate each feature over a large corpus.}
In many cases, the features within components had thematically coherent descriptions referring to common concepts or relations, e.g., most country fact task components (Appendix~\ref{subsec:interpretation_viz}).

However, these descriptions were often unreliable. For instance, none of the components obtained for prompts about Spain mentioned Spain in their feature descriptions, and some components for the word \textit{beautiful} included features about Scotland.
Additionally, some components for translation languages were described to be about programming. This suggests that even SAE features suffer from some polysemanticity.
%, possibly due to limitations in the corpora and batching strategies used for training SAEs.
% NH: perhaps not necessary
Notably, this shows that it is impossible to consistently extract comparable components by simple keyword search over SAE feature descriptions\revision{, necessitating methods like ours to group features based on task-specific activations.}

To address this in the country facts task, we ablated components and observed the resulting changes in the model's top predicted tokens, selecting those that yielded the most compositional changes.
As an example, the left 3 columns of Table \ref{tab:ablation_top_tokens} present results of ablating selected components obtained from prompts for the capital of China and Nigeria.
Promisingly, when ablating country components, the model's top predicted tokens shifted to the capitals of other countries. When ablating fact components, the model assigned higher probabilities to country names.

For the remaining tasks, this approach was unsuccessful, possibly due to more complex composition operations at play than can be captured by a single component for a concept or relation.
In these tasks, we instead represented each concept and relation by the union of multiple seemingly relevant causally impactful components. 
As shown in the other columns of Table~\ref{tab:ablation_top_tokens}, ablating these combined components for words/verbs in these tasks similarly caused the model to output other words in the prompted language of Spanish, or other capitalized words. Meanwhile, ablating relation components caused the model to repeat the input word.

% \begin{figure}
%     \centering
%     \includegraphics[width=1\textwidth]{figures/gemma-2-2b/components/china_capital.png}
%     \caption{Extracted components for China (blue) and country language (green).}
%     \label{fig:china_component_capital_currency}
% \end{figure}

\begin{figure*}[h!]
    \centering
    \includegraphics[width=0.98\textwidth]{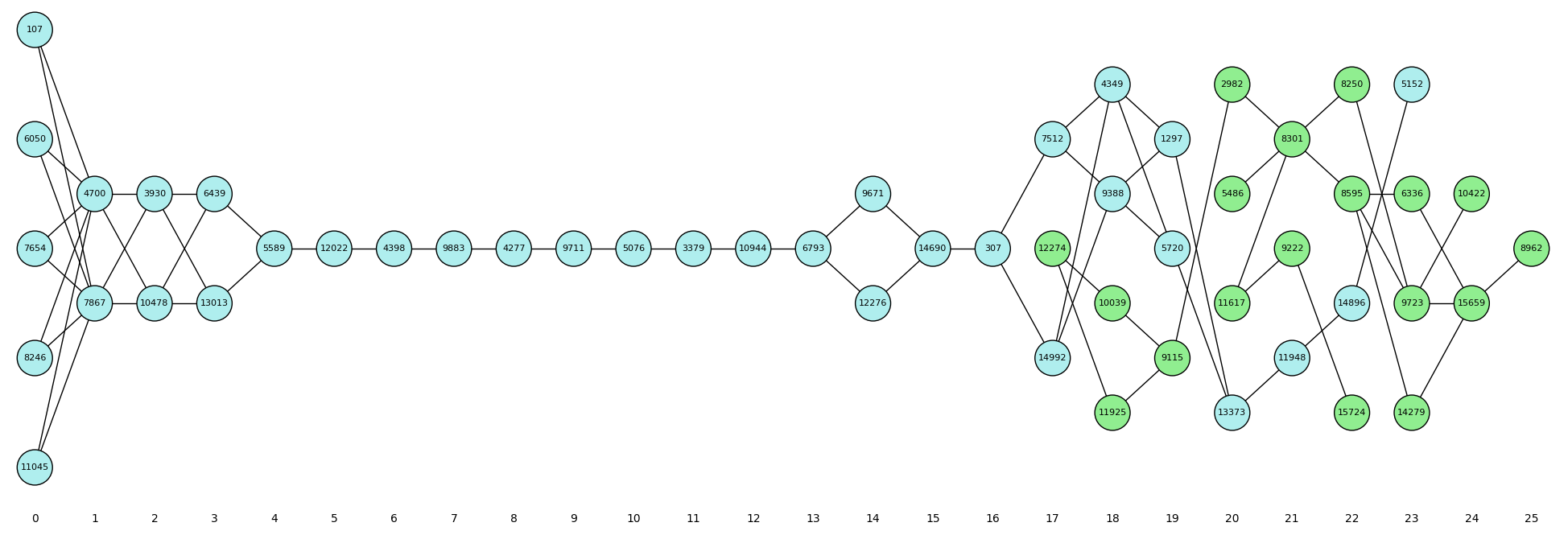}

    \vspace{-5pt}
    \caption{Components representing China (blue) and country language (green), visualized by network layer.}
    \label{fig:china_language_components}
\end{figure*}

% \begin{figure}
%     \centering
%     \begin{minipage}[b]{0.45\textwidth}
%         \centering
%         \includegraphics[width=\textwidth]{figures/gemma-2-2b/components/language_china.png}
%     \end{minipage}
%     \hspace{0.05cm}
%     \begin{minipage}[b]{0.45\textwidth}
%         \centering
%         \includegraphics[width=\textwidth]{figures/gemma-2-2b/components/language_nigeria.png}
%     \end{minipage}
%     \caption{Language components extracted from China and Nigeria prompts are nearly identical.}
%     \label{fig:language_component_china_nigeria}
% \end{figure}

\paragraph{Context consistency.} Most concept and relation components were remarkably consistent across different contexts, e.g., the top China components obtained from capital and currency prompts were identical, and the language components from China and Nigeria prompts differed only by a few edges. Therefore, 
for country facts and translation,
% except animal taxonomy
we define each concept component as the intersection of all the components for that concept across relations; similarly, each relation component is defined as the intersection of all the components for that relation across concepts. For verb transformations, while many components appeared context-insensitive, using the specific components identified for each prompted concept-relation pair maximized steering performance. Figure~\ref{fig:china_language_components} visualizes the resulting China and language components, and more examples are provided in Appendix~\ref{apx:more components}.

\begin{table*}
    \centering
    \setlength{\tabcolsep}{4.2pt}
    \small
    \begin{tabular}{cccccccccccccccc}\toprule
        & & \multicolumn{4}{c}{\textit{Country Facts}}  & \multicolumn{4}{c}{\textit{Word Translation}} & \multicolumn{6}{c}{\textit{Verb Transformation}} \\
       \textbf{C} & \textbf{R} & \textbf{Cap.} & \textbf{Curr.} & \textbf{Lang.} & \textbf{Avg.} & \textbf{Spanish} & \textbf{French} & \textbf{German} & \textbf{Avg.} & \textbf{Syn.} & \textbf{Ant.} & \textbf{Past T.} & \textbf{Cap.} & \textbf{1st L.} & \textbf{Avg.} \\\cmidrule(lr){1-2}\cmidrule(lr){3-6}\cmidrule(lr){7-10}\cmidrule(lr){11-16}
       \cmark & \xmark & .98 & .93 & .96 & .96  & .74 & .74 & .78 & .75 & .61 & .45 & .61 & .18 & .57 & .48 \\\cmidrule(lr){1-2}\cmidrule(lr){3-6}\cmidrule(lr){7-10}\cmidrule(lr){11-16}
       \xmark & \cmark & 1.0 & .85 & .95 & .93  & 1.0 & .95 & 1.0 & .98 & .00 & .00 & .00 & .31 & .81 & .23 \\\cmidrule(lr){1-2}\cmidrule(lr){3-6}\cmidrule(lr){7-10}\cmidrule(lr){11-16}
       \cmark & \cmark & 1.0 & .77 & .93 & .90 & .57 & .69 & .67 & .64 & .02 & .00 & .00 & .16 & .79 & .19  \\\bottomrule
     \end{tabular}
    \caption{Steering success rates (SR) for all tasks by relation and averaged. Row 1: SR of steering from one concept to another (C\cmark) while preserving relation (R\xmark). Row 2: SR of steering to each relation from others (R\cmark) while preserving concept (C\xmark). Row 3: SR of steering to concept-relation pairs from others by target relation (C\cmark, R\cmark). }

    % \vspace{-10pt}
    
    \label{tab:all steering}
\end{table*}

\subsection{Component Steering}
\label{sec:component_steering}

Having identified distinct components for each concept and relation, we next investigate whether they can be used to steer model outputs individually and in combination.\footnote{See Appendix~\ref{apx:steering strength} for details on steering strength tuning.}
To test whether components generalize across different contexts, we applied a zero-shot prompt template different from the one used to collect the initial activations, e.g., for translation: \textit{``Q: What is the }\{\textit{Spanish/French/German}\} word for '\{\textit{word}\}'\textit{? Answer directly (two words max). A:''}. We used comparable prompts for other tasks, as listed in Appendix~\ref{apx:prompt templates}.
% For animal taxonomy, the model performed poorly zero-shot, so we reverted to few-shot prompts for this task.} 

Table~\ref{tab:all steering} summarizes the results by task and relation, while Appendix~\ref{apx: additional tables} further breaks down performance by concepts, and  Appendix~\ref{apx:steering failure analysis} discusses common and interesting steering failure cases.

\begin{table*}
    \centering
    \setlength{\tabcolsep}{2.2pt}
    \small
    \begin{tabular}{llllllllllllllllll}
    \toprule
    \multicolumn{6}{c}{\textbf{Language, CN}} & \multicolumn{6}{c}{\textbf{French, love}} & \multicolumn{6}{c}{\textbf{Past Tense, understand}} \\
    \cmidrule(lr){1-6} \cmidrule(lr){7-12} \cmidrule(lr){13-18}
    \multicolumn{2}{c}{\textit{$\cdot$, NG}} & \multicolumn{2}{c}{\textit{Capital, $\cdot$}} & \multicolumn{2}{c}{\textit{Currency, $\cdot$}} & \multicolumn{2}{c}{$\cdot$, \textit{red}} & \multicolumn{2}{c}{\textit{Spanish, $\cdot$}} & \multicolumn{2}{c}{\textit{German, $\cdot$}} & \multicolumn{2}{c}{\textit{$\cdot$, break}} & \multicolumn{2}{c}{\textit{Capitalize, $\cdot$}} & \multicolumn{2}{c}{\textit{1st Letter, $\cdot$}} \\
    \cmidrule(lr){1-2}\cmidrule(lr){3-4}\cmidrule(lr){5-6}\cmidrule(lr){7-8}\cmidrule(lr){9-10}\cmidrule(lr){11-12}\cmidrule(lr){13-14}\cmidrule(lr){15-16}\cmidrule(lr){17-18}
    \hl{\_English} & \scriptsize{71.} & \hl{\_Beijing} & \scriptsize{96.} & \hl{\_Yuan} & \scriptsize{12.} & \hl{\_rouge} & \scriptsize{44.} & \hl{\_amor} & \scriptsize{41.} & \hl{\_Liebe} & \scriptsize{68.} &  \hl{\_broke} & \scriptsize{28.} & \hl{\_Understand} & \scriptsize{6.7} & \hl{\_U} & \scriptsize{39.} \\
    {\_Yoruba} & \scriptsize{5.6} & \hl{\_Be} & \scriptsize{2.0} & \hl{\_Ren} & \scriptsize{8.4} & \hl{\_Rouge} & \scriptsize{30.} & \hl{\_Amor} & \scriptsize{13.} & {\_die} & \scriptsize{2.4} & \hl{\_Broke} & \scriptsize{21.} & {\_Know} & \scriptsize{6.4} & {\_understood} & \scriptsize{9.3} \\
    {\_Ha} & \scriptsize{4.8} & \hl{\_Peking} & \scriptsize{.93} & {\_China} & \scriptsize{6.1} & {\_Sang} & \scriptsize{2.5} & {\_el} & \scriptsize{6.9} & {\_„} & \scriptsize{2.1} & \hl{\_BRO} & \scriptsize{8.5} & {\_Past} & \scriptsize{5.9} & {\_Understand} & \scriptsize{8.5} \\
    {\_Nigeria} & \scriptsize{4.4} & \hl{\_BE} & \scriptsize{.53} & {\_Chinese} & \scriptsize{5.9} & \hl{\_Rou} & \scriptsize{1.8}    & {\_"} & \scriptsize{6.1}      & \hl{\_lieben} & \scriptsize{1.7}  & {\_Bre} & \scriptsize{6.3} & {\_Answer} & \scriptsize{5.7} & \hl{\_u} & \scriptsize{3.6} \\
    {\_Igbo} & \scriptsize{3.4} & \hl{Beijing} & \scriptsize{.37} & \hl{\_RMB} & \scriptsize{5.9} & {\_} & \scriptsize{1.2}       & \hl{\_A} & \scriptsize{3.1}      & {\_Ich} & \scriptsize{1.6} & {\_brake} & \scriptsize{2.6} & {\_} & \scriptsize{4.5} & {\_United} & \scriptsize{3.2} \\
    
    \bottomrule
    \end{tabular}
    \caption{Top 5 output tokens and likelihoods for prompts about China's language (columns 1–3), \textit{love}'s French translation (columns 4–6), and \textit{break}'s past tense (columns 7–9), after ablating an in-prompt concept or relation component (row 1 headings) and amplifying a target concept or relation component (row 2 headings). Highlighted tokens are or may begin correct answers for steered components. More examples in Appendix~\ref{apx: additional examples ind steering}. }
    % \vspace{-10pt}
    
    \label{tab:steering_top_tokens}
\end{table*}

\paragraph{Concept steering.} 

% \jycc{how to amplify? say a few words here since this operation is used throughout causal analysis.}

By ablating an \textbf{in-prompt concept} component and amplifying a \textbf{target concept} component, we successfully directed the LLM to respond to questions about relations for the {in-prompt concept} with \textit{counterfactual answers} (i.e., from applying the relation to the {target concept} which is not actually queried in the prompt) 48-96\% of the time on average (Table~\ref{tab:all steering}). 
% As shown in the first row of Table~\ref{tab:all steering}, we observed respective average steering success rates of 96\%, 75\%, and 48\% for country facts, word translation, and verb transformation. 
For example, as shown in Table~\ref{tab:steering_top_tokens}, the model consistently disregarded prompts' references to China, \textit{love}, and \textit{understand}, instead outputting counterfactual answers ``\textit{English}'' (Nigeria's language), ``\textit{rouge}'' (\textit{red} in French), and ``broke'' (\textit{break}'s past tense). 
Even when incorrect answers were among top tokens, they were often relevant to the in-prompt relation and target concept, e.g., the model ranked ``\textit{Yoruba},'' ``\textit{Ha$\lbrack$usa$\rbrack$},'' and ``\textit{Igbo}'' (other commonly spoken languages in Nigeria) just behind the correct answer of ``\textit{English},'' while it ranked a French word for \textit{blood} just behind words for the target concept of \textit{red}.
% Additionally, when we ablated the \textit{love} component and amplified the \textit{red} component, the model responded with Spanish, French, and German words for \textit{red} despite being prompted to translate \textit{love}. 
% \textcolor{red}{Tables \ref{tab:country_steering}-\ref{tab:translation concept steering}} present results across concepts for all tasks, with concept steering successfully producing the desired counterfactual answers \textcolor{red}{96\%} of the time. 
This confirms that our identified concept components encode concept-specific information that causally determines model outputs.

%We observed the most failures in the verb transformations task. Some verbs had much lower concept steering success rates than others — for example, possess (6\%) and sink (20\%). These verbs were disproportionately responsible for failures both as in-prompt verbs (where their components could not be effectively ablated) and as target verbs (where their components could not effectively steer the model). This suggests that our method did not yield stable components for these particular verbs. Further, when steering from break and hide to sink under the past tense relation, the model generates sought and slid rather than sank — both past tense forms of other verbs beginning with s. This raises the possibility that the component identified for sink may capture words beginning with s rather than the verb sink specifically.

We observed the most failures in the verb transformations task. Notably, some verbs had a much lower concept steering success rate than others, e.g., \textit{possess} (6\%) and \textit{sink} (20\%).
%{\color{red}We observed that these verbs were often in-prompt verbs in failures to steer to other target verbs.}
{{These same verbs caused failures in both directions: steering to them (as target verbs) and steering away from them (as in-prompt verbs).}}
This suggests that our method did not yield stable components for these particular verbs.
%, causing both failures to ablate and steer based on them.
Interestingly, we saw that when steering to \textit{sink} from \textit{break} and \textit{hide} {under the past tense relation}, the model respectively generated ``\textit{sought}'' and ``\textit{slid}'', possibly suggesting that components for \textit{sink} may instead represent words that begin with \textit{s}. Meanwhile, the capitalize relation had the lowest success rate; we observed that the model usually output the target word, but not capitalized. For \textit{hide}, the model instead output the target verb in all capital letters. This may suggest that components for specific verbs can override prompt instructions when used for steering, perhaps indicating a lower degree of composability than other tasks.

\begin{table*}
    \centering
    \setlength{\tabcolsep}{9.9pt}
    \small
    \begin{tabular}{llllllllllll}
    \toprule
    \multicolumn{4}{c}{\textbf{Currency, China}} & \multicolumn{4}{c}{\textbf{German, love}} & \multicolumn{4}{c}{\textbf{Antonym, break}} \\
    \cmidrule(lr){1-4} \cmidrule(lr){5-8} \cmidrule(lr){9-12}
    \multicolumn{2}{c}{\textit{Capital, Nigeria}} & \multicolumn{2}{c}{\textit{Language, Nigeria}} & \multicolumn{2}{c}{\textit{Spanish, red}} & \multicolumn{2}{c}{\textit{French, red}} & \multicolumn{2}{c}{\textit{Synonym, like}} & \multicolumn{2}{c}{\textit{1st Letter, like}} \\
    \cmidrule(lr){1-2}\cmidrule(lr){3-4}\cmidrule(lr){5-6}\cmidrule(lr){7-8}\cmidrule(lr){9-10}\cmidrule(lr){11-12}
    \hl{\_Abuja} & \scriptsize{70.} & \hl{\_English} & \scriptsize{96.} & \hl{\_rojo} & \scriptsize{53.} & \hl{\_rouge} & \scriptsize{45.} & \hl{\_love} & \scriptsize{56.} & {\_like} & \scriptsize{31.} \\
    {\_Lagos} & \scriptsize{25.} & {\_Yoruba} & \scriptsize{1.1} & {\_el} & \scriptsize{7.2} & {\_le} & \scriptsize{6.7} & \_loved & \scriptsize{8.5} & \_Like & \scriptsize{20.} \\
    \_ & \scriptsize{3.3} & {\_French} & \scriptsize{.74} & \hl{\_roja} & \scriptsize{5.6} & \hl{\_Rouge} & \scriptsize{4.2} & \_few & \scriptsize{3.8} & \hl{\_L} & \scriptsize{6.3} \\
    {\_Nigeria} & \scriptsize{.34} & {\_Spanish} & \scriptsize{.60} & {\_"} & \scriptsize{3.1} & {\_} & \scriptsize{3.7} & \hl{\_loves} & \scriptsize{3.6} & \_ant & \scriptsize{3.2} \\
    {\_...} & \scriptsize{.32} & {\_Igbo} & \scriptsize{.54} & {\_El} & \scriptsize{2.2}         & {\_la} & \scriptsize{3.4} & \_lot & \scriptsize{3.4} & \_Ant & \scriptsize{2.5} \\
    \bottomrule
    \end{tabular}
    \caption{Top 5 output tokens and their likelihoods for prompts about China's currency (columns 1–2), \textit{love}'s German translation (columns 3–4), and \textit{break}'s antonym (columns 5–6), after ablating both in-prompt components and amplifying components for a target concept-relation pair. More examples in Appendix~\ref{apx: additional examples comp steering}.}

    % \vspace{-10pt}
    
    \label{tab:composed_steering_top_tokens}
\end{table*}

\paragraph{Relation steering.} By ablating an \textit{in-prompt relation} while amplifying a \textit{target relation}, we also successfully directed the model to respond to queries about the {in-prompt relation} as though they concerned the {target relation} 23-98\% of the time (Table~\ref{tab:all steering}). 
% As shown in the second row of Table~\ref{tab:all steering}, we observed respective average relation steering success rates of 93\%, 92\%, and 26\% for country facts, word translation, and verb transformation. 
As shown in Table~\ref{tab:steering_top_tokens}, the model disregarded prompts' references to country language, French translation, and past tense transformation, and produced correct counterfactual answers for other relations within each task, e.g., answering prompts to translate \textit{love} into French with Spanish and German words for \textit{love}. 
Even when incorrect answers were among top tokens, they were often relevant, e.g., words in the target language, or ``\textit{Know}'' for the capitalized form of \textit{understand} (a synonym for the correct answer). In some cases, correct answers competed with answers related to the prompt context, suggesting that some in-prompt information remains activated. For example, ``\textit{Chinese}'' appeared among the top tokens when steering to the currency relation even after the in-prompt component for the language relation was ablated. Similarly, ``\textit{understood}'' appeared among top tokens when steering to first letter after the in-prompt component for past tense was ablated.
% The only failure was the target transformation of capitalize, which yielded capitalized words other than the in-prompt verb \textit{break}.

% Table~\ref{tab:relation_steering} shows that relation steering is effective across all three relations, achieving an average success rate of 92\%.

Unsurprisingly, the most failures again occurred in this task, specifically for the synonym, antonym, and past tense transformations. For the former two, this may be expected. Unlike the capital city or currency of a country, synonym and antonym do not have objective answers, and actually depend on a variety of finer-grained dimensions, e.g., word sense and part of speech. As observed in previous visualized token distributions after ablation and steering, the country fact and word translation components have clear roles in model outputs: to promote tokens related to specific countries or words, or that are relevant to a target fact (languages, currencies, or cities) or in a target language.
However, a mechanism for synonym or antonym relations should not be expected to work like this, as many tokens could be a synonym or antonym, and reasonable answers thus depend more heavily on context. 
While past tense forms of verbs are more consistent, this relation is much broader and less specific than relations in other tasks, as every verb has a past tense form, and thus an LLM learning a specialized module to promote past tense inflections may be inefficient.
Such relations are thus likely better captured in attention heads, which aligns with \citet{todd2024function}'s finding that antonym and past tense are modeled well by function vectors.

\paragraph{Composite steering.} We conducted composite steering experiments where both concept and relation components were manipulated at once. By ablating both the in-prompt concept and {in-prompt relation} components while amplifying the {target concept} and {target relation} components, it was possible to steer the model to ignore both the {in-prompt concept} and {in-prompt relation} and answer about a different concept-relation pair 19-90\% of the time (Table~\ref{tab:all steering}). Table~\ref{tab:composed_steering_top_tokens} provides specific examples of composite steering success. As shown in the first 2 columns, when we ablated both the China and capital components while amplifying the Nigeria and currency components, the model answered ``\textit{Naira}'' despite being asked about China's capital. In the middle 2 columns, we see similar success in steering the model's outputs to Spanish and French words for \textit{red}, despite being prompted for the German translation of \textit{love}. In the last 2 columns, when prompted for an antonym of \textit{break}, we successfully steer the model to output a synonym of \textit{like}, while steering it to output the first letter of \textit{like} was less successful, with the correct answer only ranked third after ``\textit{like}'' and ``\textit{Like}''. We also observed interesting failure cases where in-prompt information intermingled with target information, e.g., steering from the antonym of \textit{break} to \textit{include} capitalized yielded ``\textit{Exclude},'' a capitalized antonym of \textit{include}.
Nonetheless, these findings demonstrate significant composability of many components.

% \footnote{For example, the forthcoming steering failure analysis as steering with a particular component for \textit{here} caused the model to output common locations for the prompted word, e.g., a word meaning \textit{ocean} for the prompted word \textit{fish} in both Spanish and French, a nontrivial operation.} 

% Table~\ref{tab:relation_country_steering} presents comprehensive results for all concept-relation pairs, showing an average steering success rate of 90\%.

% \section{Analysis Results}
% Lastly, we present additional analyses to provide more information about the nature of extracted components. 
% Lastly, in Section~\ref{sec:syntactic vs semantic}, we explore the generalizability of our approach beyond declarative fixed knowledge to context-sensitive linguistic tasks.

\subsection{Component Organization}
\label{sec:component_organization}

Having established the causal role and composability of concept and relation components, we next analyze their distribution across network layers and the relative importance of nodes within these components.
% specifically for the country facts task where we achieved the best performance. 
To quantify the causal importance of an individual node in a concept component, we compute the average post-ablation KL divergence from the original output distribution across all relations. For nodes in relation components, we compute the same average across all concepts. 
Figure~\ref{fig:country_node_importance} displays results for the country facts task, with more examples in Appendix~\ref{apx:component_organization}. 

Concept and relation components show distinct distribution patterns across network layers. 8 of 10 country components begin in the first layer, as do all word/verb components in other tasks. Some concept components span nearly all layers (e.g., for China and \textit{love}), while others concentrate in early to middle layers (e.g., for Nigeria and \textit{red}). In contrast, relation components concentrate in later layers, with most spanning only the last quarter to half of the network's layers (e.g., for country facts and target translation languages), and some spanning further across the network while having greater densities of nodes in later layers (e.g., for verb transformations). This may suggest that representations of more concrete entities are established earlier in processing, while more abstract relations emerge later. 
Concept component nodes exhibit various trends in KL divergence across layers. Meanwhile, relation components for country facts, past tense, and first letter have stronger causal impact in later layers, and the opposite trend holds for translation languages and capitalization.

% Country and fact components show distinct distribution patterns across network layers. For example, 8 of 10 country components tested begin in the first layer; some (e.g., China) span nearly all layers, while others (e.g., Nigeria) concentrate in early to middle layers. In contrast, all 3 country fact components appear only in later layers. This suggests that representations of concrete entities are established earlier in processing, while more abstract relations emerge later. Not only do all 3 country fact components concentrate in later layers, but we find that, even within each fact component, nodes from later layers tend to have a stronger causal impact on model outputs. This is not the case for country nodes, which exhibit variable relationships between layer depth and KL divergence ranging from positive to negative.

% \begin{figure}
%     \centering
%     \includegraphics[trim={0 0 0 3em},clip,width=1\linewidth]{figures/gemma-2-2b/node_kl_divergence_by_layer.pdf}
%     \vspace{-20pt}
%     \caption{Node-wise KL divergence between pre- and post-ablation output token distributions.}
%     \label{fig:node_importance}
% \end{figure}

\begin{figure*}
    \centering
    \includegraphics[width=0.99\linewidth,trim={2em 0 4em 2.5em},clip]{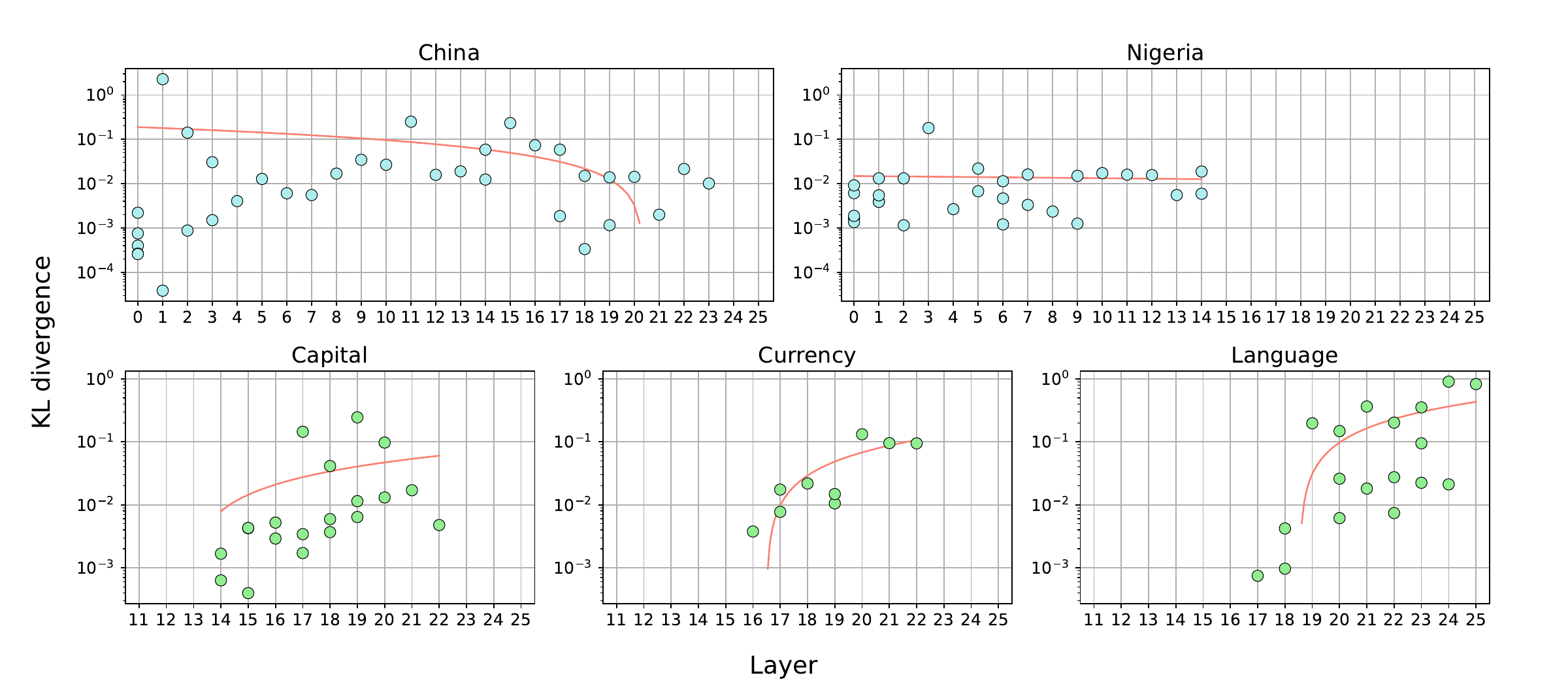}
    \vspace{-5pt}
    \caption{KL divergence between pre- and post-ablation output token distributions for each node in the China, Nigeria, and country fact components, plotted by layer. Linear regression lines plotted in red.}

    % \jycc{linear regression line not straight?}

    % \vspace{-10pt}
    
    \label{fig:country_node_importance}
\end{figure*}

% \begin{figure}
%     \centering
%     \includegraphics[width=1.02\linewidth]{figures/gemma-2-2b/relation_node_kl_divergence_by_layer.png}
%     \vspace{-15pt}
%     \caption{KL divergence between pre- and post-ablation output token distributions for each node in the relation components, plotted by layer. Linear regression line is in red. All three components emerge in later (14 and beyond) layers.}
%     \label{fig:relation_node_importance}
% \end{figure}

\subsection{Single Feature Steering Comparison}\label{sec:population steering analysis}
We hypothesize that our approach of steering based on a group of features is advantageous over steering based on an individual feature (as done in some prior works, e.g., \citealp{anthropic2023decomposing,templeton2024scaling}).
To validate this, we compared its efficacy against a simpler baseline: intervening on only the single most causally impactful feature. Specifically, we identified this top feature for each concept and relation as the one whose individual ablation produced the highest KL divergence, then in line with earlier results, used it for ablation and steering.
% This top-feature steering method was significantly less effective. 

For the country facts task, the baseline respectively achieved average success rates of 83\%, 83\%, and 75\% for concept, relation, and composite steering. These results fall notably short of the success rates achieved using full components (respectively 96\%, 93\%, and 90\%). This suggests that concepts are represented distributively across populations of features, and our method's success is largely attributable to identifying and manipulating these entire functional groups.

\subsection{Specificity Evaluation}\label{sec:crosstask}

Lastly, we explore the degree to which ablating our identified components impacts performance on other tasks, an important question for the specificity of mechanisms and practicality of manipulating models based on them. Table~\ref{tab:crosstask} presents the model's average accuracy for each task when our identified components are ablated. When ablating components from the country facts task, 100\% accuracy (or near it) is maintained in most cases, even when ablating components for some country-fact pairs and prompting the model for other ones. Accuracy on translation degrades slightly, possibly due to the intersection of some country- and language-specific information.
In ablating word translation components, accuracy remains high on other tasks, but degrades slightly on translation itself. This may be expected, given that the relations in this task are of highly similar nature, and even overlap by a small number of features. 
In ablating verb transformation components, we see moderately degraded accuracy for all tasks to as low as 62-73\%. This suggests that the components for this task are less specific than other tasks, in line with our relatively low steering performance with them.
Nonetheless, this demonstrates that extracted components are generally fairly specific to the functions we associate them with, promising for manipulating deployed models with them.

\begin{table}
    \centering
        \setlength{\tabcolsep}{3pt}
    \footnotesize
    \begin{tabular}{cccc}\toprule
         % & \multicolumn{3}{c}{\textit{Task Accuracy}}\\
        \textbf{Ablated Task} & \textbf{\thead{Country\\Facts Acc.}} & \textbf{\thead{Word\\Transl. Acc.}} & \textbf{\thead{Verb\\Transf. Acc.}} \\\midrule
        \textit{Country Facts} & 1.00 & 0.93 & 1.00  \\
        \textit{Word Transl.} & 1.00 & 0.83 & 1.00 \\
        \textit{Verb Transf.} & 0.73 & 0.64 & 0.63  \\\bottomrule
    \end{tabular}
    \caption{Prediction accuracy over concept-relation pairs in each task when components from the same or another task are ablated. If the same task, cases where prompted concept-relation pair overlaps with ablated concept-relation pair are omitted from evaluation.}

    % \vspace{-10pt}
    
    \label{tab:crosstask}
\end{table}
\section{Related Work}

% \cite{bayat2025steering} % This paper proposes a steering method based on constrative paired prompting and SAE features - show some small-scale (smaller-scaler than us) compositionality of features
% 

\paragraph{Circuits and sparse feature dictionaries.}

Prior work has applied various methods to interpret large neural networks as compositions of subnetworks that perform specific functions~\citep{olah2020zoom,elhage2021mathematical,sharkey2025openproblemsmechanisticinterpretability}. Specialized circuits have been discovered for various functions of LLMs, including in-context learning \citep{olsson2022context}, numerical comparison~\citep{hanna2023does}, indirect object recognition~\citep{wang2023interpretability}, and broader natural language fluency \cite{alkhamissi-etal-2025-llm}.
More recent work has developed automated circuit discovery methods to improve scalability \citep{goldowskydill2023localizingmodelbehaviorpath,conmy2023towards,hsuefficient,bhaskar2024finding}.
However, these approaches can be computationally expensive and yield circuits that are difficult to interpret.
As large neural networks like LLMs are thought to encode sparse features across polysemantic neurons \citep{elhage2022superposition}, dictionary learning techniques like SAEs promise greater interpretability by extracting monosemantic feature directions from them~\citep{bricken2023monosemanticity,huben2024sparse}.
This has allowed automated circuit discovery approaches to operate on these features instead of neurons~\citep{marks2025sparse}, but high computational costs remain a barrier.
Like our work, \citet{e27040344} used SAE feature coactivation patterns to analyze their geometry, finding spatial clustering of related concepts.
Standard SAEs can also suffer from inconsistent feature quality, poor reconstruction, and weak functional alignment, prompting various architectural and training improvements~\citep{braun2024identifying,rajamanoharan2024improvingdictionarylearninggated,DBLP:journals/corr/abs-2407-14435}.
Transcoders~\citep{dunefsky2024transcoders} extend SAEs by learning to approximate layerwise computations. Cross-layer transcoders comprise the Anthropic circuit tracing tool \citep{ameisen2025circuit}, which visualizes LLM responses with sparse but still {difficult-to-interpret} computation graphs. Building on these efforts, we follow \citeauthor{e27040344} to leverage SAE feature coactivation, applying node pruning based on activation density to automatically discover semantically coherent, context-consistent components that can influence LLMs' relational reasoning outputs individually and in combination. Our approach offers an efficient framework for analyzing and controlling LLM behavior without exhaustive circuit tracing.

\paragraph{Knowledge organization in LLMs.}

Orthogonally, prior work has used various methods to study the organization of lexical and factual knowledge in LLMs. Factual knowledge is believed to reside in the feedforward layers of transformer-based LLMs.
\citet{geva-etal-2021-transformer} characterized these layers as key-value memories mapping textual patterns to vocabulary distributions.
\citet{dai2022knowledge} identified ``knowledge neurons'' whose activations correlate with specific facts.
These insights informed approaches for editing factual knowledge stored in LLMs~\citep{de-cao-etal-2021-editing,mitchell2022fast,meng2022locating}.
\citet{geva2023dissecting} described the recall of factual associations as a three-step process of subject enrichment, relation propagation, and attribute extraction, while \citet{hernandez2024linearity} demonstrated that relation decoding in transformers can be approximated by simple linear transformations.
Recent work on ``knowledge circuits'' has begun tracing causal pathways underlying factual recall~\citep{yao2024knowledge,ou2025llmsacquirenewknowledge}.

Highly relevant to this work, recent work has explored the composition of knowledge in LLMs. \citet{merullo2024language} demonstrated that LLMs implement word2vec-style~\cite{mikolov2013distributed} vector arithmetic through their feedforward layers to solve some relational tasks. 
\citet{bayat2025steering} leveraged active SAE features in contrastive paired prompts to steer LLM behaviors, showing evidence of composability of behaviors.
In line with our word translation results, \citet{dumas-etal-2025-separating} used activation patching to reveal language-agnostic concept representations in LLMs, such that the translation language and concept can be steered independently.
Our work extends these efforts by identifying finer-grained modular groups of feature directions that represent human-interpretable concepts and relations and similarly compose to form responses, possibly from a single relational prompt.

\section{Conclusion}\label{sec:discussion}

By tracing sparse feature coactivation patterns across layers, we uncover semantically coherent, context-consistent components encoding both concrete entities and abstract relations.
We find that ablating or amplifying these components reliably alters model behavior.
Moreover, composing concept and relation components induces compound counterfactual responses.
The layer-wise distribution of components reveals a hierarchical organization: entity features emerge in early layers, while relational features cluster in deeper layers. Overall, our method provides a lightweight, interpretable framework for analyzing and steering LLM behavior without full-scale circuit tracing.
This work advances scalable interpretation and control of LLM behavior by leveraging pretrained SAEs to efficiently identify functional components. 
Given the adoption of LLMs by humans on a greater variety of tasks, such capabilities are vital for sustaining a human-machine enterprise that is not only reliable and safe, but also transparent, aligned with human goals, and resilient to misuse and misinterpretation.

% \section*{Reproducibility Statement}
% \textcolor{red}{TODO: write a one-paragraph statement to point to where in the paper we provide important info for reproducing results (see \url{https://iclr.cc/Conferences/2026/AuthorGuide)}}
\section*{Limitations}

We acknowledge that our study is limited in the following ways:

\paragraph{Limited datasets.}

First, due to computational constraints, our investigation was constrained to only 3 multi-relational reasoning tasks with small numbers of examples. When it comes to the task space, we attempted to add some additional tasks around Spanish verb conjugation, factual information about Nobel Prize winners, and taxonomical information about animals,\footnote{More discussion about animal taxonomy results in Appendix~\ref{apx:taxonomy}.} but Gemma 2 2B could not perform these tasks accurately off the shelf (for the former, this is likely due to Gemma not being trained on much Spanish data), thus we would not expect to identify semantic modules for these tasks. That said, the tasks we explored in this paper are diverse, representing factual and lexicosemantic knowledge, and similar forms of these tasks are also considered in prior relevant works \cite{merullo2024language,todd2024function,hernandez2024linearity}. As such, we expect that our analytical framework generalizes to other domains. 

When it comes to the datasets' sizes, scaling our method up to large numbers of examples causes some complications which we intentionally chose to avoid. First, increasing the dataset sizes causes a combinatorial explosion for the reporting of steering results, which currently involve steering from each concept/relation/concept-relation pair to every other concept/relation/concept-relation pair. Consequently, with a larger dataset, any steering results reported would have to be based on subsets of such combinations, e.g., specific interchange interventions as done in the RAVEL dataset \cite{wu-etal-2024-pyvene}. However, this has a tradeoff of providing a less clear view into the composition of the semantic modules identified in our work. 

This is also nontrivial due to the requirement of manually selecting components based on various heuristics outlined in Section~\ref{sec:results}. While we believe it may be possible to automate component selection in future work (as described below), this would still take a significant amount of time and computational power. Nevertheless, we believe the examples in our datasets comprise sufficiently representative samples of the domains, e.g., countries around the world, animals from various classes, orders, and families, words of various syntactic categories, and verbs with both common and irregular past tense transformations.

\paragraph{Limited selection of interpretability methods and models.}

Computational constraints also restricted our analysis to standard JumpReLU sparse autoencoders~\citep{DBLP:journals/corr/abs-2407-14435}, which can suffer from inconsistent feature quality, poor reconstruction, and weak functional alignment. These limitations have motivated various improved architectures~\citep{braun2024identifying,rajamanoharan2024improvingdictionarylearninggated,DBLP:journals/corr/abs-2407-14435,dunefsky2024transcoders}, which may yield even more refined functional module identification under our approach. As a proof-of-concept, we extended our translation analysis to Gemma 2 2B transcoders \citep{dunefsky2024transcoders}, successfully rediscovering some causal features for translation languages that \citeauthor{circuit-tracer} manually select, and achieving about 27\% steering success rate with our extracted components.\footnote{See Appendix~\ref{apx:transcoders} for more details.} This provides evidence that this approach can work with broader interpretability methods than SAEs.
% While  but were not able to complete a comprehensive study of transcoders. 

Related to this, most of our results were limited to one model (Gemma 2 2B), except for the supplementary results on the country facts task with Gemma 2 9B in Appendix~\ref{apx:9b-results}. Our choice of LLMs was constrained to those with pretrained SAEs available via Neuronpedia. Further, we did not use smaller models like GPT-2, as they demonstrated a weaker grasp of the factual concepts under investigation (e.g., confusing Lagos, Nigeria's largest city, with its capital) and an inability to follow in-context instructions to shorten their answers. As we continue to improve this work, we hope to expand the results to more models within the bounds of our computational constraints.

\paragraph{\revision{Reliance on manual intervention.}}

\revision{While our method of extracting components based on sparse feature coactivation is fully automated, as detailed in Section~\ref{sec:c identification}, we manually selected and/or combined components to represent each concept and relation using heuristics dependent on the task. This may give an impression that significant manual intervention is required, compromising the utility of our method. However, the focus of our work was primarily to automate the former step of component extraction, and given that, we found it relatively simple to identify semantically coherent and impactful components (the latter step). As we expanded the work to more tasks, we found that different choices maximized performance depending on the nature of the task, resulting in minor variations to our method.}

\revision{That said, after conducting this work, we believe these heuristics can be replaced with a simple algorithm in future work. First, as our approach already does, we can apply our sparse feature coactivation method to identify several components for each concept-relation prompt, and sort them by their KL divergence of the LLM's next token probability before and after ablation. Second, we can represent each unique concept and relation by identifying the most impactful component(s) which overlap among all prompts for that concept or relation, then taking the intersection of them. Another advantage of such an approach is that it removes dependence on feature descriptions, e.g., those we used from Neuronpedia.}

\paragraph{Focus on sparsely activating features.}

% precluding the use of more advanced architectures such as Gated SAEs~\citep{rajamanoharan2024improvingdictionarylearninggated} or transcoders~\citep{dunefsky2024transcoders} which may yield even more refined functional module identification.
Lastly, our analysis focused exclusively on features with low activation density.
Intriguingly, preliminary experiments revealed that ablating high-density features produces syntactically and semantically incoherent outputs, suggesting these features serve critical yet unexplored computational functions that merit further investigation.

\section*{Acknowledgments}
We thank our anonymous reviewers for their thoughtful and constructive feedback.
This research was supported in part through computational resources and services provided by Advanced Research Computing at the University of Michigan, Ann Arbor. Google Gemini 3 (\url{gemini.google.com}) and Anthropic Claude (\url{claude.ai}) were used for minor language edits and generation of boilerplate and visualization code.

\bibliography{references}

\clearpage
\appendix
\section{Gemma 2 9B Results}\label{apx:9b-results}

\begin{figure*}
    \centering

        \centering
        \includegraphics[width=0.8\textwidth]{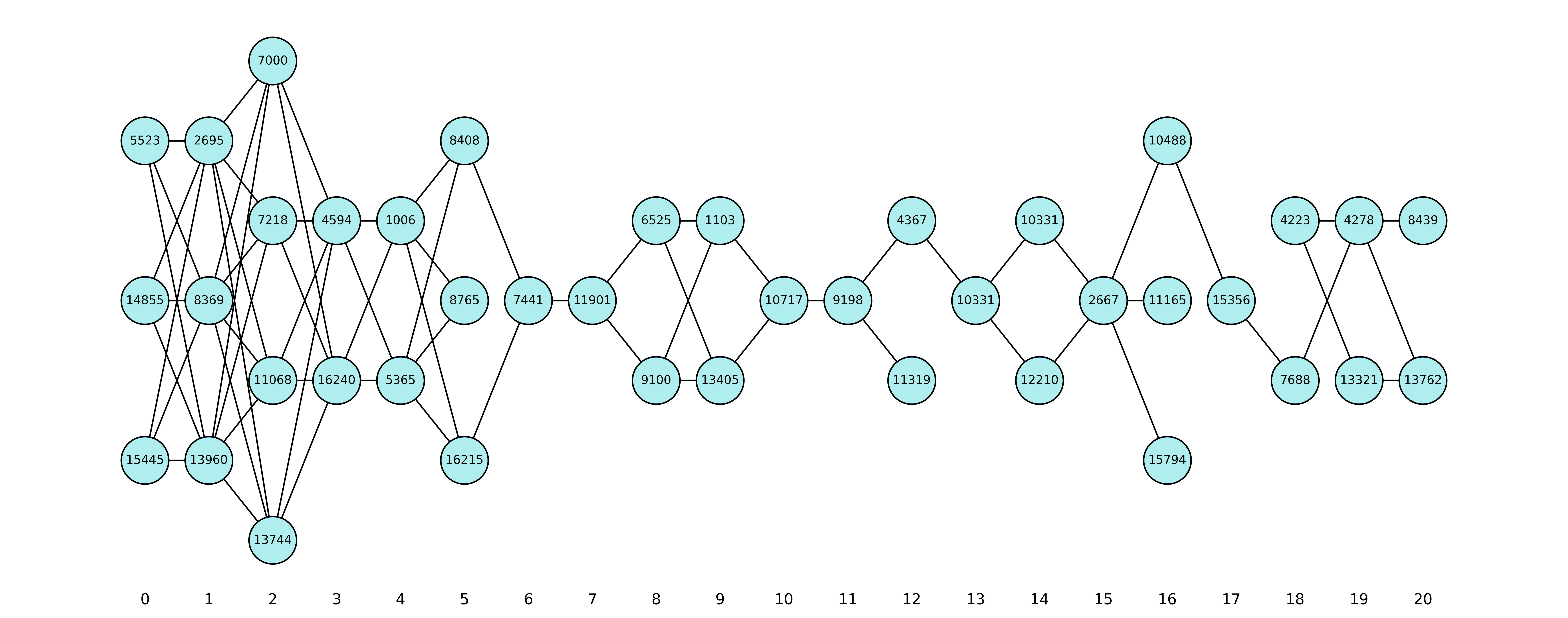}
    %\hspace{0.05cm}
        \centering
        \includegraphics[width=0.8\textwidth]{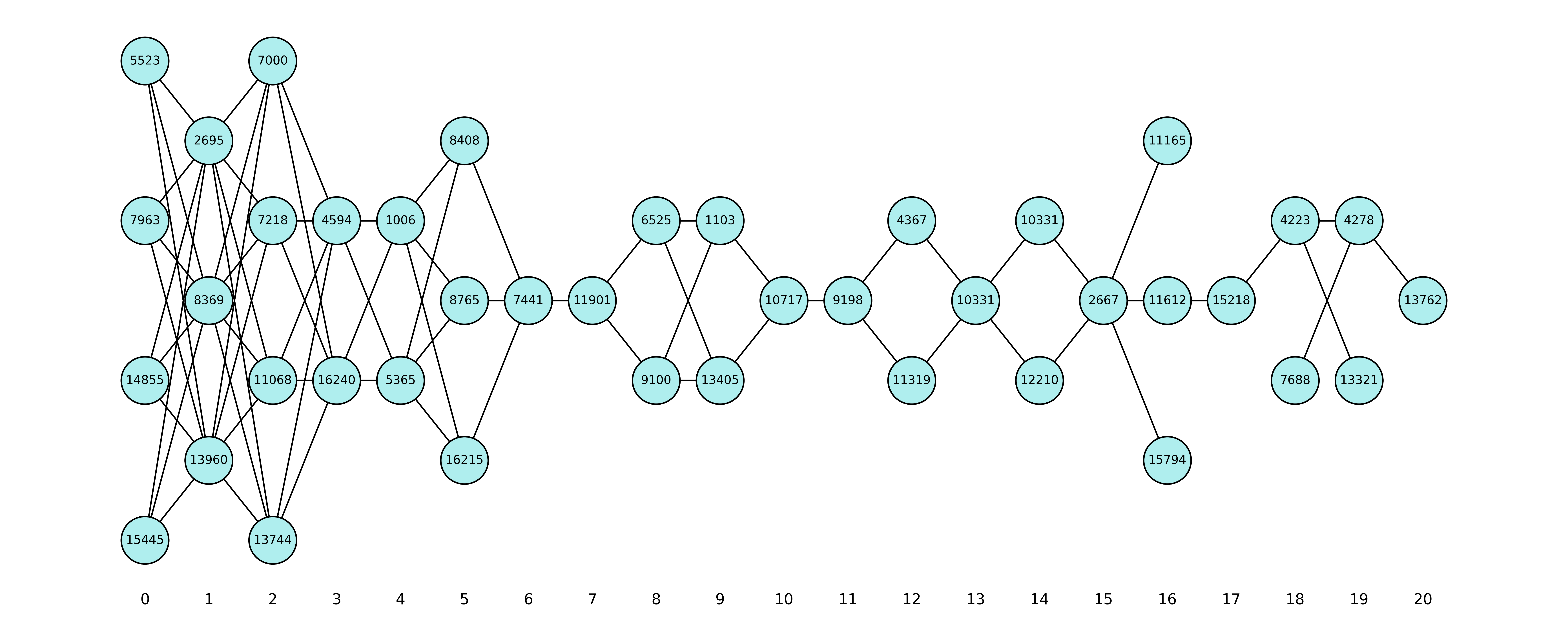}
    \caption{Gemma 2 9B China components extracted from capital and currency prompts.}
    \label{fig:china_component_capital_currency_9b}
\end{figure*}

\begin{figure*}
    \centering
    \begin{minipage}[b]{0.8\textwidth}
        \centering
        \includegraphics[width=\textwidth]{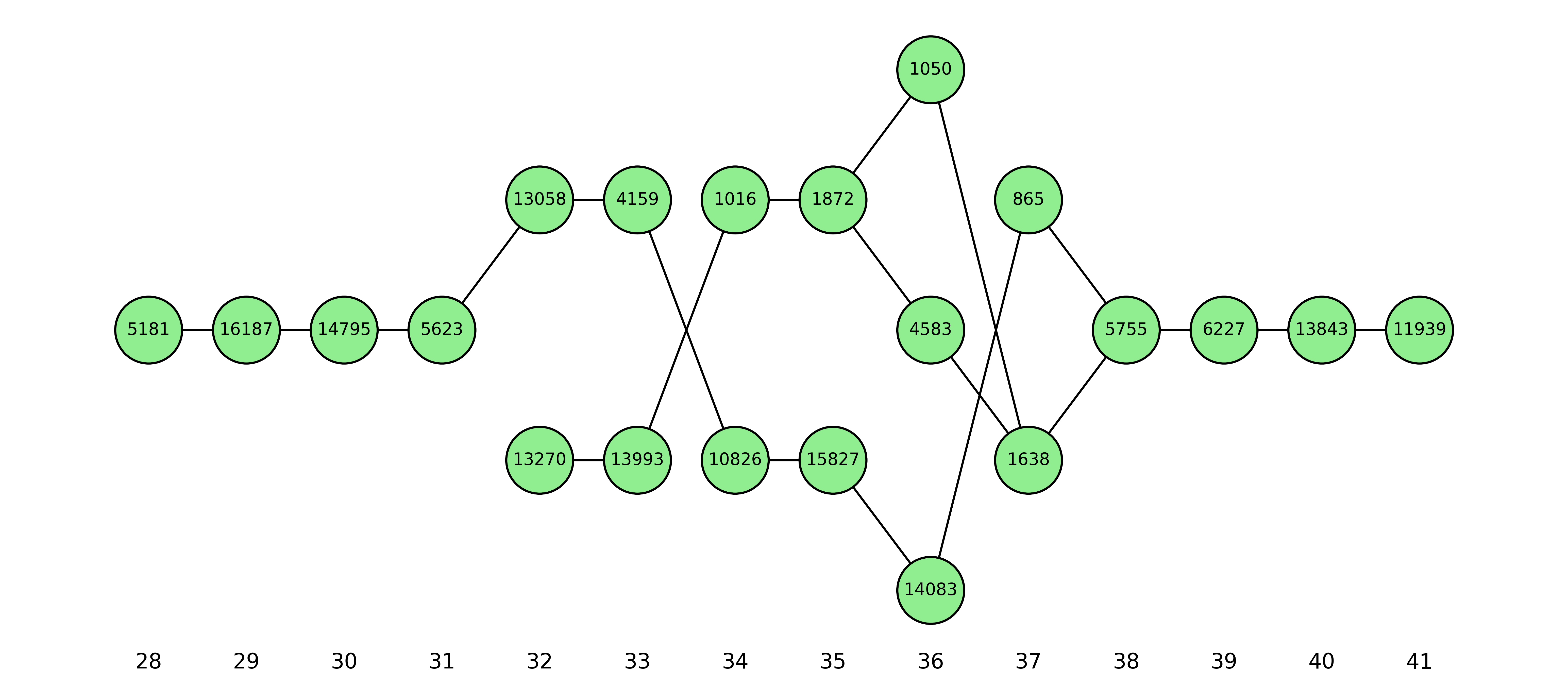}
    \end{minipage}
    %\hspace{0.05cm}
    \begin{minipage}[b]{0.8\textwidth}
        \centering
        \includegraphics[width=\textwidth]{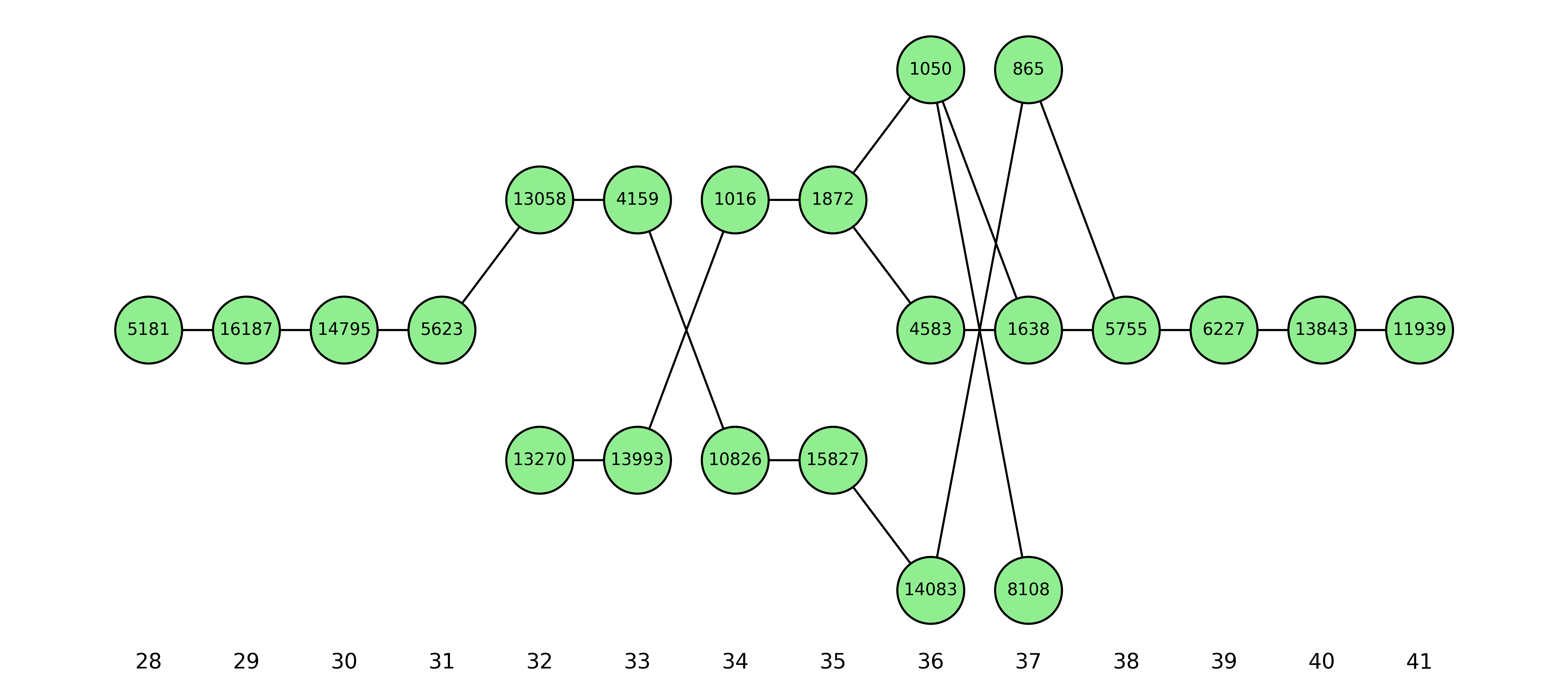}
    \end{minipage}
    \caption{Gemma 2 9B language components extracted from China and Nigeria prompts.}
    \label{fig:language_component_china_nigeria_9b}
\end{figure*}

We replicated country fact task ablation and steering using Gemma 2 9B.
Results closely mirrored those observed with Gemma 2 2B.
The model showed context consistency, with similar country components across relations (Figure~\ref{fig:china_component_capital_currency_9b}) and similar relation components across countries (Figure~\ref{fig:language_component_china_nigeria_9b}).
High success rates were achieved for country (93\%), relation (97\%), and composite (92\%) steering (see Tables~\ref{tab:country_steering_9b}–\ref{tab:relation_country_steering_9b}).

\begin{table*}
    \centering
    \setlength{\tabcolsep}{6.6pt}
    \small
    \begin{tabular}{rccccccccccc}
        \toprule
        \textit{Country} & \textbf{CN} & \textbf{FR} & \textbf{DE} & \textbf{JP} & \textbf{NG} & \textbf{PL} & \textbf{RU} & \textbf{ES} & \textbf{UK} & \textbf{US} & \textbf{Average} \\
        \cmidrule(lr){1-1}\cmidrule(lr){2-11}\cmidrule(lr){12-12}
        \textit{Success Rate} & 1.00 & 0.78 & 0.85 & 1.00 & 0.89 & 0.81 & 1.00 & 0.96 & 1.00 & 0.96 & 0.93 \\
        \bottomrule
    \end{tabular}
    \vspace{5pt}
    \caption{Gemma 2 9B steering success rates for each target country across all 27 prompts that do not already query that country.}
    \label{tab:country_steering_9b}
\end{table*}

\begin{table}
    \centering
    \setlength{\tabcolsep}{3pt}
    \small
    \begin{tabular}{rcccc}
        \toprule
        \textit{Relation} & \textbf{Capital} & \textbf{Currency} & \textbf{Language} & \textbf{Average} \\
        \cmidrule(lr){1-1}\cmidrule(lr){2-4}\cmidrule(lr){5-5}
        \textit{Success Rate} & 0.90 & 1.00 & 1.00 & 0.97 \\
        \bottomrule
    \end{tabular}
    \vspace{5pt}
    \caption{Gemma 2 9B steering success rates for each target country fact across all 20 prompts that do not already query that country fact.}
    \label{tab:relation_steering_9b}
\end{table}

\begin{table*}
    \centering
    \setlength{\tabcolsep}{7.5pt}    
    \small
    \begin{tabular}{rccccccccccc}
        \toprule
        \textit{Ctry. / Rel.} & \textbf{CN} & \textbf{FR} & \textbf{DE} & \textbf{JP} & \textbf{NG} & \textbf{PL} & \textbf{RU} & \textbf{ES} & \textbf{UK} & \textbf{US} & \textbf{Avg.} \\
        \cmidrule(lr){1-1}\cmidrule(lr){2-11}\cmidrule(lr){12-12}
        \textit{Capital} & 1.00 & 1.00 & 1.00 & 1.00 & 1.00 & 1.00 & 1.00 & 1.00 & 1.00 & 1.00 & 1.00 \\
        \textit{Currency} & 1.00 & 1.00 & 1.00 & 0.72 & 1.00 & 0.50 & 0.11 & 0.94 & 1.00 & 1.00 & 0.83 \\
        \textit{Language} & 0.50 & 1.00 & 1.00 & 1.00 & 1.00 & 0.78 & 1.00 & 1.00 & 1.00 & 1.00 & 0.93 \\
        \cmidrule(lr){1-1}\cmidrule(lr){2-11}\cmidrule(lr){12-12}
        \textit{Average} & 0.83 & 1.00 & 1.00 & 0.91 & 1.00 & 0.76 & 0.70 & 0.98 & 1.00 & 1.00 & 0.92 \\
        \bottomrule
    \end{tabular}
    \vspace{5pt}
    \caption{Gemma 2 9B composite steering success rates for each target country-fact pair across all 18 prompts that do not already query the target country or fact.}
    \label{tab:relation_country_steering_9b}
\end{table*}

\section{Full Task Details}\label{apx:evaluation schemes}

In this appendix, we provide the full list of concepts included in our analysis for each task. Additionally, as our accuracy and steering success evaluations importantly accounted for multiple possible correct answers to prompts, we provide lists of correct answers for each explored task's concept-relation pairs below. During evaluation, LLMs generate an answer of no more than 5 tokens. 
An LLM's answer is generally counted as correct if it contains any of the correct answers in the lists below, except for the following caveats.

In the first letter transformation and translation cases where correct answers are less than 4 characters long, the LLM must generate a correct answer on its own surrounded by whitespace or non-letters. This constraint reduces the possibility of mistakenly marking an LLM output that is not a correct answer but happens to contain one as correct, e.g., the string "te" appears in many Spanish and French words, but not all such words should be taken as correct answers for translating \textit{you}.
However, it is also common in translation for an LLM to output a token pertaining to a correct answer, then follow it up with tokens that morph its meaning (e.g., "Rotwein" when steering the LLM to output the German word for \textit{red}). To account for this, in translation, we always mark LLM outputs beginning with a correct answer as correct.
Evaluation is case-insensitive, except for the capitalization relation of the verb transformations task.

The below lists of correct answers were refined through inspecting correct answers counted as incorrect in intermediate results. Importantly, the off-the-shelf tested models achieve 100\% accuracy on all tasks under the few-shot prompts used to identify components, as well as the zero-shot prompts used for ablation and steering experiments. 
% The taxonomy task is the only exception, where Gemma 2 2B achieved poor zero-shot accuracy, and thus we used few-shot prompts for all experiments. 
% The model's inconsistent performance may be another reason for the relatively low steering success rates on this task.

\subsection{Country Facts}
For the country facts, the full list of countries and their corresponding facts is below:

\begin{itemize}
    \item Capital:
    \begin{itemize}
        \item \textit{China} $\rightarrow$ \textit{Beijing, Peking}
        \item \textit{France} $\rightarrow$ \textit{Paris}
        \item \textit{Germany} $\rightarrow$ \textit{Berlin}
        \item \textit{Japan} $\rightarrow$ \textit{Tokyo}
        \item \textit{Nigeria} $\rightarrow$ \textit{Abuja}
        \item \textit{Poland} $\rightarrow$ \textit{Warsaw}
        \item \textit{Russia} $\rightarrow$ \textit{Moscow, \foreignlanguage{russian}{Москва}}
        \item \textit{Spain} $\rightarrow$ \textit{Madrid}
        \item \textit{UK} $\rightarrow$ \textit{London}
        \item \textit{USA} $\rightarrow$ \textit{Washington}
    \end{itemize}

    \item Currency:
    \begin{itemize}
        \item \textit{China} $\rightarrow$ \textit{Yuan, ¥, RMB, CNY, renmin}
        \item \textit{France} $\rightarrow$ \textit{Euro, Franc, €}
        \item \textit{Germany} $\rightarrow$ \textit{Euro, €, Mark}
        \item \textit{Japan} $\rightarrow$ \textit{Yen, ¥}
        \item \textit{Nigeria} $\rightarrow$ \textit{Naira, ₦}
        \item \textit{Poland} $\rightarrow$ \textit{Złoty, Zloty, PLN, Zlo}
        \item \textit{Russia} $\rightarrow$ \textit{Ruble, \faRub, RUB, \foreignlanguage{russian}{руб}}
        \item \textit{Spain} $\rightarrow$ \textit{Euro, €}
        \item \textit{UK} $\rightarrow$ \textit{Pound, £}
        \item \textit{USA} $\rightarrow$ \textit{Dollar, \$}
    \end{itemize}

    \item Language:
    \begin{itemize}
        \item \textit{China} $\rightarrow$ \textit{Mandarin, Chinese}
        \item \textit{France} $\rightarrow$ \textit{French, Français}
        \item \textit{Germany} $\rightarrow$ \textit{German, Deutsch}
        \begin{CJK}{UTF8}{min}
            \item \textit{Japan} $\rightarrow$ \textit{Japanese, 日本語}
        \end{CJK}
        \item \textit{Nigeria} $\rightarrow$ \textit{English}
        \item \textit{Poland} $\rightarrow$ \textit{Polish}
        \item \textit{Russia} $\rightarrow$ \textit{Russian}
        \item \textit{Spain} $\rightarrow$ \textit{Spanish}
        \item \textit{UK} $\rightarrow$ \textit{English}
        \item \textit{USA} $\rightarrow$ \textit{English}
    \end{itemize}
\end{itemize}

\subsection{Word Translation}
For translation, our evaluation scheme accounts for various aspects:
\begin{itemize}
    \item \textbf{Parts of speech}, e.g., \textit{poisson} (fish) and \textit{p{\^e}cher} (to fish) as correct French translations of \textit{fish}
    \item \textbf{Verb conjugations}, e.g., \textit{amo} and \textit{amas} as correct Spanish translations of \textit{love} (despite this verb taking the same form in English regardless of grammatical first or second person)
    \item \textbf{Semantic relevance}, e.g., \textit{coureur} (runner) as a valid French translation of \textit{run} despite being syntactically incorrect
\end{itemize}

These choices allow us to separate syntactic errors from semantic errors (the latter of which we focus on). The full list of correct answers is below:

\begin{itemize}
    \item Spanish:
    \begin{itemize}
        \item \textit{beautiful} $\rightarrow$ \textit{hermosa, hermoso, bello, bella, bonito, bonita, lindo, linda, belleza, bellisima}
        \item \textit{cat} $\rightarrow$ \textit{gato, gata, gatos, gatas, felino}
        \item \textit{dog} $\rightarrow$ \textit{perro, perra, perros, perras}
        \item \textit{fish} $\rightarrow$ \textit{pez, pescar, pesco, pescas, pescan, pescado, pesca, faenar}
        \item \textit{good} $\rightarrow$ \textit{buen, bueno, buena, bien, buenos, buenas}
        \item \textit{here} $\rightarrow$ \textit{aquí, acá, aqui}
        \item \textit{learn} $\rightarrow$ \textit{aprender, aprendo, aprendes, aprenden, aprende, Saber}
        \item \textit{love} $\rightarrow$ \textit{amar, amor, amo, amas, aman, encanta}
        \item \textit{red} $\rightarrow$ \textit{rojo, roja}
        \item \textit{run} $\rightarrow$ \textit{correr, corro, corres, corren, corrida, corre, carrera, corredor}
        \item \textit{you} $\rightarrow$ \textit{tú, eres, usted, le, te, ustedes, vosotros}
    \end{itemize}

    \item French:
    \begin{itemize}
        \item \textit{beautiful} $\rightarrow$ \textit{belle, beau, bel, beauté}
        \item \textit{cat} $\rightarrow$ \textit{chat, chatte, chats, chattes, félin}
        \item \textit{dog} $\rightarrow$ \textit{chien, chienne, chiens, chiennes}
        \item \textit{fish} $\rightarrow$ \textit{poisson, pêcher, pêche}
        \item \textit{good} $\rightarrow$ \textit{bien, bon, grand}
        \item \textit{here} $\rightarrow$ \textit{ici}
        \item \textit{learn} $\rightarrow$ \textit{apprendre, apprends, apprenant, savoir}
        \item \textit{love} $\rightarrow$ \textit{amour, aimer, aime, aiment, aimes, aimez, aim}
        \item \textit{red} $\rightarrow$ \textit{rouge}
        \item \textit{run} $\rightarrow$ \textit{courir, cours, course, couler, coureur}
        \item \textit{you} $\rightarrow$ \textit{tu, toi, vous, on, te}
    \end{itemize}

    \item German:
    \begin{itemize}
        \item \textit{beautiful} $\rightarrow$ \textit{schön, schöne, schönen}
        \item \textit{cat} $\rightarrow$ \textit{Katze, Katzen}
        \item \textit{dog} $\rightarrow$ \textit{Hund, Hunde}
        \item \textit{fish} $\rightarrow$ \textit{Fisch}
        \item \textit{good} $\rightarrow$ \textit{gut, gute, guten}
        \item \textit{here} $\rightarrow$ \textit{hier}
        \item \textit{learn} $\rightarrow$ \textit{lernen, lerne, lernst, lernt, lern}
        \item \textit{love} $\rightarrow$ \textit{lieben, liebe, liebst, liebt, lieb}
        \item \textit{red} $\rightarrow$ \textit{rot, rote, roten}
        \item \textit{run} $\rightarrow$ \textit{laufen, laufe, läufst, läuft, lauf, rennen, renne, rennst, rennt, renn, Flucht}
        \item \textit{you} $\rightarrow$ \textit{du, Sie, duzen}
    \end{itemize}
\end{itemize}

\subsection{Verb Transformation}

For verb transformation, the full list of verbs and their corresponding transformations is below:

\begin{itemize}
    \item Synonym:
    \begin{itemize}
        \item \textit{break} $\rightarrow$ \textit{shatter, destroy, split, pause, end, fracture, stop}
        \item \textit{focus} $\rightarrow$ \textit{concentrate, center, attention, attend, sharpen}
        \item \textit{hide} $\rightarrow$ \textit{conceal, camouflage, bury, stash, secret}
        \item \textit{include} $\rightarrow$ \textit{incorporate, contain, comprise, cover, involve, add, insert, append, package}
        \item \textit{like} $\rightarrow$ \textit{enjoy, similar, alike, love, prefer}
        \item \textit{possess} $\rightarrow$ \textit{have, hold, own, control, dominate, influence, keep, succeed}
        \item \textit{sink} $\rightarrow$ \textit{submerge, dip, drop, descend, plunge, drain, basin, fall, destination}
        \item \textit{understand} $\rightarrow$ \textit{realize, realise, grasp, know, comprehend, get}
    \end{itemize}

    \item Antonym:
    \begin{itemize}
        \item \textit{break} $\rightarrow$ \textit{fix, mend, repair, continue, resume, make, whole, hold}
        \item \textit{focus} $\rightarrow$ \textit{distract, unfocus, ignore, blur, lose, disperse, diversion}
        \item \textit{hide} $\rightarrow$ \textit{expose, reveal, flaunt, show}
        \item \textit{include} $\rightarrow$ \textit{exclude, omit, remove, delete}
        \item \textit{like} $\rightarrow$ \textit{dislike, hate, avoid, unlike}
        \item \textit{possess} $\rightarrow$ \textit{dispossess, lose, want, need, lack, abandon, miss, disarray, fail, not have}
        \item \textit{sink} $\rightarrow$ \textit{float, rise, ascend, stand}
        \item \textit{understand} $\rightarrow$ \textit{misunderstand, misinterpret, ignore, not understand}
    \end{itemize}

    \item Past Tense:
    \begin{itemize}
        \item \textit{break} $\rightarrow$ \textit{broke, broken}
        \item \textit{focus} $\rightarrow$ \textit{focused}
        \item \textit{hide} $\rightarrow$ \textit{hid, hidden}
        \item \textit{include} $\rightarrow$ \textit{included}
        \item \textit{like} $\rightarrow$ \textit{liked}
        \item \textit{possess} $\rightarrow$ \textit{possessed}
        \item \textit{sink} $\rightarrow$ \textit{sank, sunk}
        \item \textit{understand} $\rightarrow$ \textit{understood}
    \end{itemize}

    \item Capitalize:
    \begin{itemize}
        \item \textit{break} $\rightarrow$ \textit{Break}
        \item \textit{focus} $\rightarrow$ \textit{Focus}
        \item \textit{hide} $\rightarrow$ \textit{Hide}
        \item \textit{include} $\rightarrow$ \textit{Include}
        \item \textit{like} $\rightarrow$ \textit{Like}
        \item \textit{possess} $\rightarrow$ \textit{Possess}
        \item \textit{sink} $\rightarrow$ \textit{Sink}
        \item \textit{understand} $\rightarrow$ \textit{Understand}
    \end{itemize}

    \item First Letter:
    \begin{itemize}
        \item \textit{break} $\rightarrow$ \textit{B}
        \item \textit{focus} $\rightarrow$ \textit{F}
        \item \textit{hide} $\rightarrow$ \textit{H}
        \item \textit{include} $\rightarrow$ \textit{I}
        \item \textit{like} $\rightarrow$ \textit{L}
        \item \textit{possess} $\rightarrow$ \textit{P}
        \item \textit{sink} $\rightarrow$ \textit{S}
        \item \textit{understand} $\rightarrow$ \textit{U}
    \end{itemize}
\end{itemize}

Like translation, the synonym and antonym relations similarly account for a variety of interpretations of the input word, as well as some answers which have a different part of speech than the input word.

\section{Prompt Templates}\label{apx:prompt templates}

In this appendix, we list component selection and steering evaluation prompt templates for all tasks' relations.

\subsection{Country Facts}

\paragraph{Capital city.}

For component selection, we use:

\begin{quote}
    \textit{The capital city of Peru is Lima. The capital city of South Korea is Seoul. The capital city of Saudi Arabia is Riyadh. The capital city of \texttt{\{in-prompt country\}} is}
\end{quote}

For steering evaluation, we use:

\begin{quote}
    \textit{Q: What is the capital city of \texttt{\{in-prompt country\}}? Answer directly (two words max). \\    
    A:}
\end{quote}

\paragraph{Currency.}

For component selection, we use:

\begin{quote}
    \textit{The currency of Peru is the Sol. The currency of South Korea is the Won. The currency of Saudi Arabia is the Riyal. The currency of \texttt{\{in-prompt country\}} is the}
\end{quote}

For steering evaluation, we use:

\begin{quote}
    \textit{Q: What is the currency of \texttt{\{in-prompt country\}}? Answer directly (two words max). \\    
    A:}
\end{quote}

\paragraph{Language.}

For component selection, we use:

\begin{quote}
    \textit{The main language in Peru is Spanish. The main language in South Korea is Korean. The main language in Saudi Arabia is Arabic. The main language in \texttt{\{in-prompt country\}} is}
\end{quote}

For steering evaluation, we use:

\begin{quote}
    \textit{Q: What is the main language in \texttt{\{in-prompt country\}}? Answer directly (two words max). \\    
    A:}
\end{quote}

\subsection{Word Translation}

\paragraph{Spanish.}

For component selection, we use:

\begin{quote}
    \textit{The Spanish word for “sing" is cantar. The Spanish word for "he" is él. The Spanish word for "bird" is pájaro. The Spanish word for "\texttt{\{in-prompt word\}}" is}
\end{quote}

For steering evaluation, we use:

\begin{quote}
    \textit{Q: What is the Spanish word for "\texttt{\{in-prompt word\}}"? Answer directly (two words max).\\
    A:}
\end{quote}

\paragraph{French.}

For component selection, we use:

\begin{quote}
    \textit{The French word for “sing" is chanter. The French word for "he" is il. The French word for "bird" is oiseau. The French word for "\texttt{\{in-prompt word\}}" is}
\end{quote}

For steering evaluation, we use:

\begin{quote}
    \textit{Q: What is the French word for "\texttt{\{in-prompt word\}}"? Answer directly (two words max).\\
    A:}
\end{quote}

\paragraph{German.}

For component selection, we use:

\begin{quote}
    \textit{The German word for “sing" is singen. The German word for "he" is er. The German word for "bird" is Vogel. The German word for "\texttt{\{in-prompt word\}}" is}
\end{quote}

For steering evaluation, we use:

\begin{quote}
    \textit{Q: What is the German word for "\texttt{\{in-prompt word\}}"? Answer directly (two words max).\\
    A:}
\end{quote}

\subsection{Verb Transformation}

\paragraph{Synonym.}

For component selection, we use:

\begin{quote}
    \textit{A synonym of throw is toss. A synonym of go is move. A synonym of find is discover. A synonym of \texttt{\{in-prompt verb\}} is}
\end{quote}

For steering evaluation, we use:

\begin{quote}
    \textit{Q: What is a synonym of \texttt{\{in-prompt verb\}}? Answer directly (one word max).\\
    A:}
\end{quote}

\paragraph{Antonym.}

For component selection, we use:

\begin{quote}
    \textit{An antonym of throw is catch. An antonym of go is stop. An antonym of find is lose. An antonym of \texttt{\{in-prompt verb\}} is}
\end{quote}

For steering evaluation, we use:

\begin{quote}
    \textit{Q: What is an antonym of \texttt{\{in-prompt verb\}}? Answer directly (one word max).\\
    A:}
\end{quote}

\paragraph{Past tense.}

For component selection, we use:

\begin{quote}
    \textit{The past tense of throw is threw. The past tense of go is went. The past tense of find is found. The past tense of \texttt{\{in-prompt verb\}} is}
\end{quote}

For steering evaluation, we use:

\begin{quote}
    \textit{Q: What is the past tense of \texttt{\{in-prompt verb\}}? Answer directly (one word max).\\
    A:}
\end{quote}

\paragraph{Capitalize.}

For component selection, we use:

\begin{quote}
    \textit{The word throw with the first letter capitalized is Throw. The word go with the first letter capitalized is Go. The word find with the first letter capitalized is Find. The word \texttt{\{in-prompt verb\}} with the first letter capitalized is}
\end{quote}

For steering evaluation, we use:

\begin{quote}
    \textit{Q: Capitalize the word \texttt{\{in-prompt verb\}}. Answer directly (one word max).\\
    A:}
\end{quote}

\paragraph{First letter.}

For component selection, we use:

\begin{quote}
    \textit{The first letter of throw is T. The first letter of go is G. The first letter of find is F. The first letter of \texttt{\{in-prompt verb\}} is}
\end{quote}

For steering evaluation, we use:

\begin{quote}
    \textit{Q: What is the first letter of the word \texttt{\{in-prompt verb\}}? Answer directly (one word max).\\
    A:}
\end{quote}

\section{Additional Component KL Divergence Plots}\label{apx:component_importance}

\begin{figure*}
    \centering
    \includegraphics[width=0.94\linewidth]{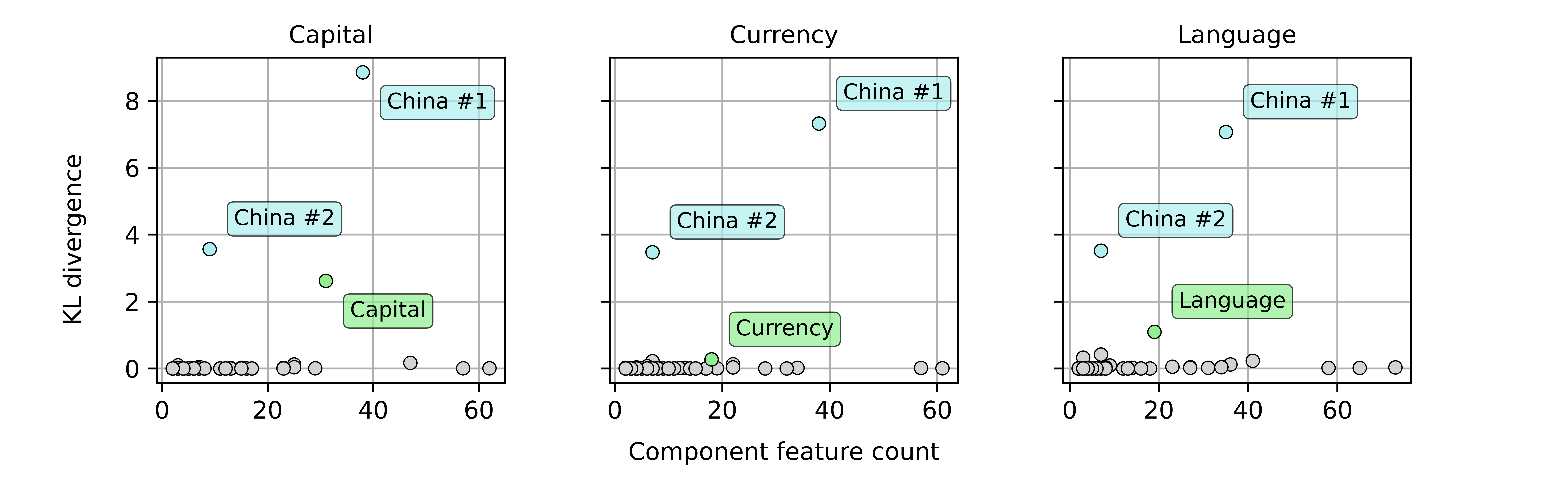}

    \vspace{-10pt}
    
    \caption{Plot of component feature counts versus KL divergence between pre- and post-ablation output token distributions for sparse components extracted for the capital city, currency, and language of China.} %\jycc{add somewhere X axis is... Y is .... }}
    \label{fig:component_importance countries} 
\end{figure*}

\begin{figure}
    \centering
    \includegraphics[width=0.98\linewidth]{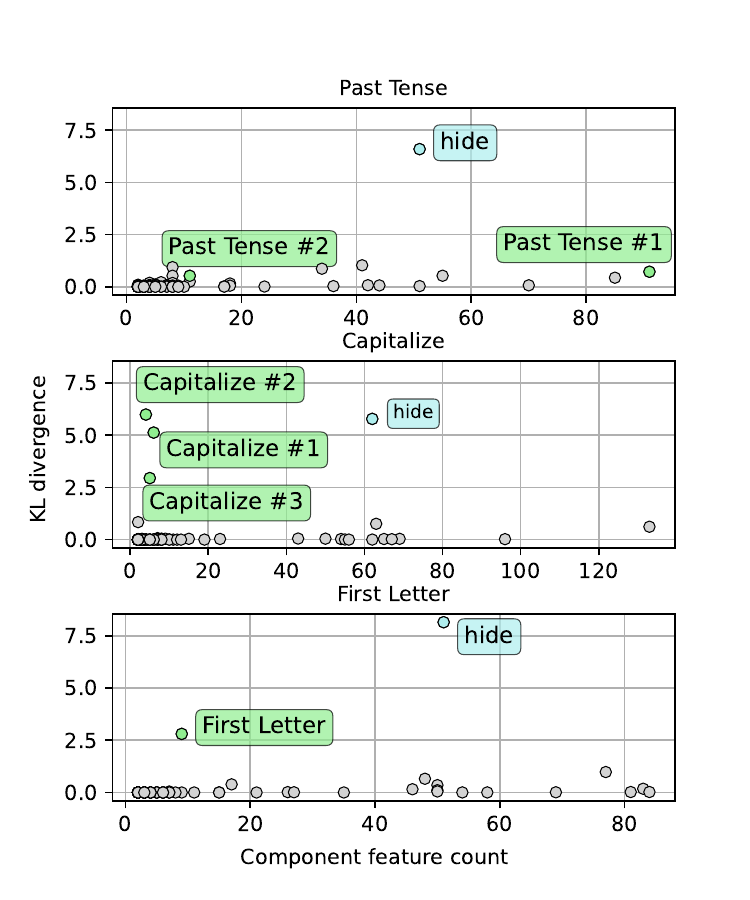}
    \caption{Plot of component feature counts versus KL divergence between pre- and post-ablation output token distributions for sparse components extracted for the capitalization, first letter, and past tense of \textit{hide}.}
    \label{fig:component_importance verbs}
\end{figure}

In Figure~\ref{fig:component_importance}, we plotted all sparse components for \textit{love} and the three languages in the word translation task. We provide some additional examples in this appendix. Specifically, Figure~\ref{fig:component_importance countries} visualizes the component KL divergences for China and the capital city, currency, and language relations in the country facts task. Further, Figure~\ref{fig:component_importance verbs} visualizes the component KL divergences for \textit{hide} and the capitalize, first letter, and past tense relations in the verb transformation task.

\section{Country Facts Component Feature Description Visualizations}
\label{subsec:interpretation_viz}

\begin{figure}[htbp]
    \centering
    % Top row with two figures side by side
    \begin{minipage}[b]{0.4\textwidth}
        \centering
        \setlength{\fboxrule}{0.4pt}
        \setlength{\fboxsep}{0pt}
        \fbox{\includegraphics[width=\textwidth]{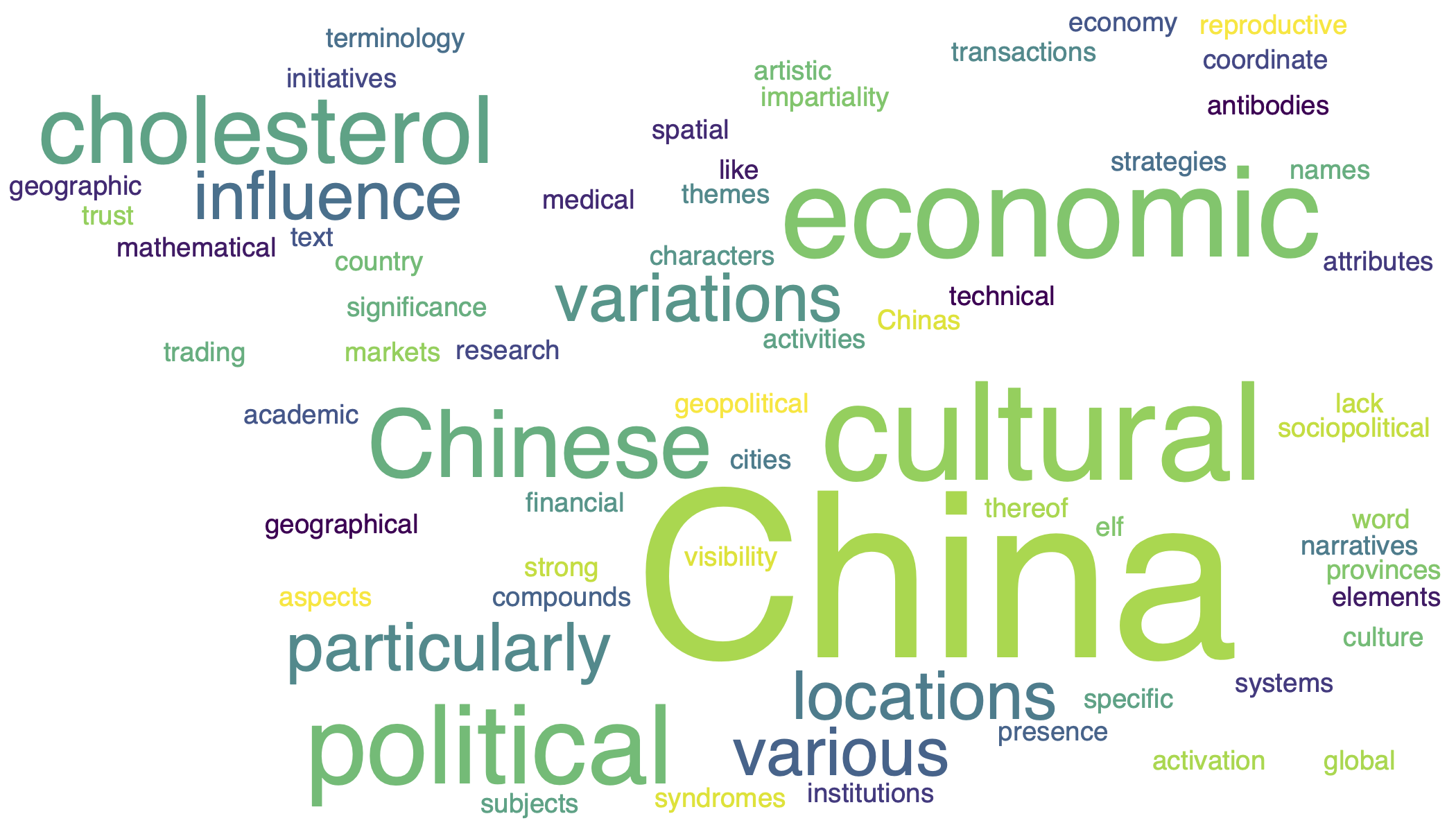}}
        \textbf{China Component}
    \end{minipage}
    \hspace{0.1\textwidth}
    \begin{minipage}[b]{0.4\textwidth}
        \centering
        \setlength{\fboxrule}{0.4pt}
        \setlength{\fboxsep}{0pt}
        \fbox{\includegraphics[width=\textwidth]{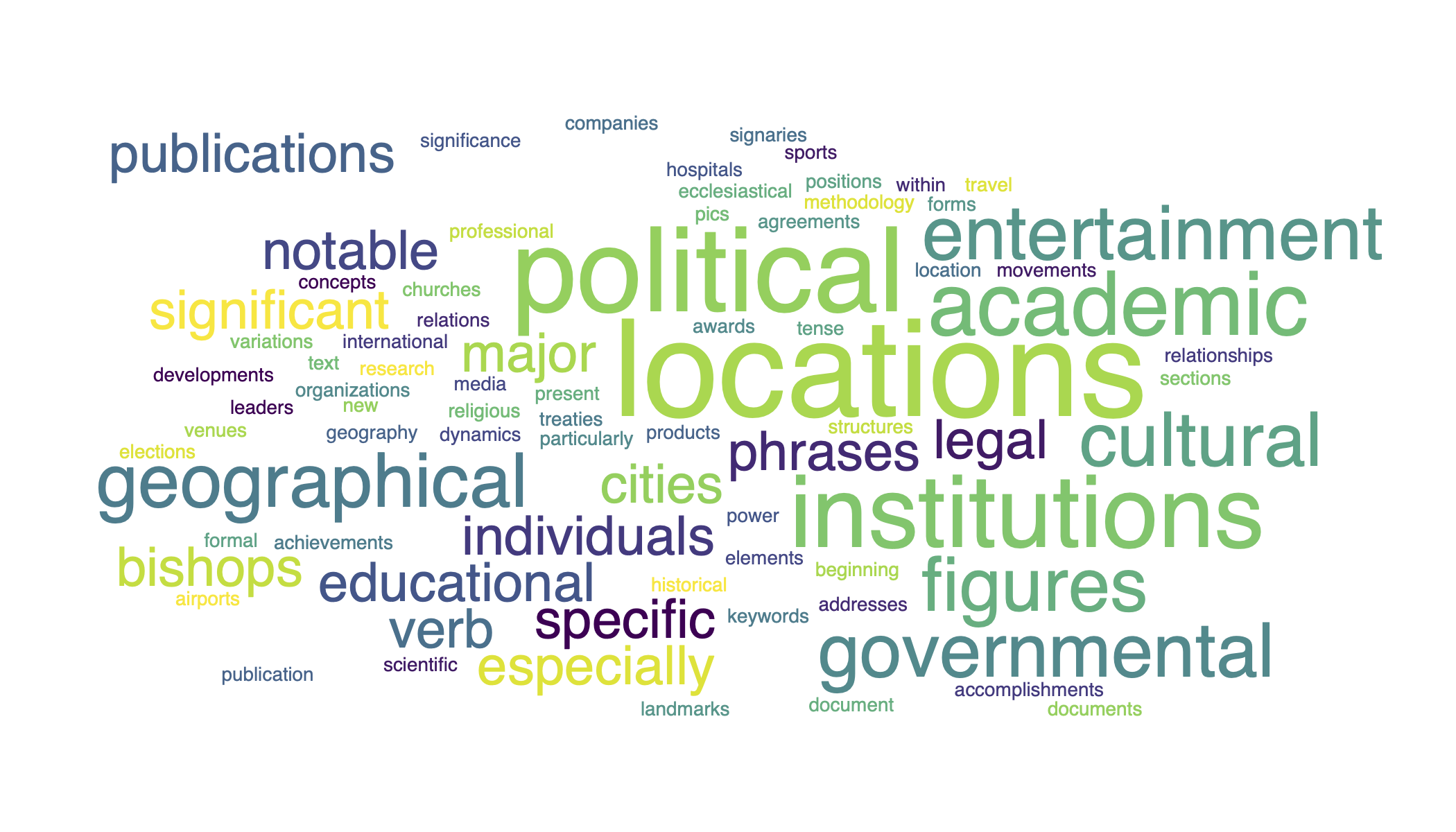}}
        \textbf{Capital Component}
    \end{minipage}
    
    \vspace{1em}
    
    % Bottom row with two figures side by side
    \begin{minipage}[b]{0.4\textwidth}
        \centering
        \setlength{\fboxrule}{0.4pt}
        \setlength{\fboxsep}{0pt}
        \fbox{\includegraphics[width=\textwidth]{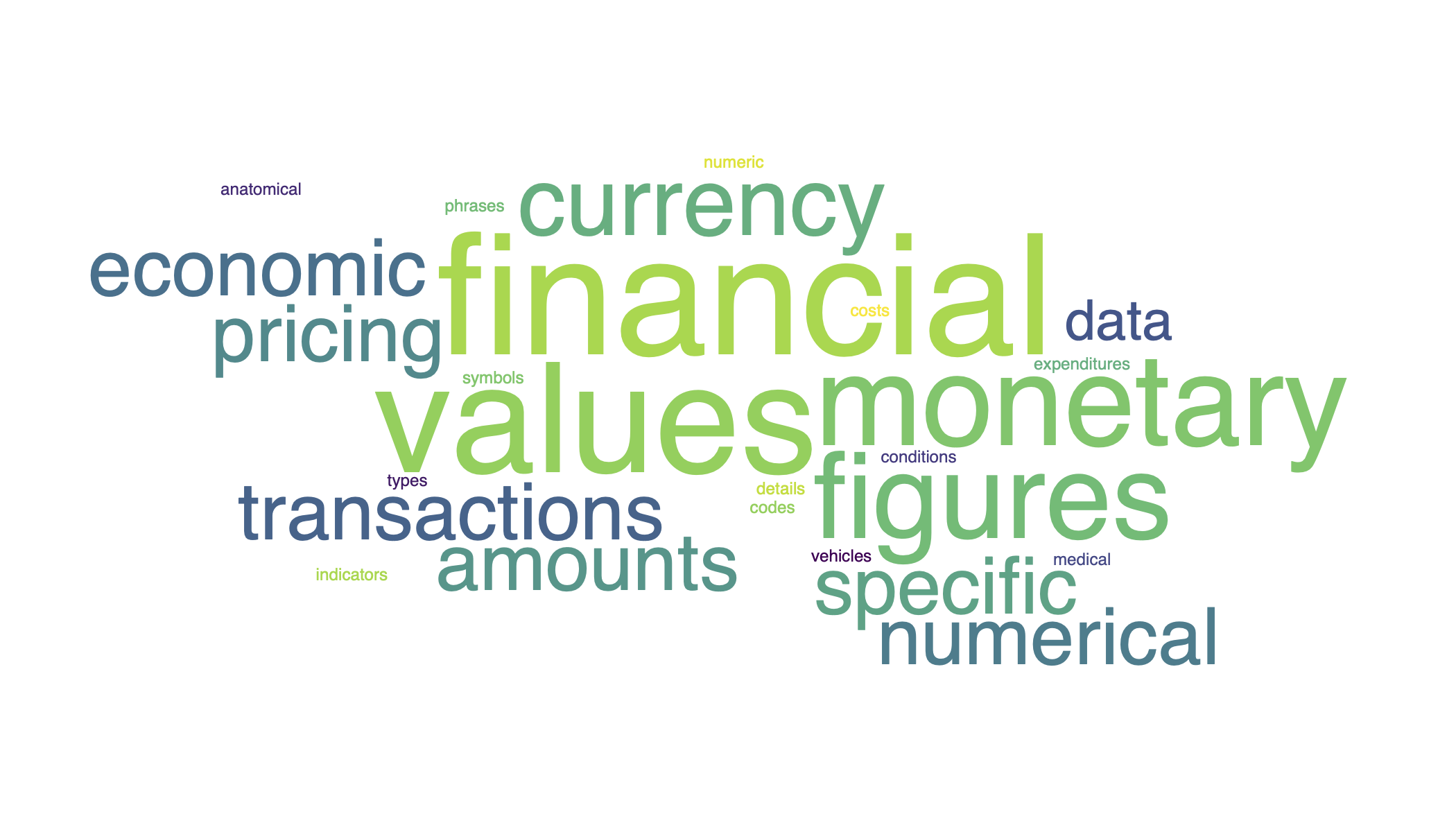}}
        \textbf{Currency Component}
    \end{minipage}
    \hspace{0.1\textwidth}
    \begin{minipage}[b]{0.4\textwidth}
        \centering
        \setlength{\fboxrule}{0.4pt}
        \setlength{\fboxsep}{0pt}
        \fbox{\includegraphics[width=\textwidth]{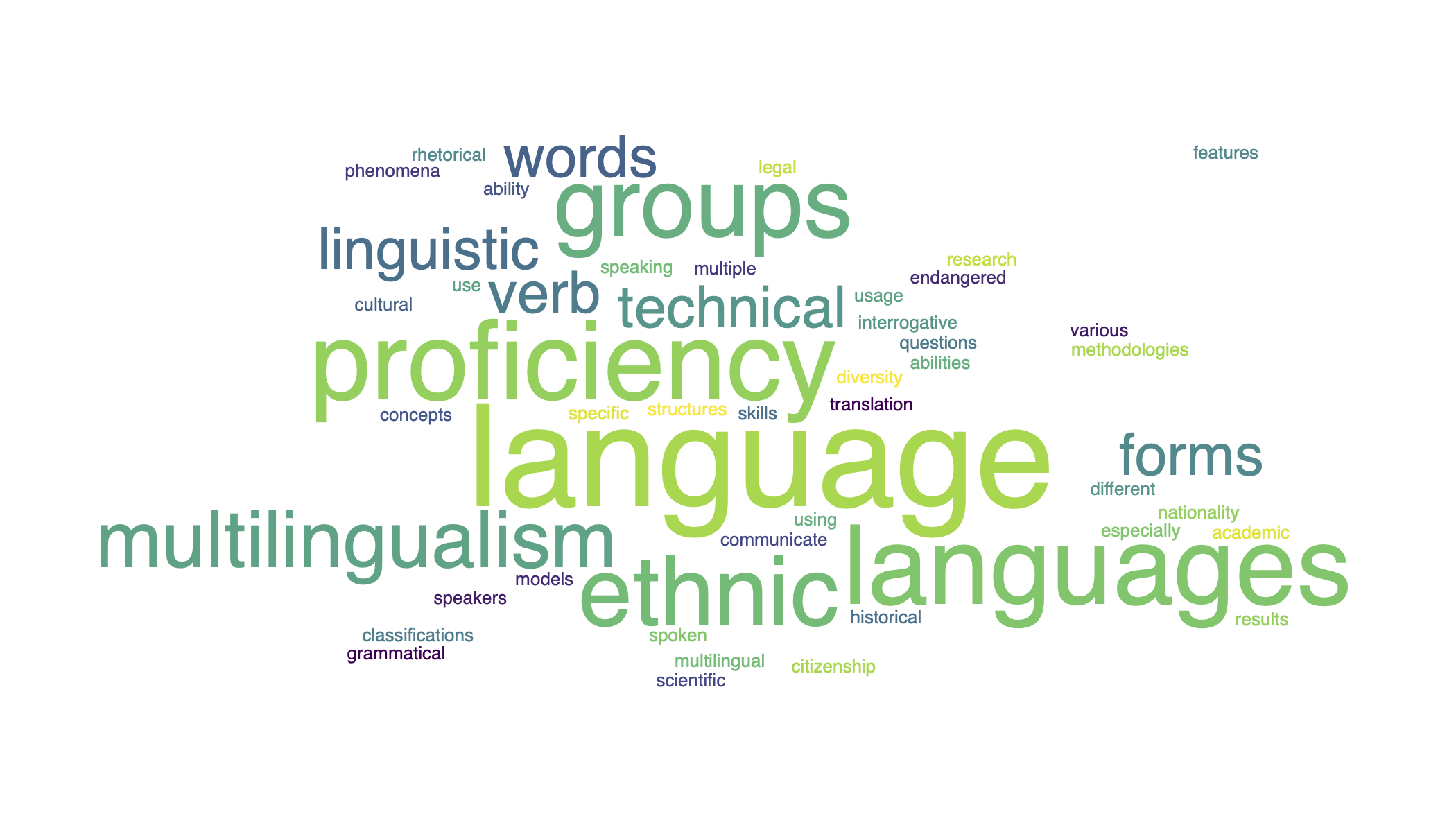}}

        \textbf{Language Component}
    \end{minipage}
    \caption{Word clouds for LLM-generated descriptions of SAE features within the China, capital, currency, and language components.}
    \label{fig:component_word_clouds}
\end{figure}

Figure~\ref{fig:component_word_clouds} visualizes the feature descriptions for selected components in the country fact task.

%See Figure~\ref{fig:component_word_clouds}.

\section{Additional Concept and Relation Component Visualizations}\label{apx:more components}

\begin{figure*}

\centering

    \begin{multicols}{3}

    \includegraphics[width=0.32\textwidth]{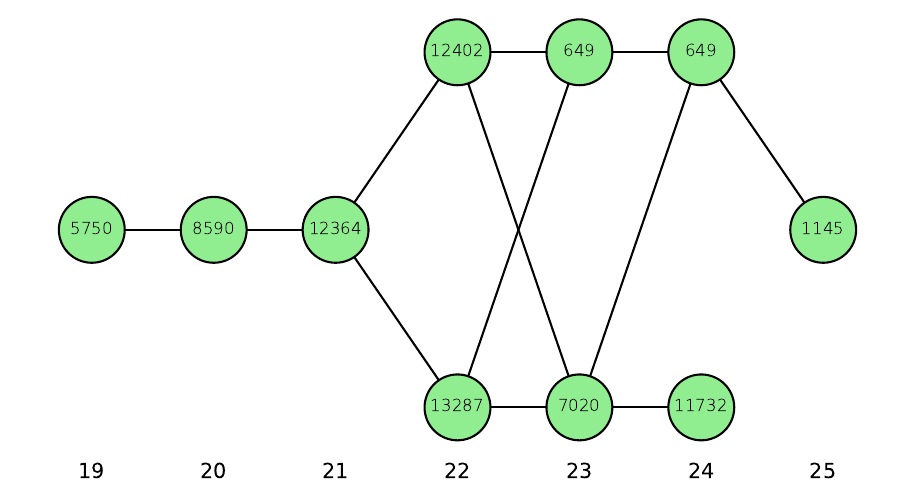}

    \includegraphics[width=0.32\textwidth]{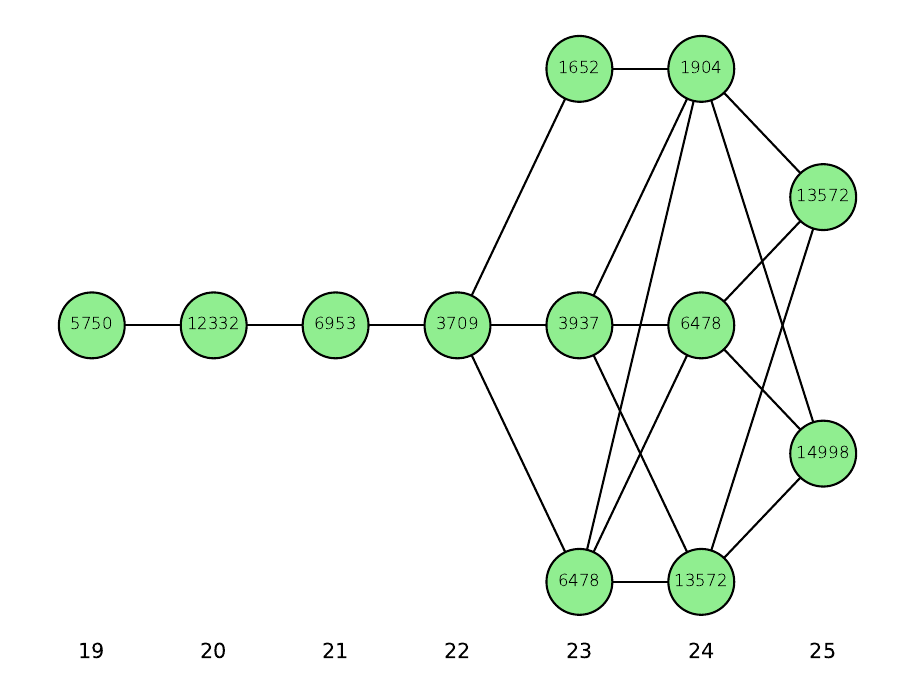}    

    \includegraphics[width=0.32\textwidth]{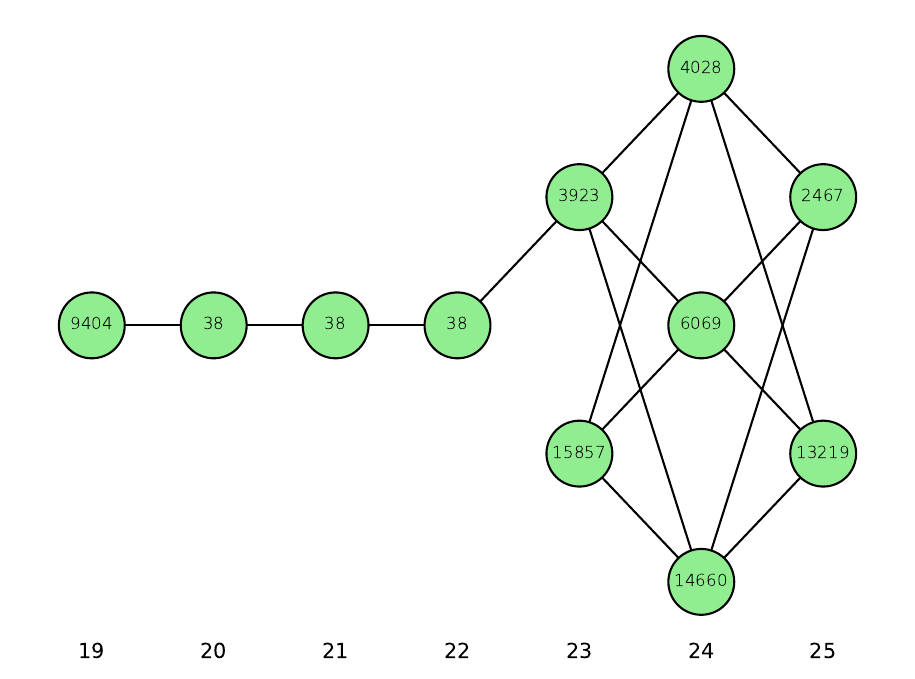}    

    \end{multicols}
    
    \caption{Components representing Spanish, French, and German translation (from left to right).}
    \label{fig:translation lang components}
\end{figure*}

\begin{figure*}

\centering

    \includegraphics[width=0.8\textwidth]{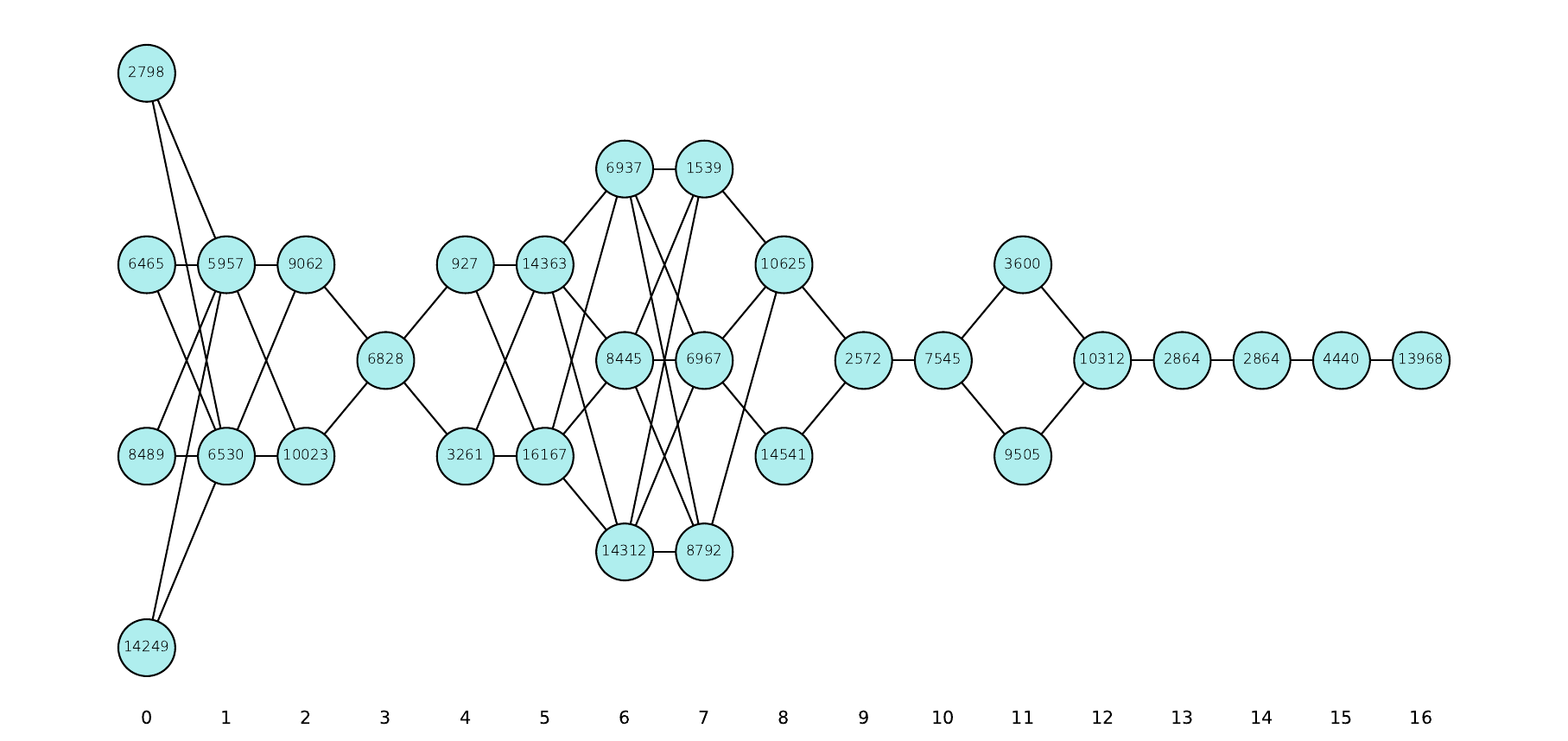}
    \includegraphics[width=0.8\textwidth]{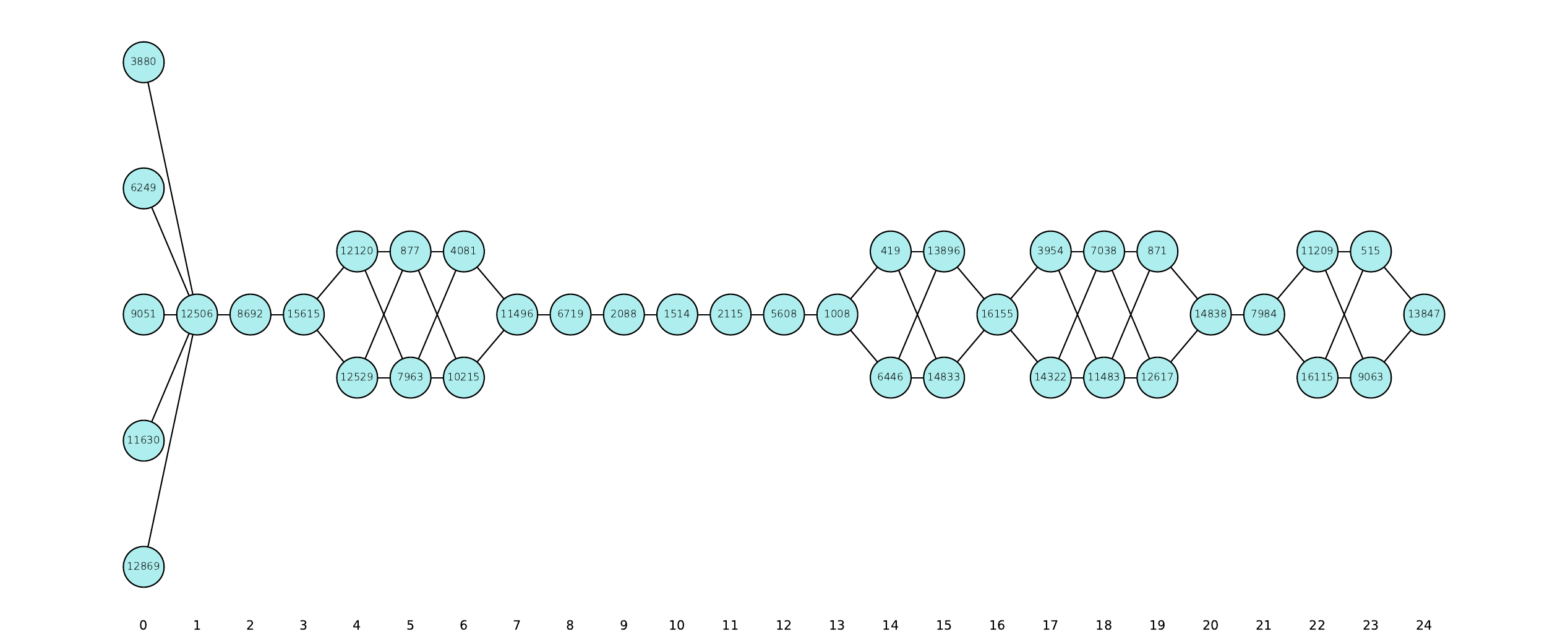}
    \includegraphics[width=0.8\textwidth]{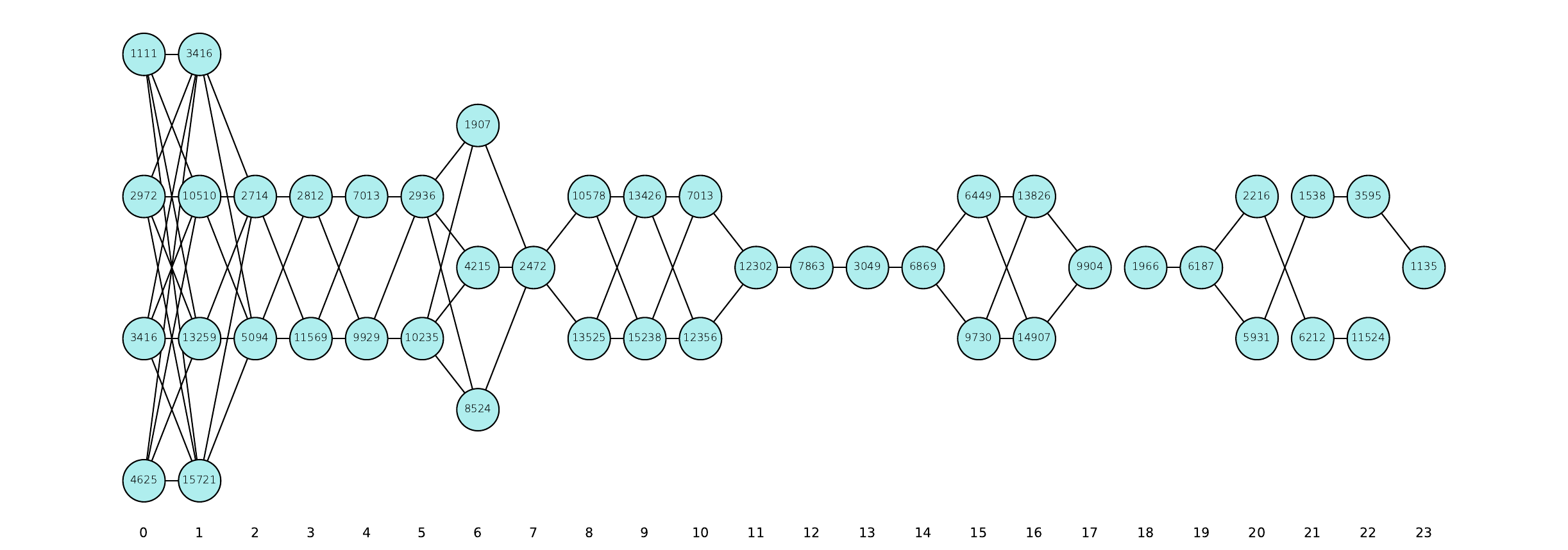}

    \caption{Components representing the translated words \textit{beautiful}, \textit{dog}, and \textit{love} (from left to right).}
    \label{fig:translation word components}
\end{figure*}

\begin{figure*}

\centering

    \begin{multicols}{2}
        \includegraphics[width=0.4\textwidth]{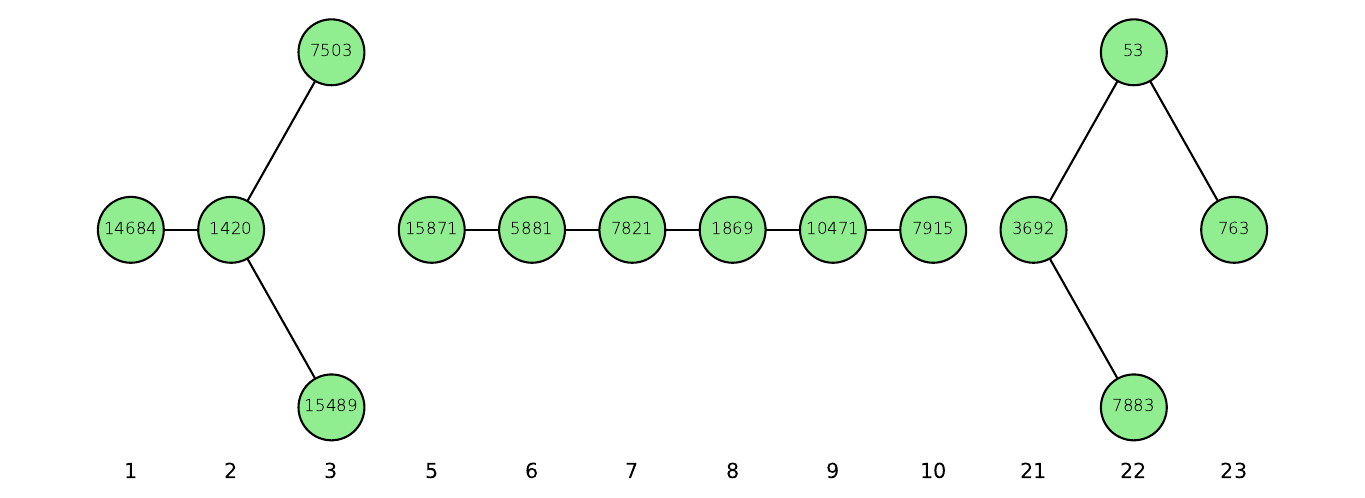}

        \includegraphics[width=0.4\textwidth]{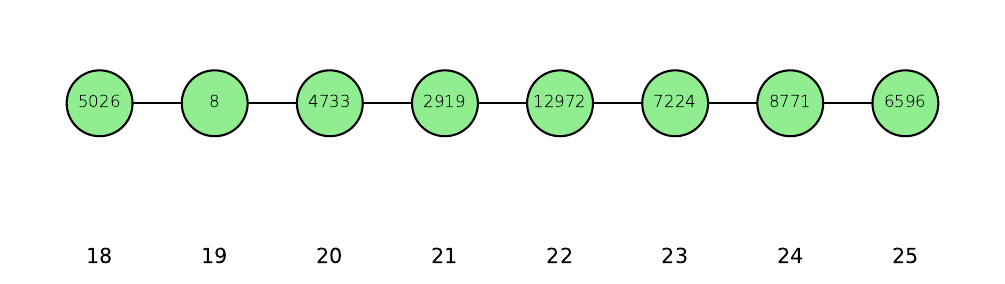}
    \end{multicols}

    \includegraphics[width=0.8\textwidth]{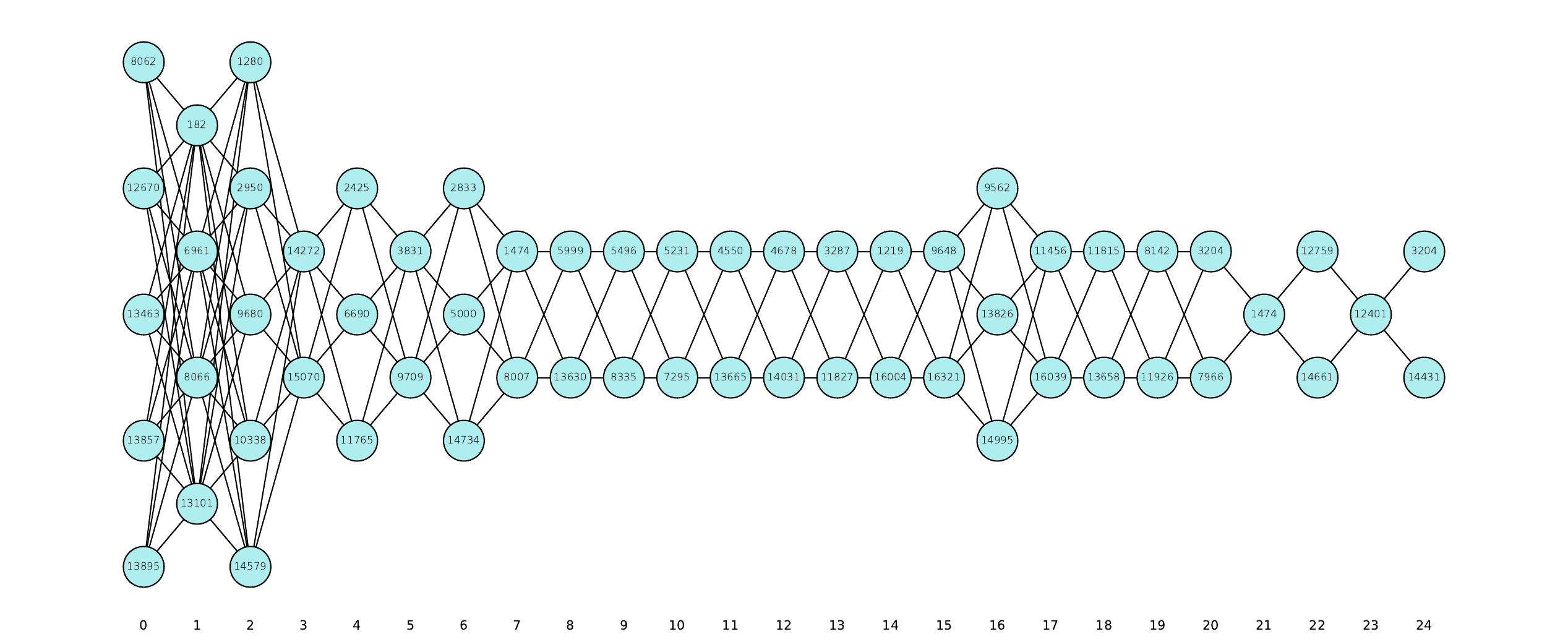}

    \includegraphics[width=0.8\textwidth]{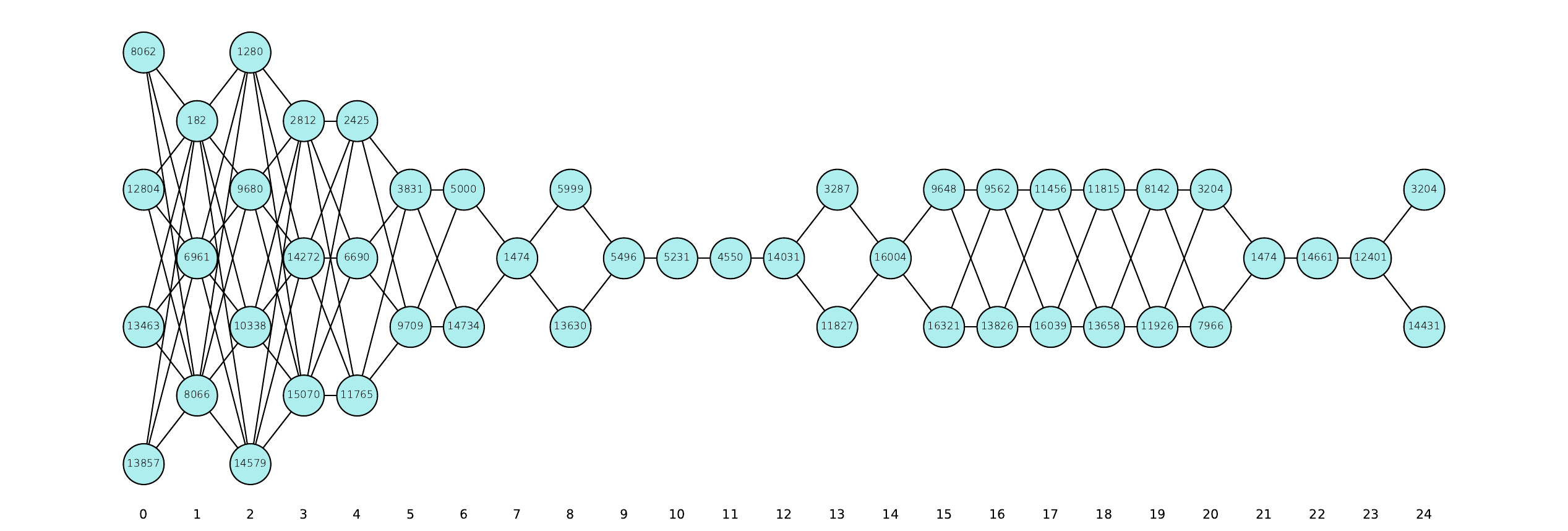}

    \caption{Components representing the capitalize (upper left) and first letter (upper right) transformations in the context of the verb \textit{like}, and the components representing the verb \textit{like} in the  context of the capitalize (middle row) and first letter (bottom row) transformations.}
    \label{fig:verbs components}
\end{figure*}

In this appendix, we include some additional visualizations of components extracted from studied tasks. Figures~\ref{fig:translation lang components} and \ref{fig:translation word components} visualize selected components from the word translation task, while Figure~\ref{fig:verbs components} visualizes selected components from the verb transformation task.

\section{Steering Details and Hyperparameters}\label{apx:steering strength}

For individual concept and relation steering, we selected steering strengths $\alpha_c, \alpha_r$ from $\{k \cdot 0.05 : k \in \mathbb{Z}\} \cap (0,1]$ that achieved the highest respective success rates. For composite steering, we selected the $(\alpha_c', \alpha_r')$ pair from $\{\alpha_c-0.05, \alpha_c, \alpha_c+0.05\} \times \{\alpha_r-0.05, \alpha_r, \alpha_r+0.05\}$ that achieved the highest success rate. This procedure yielded the following parameters for each task:

\begin{itemize}
    \item Country facts: $\alpha_c=0.1$, $\alpha_r=0.4$, $\alpha'_c=0.15$, and $\alpha'_r=0.45$
    \item Word translation: $\alpha_c=0.05$, $\alpha_r=0.4$, $\alpha'_c=0.05$, and $\alpha'_r=0.35$
    \item Verb transformation: $\alpha_c=0.05$, $\alpha_r=0.65$, $\alpha'_c=0.05$, and $\alpha'_r=0.7$
\end{itemize}

% To ensure consistency of viable steering values, in cases where features belonged to both an ablated component and steered component, we did not ablate them. This was particularly important in the animal taxonomy task, where components were overlapping and nested within each other.

For the verb transformation task, we additionally tried tuning a separate steering strength for each (target) relation. For individual concept steering, this made no difference in performance. For relation steering, past tense achieved up to 3.1\%, capitalization 50.0\%, and first letter 87.5\%. However, since antonym and synonym remained at 0\% steering success rate, choosing a single $\alpha_r$ in facilitating composite steering was not possible. As such, we chose to omit these results to avoid unnecessarily complicating our experimental paradigm. However, for future use of this method, it may be advantageous to explore more fine-grained selection of steering strengths for heterogeneous tasks like this one.

To provide more information about how performance varies with respect to steering strength, Tables~\ref{tab: steering strength search individual concept}, \ref{tab: steering strength search individual relation}, and \ref{tab: steering strength search composite concept-relation} list the individual concept/relation and composite steering accuracies across all searched steering strengths and tasks.

\begin{table}[h]
\centering
\footnotesize
\begin{tabular}{cccc}
\toprule
\textbf{\thead{Steering \\ Strength}} & \textbf{\thead{Country \\ Facts}} & \textbf{\thead{Word \\ Translation}} & \textbf{\thead{Verb \\ Transformation}} \\ \cmidrule(lr){1-1}\cmidrule(lr){2-4}
0.05 & {0.87} & \textbf{0.75} & \textbf{0.48} \\
0.10 & \textbf{0.96} & 0.51 & 0.30 \\
0.15 & 0.84 & 0.18 & 0.11 \\
0.20 & 0.54 & 0.09 & 0.06 \\
0.25 & 0.22 & 0.03 & 0.06 \\
0.30 & 0.17 & 0.00 & 0.10 \\
0.35 & 0.15 & 0.00 & 0.06 \\
0.40 & 0.15 & 0.00 & 0.08 \\
0.45 & 0.16 & 0.00 & 0.08 \\
0.50 & 0.16 & 0.00 & 0.08 \\
0.55 & 0.16 & 0.00 & 0.08 \\
0.60 & 0.15 & 0.00 & 0.07 \\
0.65 & 0.15 & 0.00 & 0.08 \\
0.70 & 0.16 & 0.00 & 0.08 \\
0.75 & 0.17 & 0.00 & 0.10 \\
0.80 & 0.20 & 0.00 & 0.09 \\
0.85 & 0.20 & 0.03 & 0.06 \\
0.90 & 0.20 & 0.03 & 0.06 \\
0.95 & 0.19 & 0.03 & 0.06 \\
1.00 & 0.18 & 0.03 & 0.06 \\
\bottomrule
\end{tabular}
\normalsize
\caption{Overall task accuracy for individual concept steering across candidate steering strengths. Accuracy under selected steering strength for each task in bold.}
\label{tab: steering strength search individual concept}
\end{table}

\begin{table}[h]
\centering
\footnotesize
\begin{tabular}{cccc}
\toprule
\textbf{\thead{Steering \\ Strength}} & \textbf{\thead{Country \\ Facts}} & \textbf{\thead{Word \\ Translation}} & \textbf{\thead{Verb \\ Transformation}} \\ \cmidrule(lr){1-1}\cmidrule(lr){2-4}
0.05 & 0.17 & 0.03 & 0.10 \\
0.10 & 0.60 & 0.35 & 0.09 \\
0.15 & 0.68 & 0.73 & 0.11 \\
0.20 & 0.67 & 0.94 & 0.11 \\
0.25 & 0.70 & 0.98 & 0.11 \\
0.30 & 0.77 & 0.98 & 0.10 \\
0.35 & 0.85 & 0.98 & 0.14 \\
0.40 & \textbf{0.93} & \textbf{0.98} & 0.16 \\
0.45 & 0.92 & 0.91 & 0.19 \\
0.50 & 0.92 & 0.82 & 0.19 \\
0.55 & 0.88 & 0.65 & 0.21 \\
0.60 & 0.83 & 0.48 & 0.22 \\
0.65 & 0.78 & 0.35 & \textbf{0.22} \\
0.70 & 0.78 & 0.20 & 0.21 \\
0.75 & 0.77 & 0.17 & 0.19 \\
0.80 & 0.75 & 0.14 & 0.19 \\
0.85 & 0.73 & 0.09 & 0.19 \\
0.90 & 0.70 & 0.09 & 0.17 \\
0.95 & 0.67 & 0.06 & 0.17 \\
1.00 & 0.63 & 0.02 & 0.17 \\
\bottomrule
\end{tabular}
\normalsize
\caption{Overall task accuracy for individual relation steering across candidate steering strengths. Accuracy under selected steering strength for each task in bold.}
\label{tab: steering strength search individual relation}
\end{table}

\begin{table}[h]
\centering
\footnotesize
\setlength{\tabcolsep}{3.5pt}
\begin{tabular}{cccc}
\toprule
\textbf{\thead{Steering \\ Strengths}} & \textbf{\thead{Country \\ Facts}} & \textbf{\thead{Word \\ Translation}} & \textbf{\thead{Verb \\ Transformation}} \\ \cmidrule(lr){1-1}\cmidrule(lr){2-4}
0.00, 0.35 & -- & 0.01 & -- \\
0.00, 0.40 & -- & 0.01 & -- \\
0.00, 0.45 & -- & 0.01 & -- \\
0.00, 0.60 & -- & -- & 0.15 \\
0.00, 0.65 & -- & -- & 0.16 \\
0.00, 0.70 & -- & -- & 0.16 \\
0.05, 0.35 & 0.74 & \textbf{0.64} & -- \\
0.05, 0.40 & 0.78 & 0.59 & -- \\
0.05, 0.45 & 0.77 & 0.48 & -- \\
0.05, 0.60 & -- & -- & 0.17 \\
0.05, 0.65 & -- & -- & 0.18 \\
0.05, 0.70 & -- & -- & \textbf{0.19} \\
0.10, 0.35 & 0.85 & 0.53 & -- \\
0.10, 0.40 & 0.85 & 0.47 & -- \\
0.10, 0.45 & 0.86 & 0.36 & -- \\
0.10, 0.60 & -- & -- & 0.16 \\
0.10, 0.65 & -- & -- & 0.18 \\
0.10, 0.70 & -- & -- & 0.18 \\
0.15, 0.35 & 0.86 & -- & -- \\
0.15, 0.40 & 0.88 & -- & -- \\
0.15, 0.45 & \textbf{0.90} & -- & -- \\
\bottomrule
\end{tabular}
\normalsize
\caption{Overall task accuracy for composite concept-relation steering across candidate steering strength pairs. Accuracy under selected steering strength pair for each task in bold.}
\label{tab: steering strength search composite concept-relation}
\end{table}

\section{Supplementary Ablation and Steering Next Token Distributions}\label{apx: additional examples}

In this appendix, we provide additional detailed examples of how the next token distributions change under ablating and steering concept and relation components.

\subsection{Individual Concept and Relation Ablation}\label{apx: additional examples ablation}

Extending Table~\ref{tab:ablation_top_tokens}, we provide three comprehensive sets of examples for ablating individual concepts and relations: Table~\ref{tab:ablation_top_tokens countries} (for country facts), Table~\ref{tab:ablation_top_tokens translation} (for word translation), and Table~\ref{tab:ablation_top_tokens verbs} (for verb transformations).
While the results largely mirror those highlighted in Table~\ref{tab:ablation_top_tokens}, we observe that when ablating the first letter transformation, the model still ranks tokens representing single letters highly, and many of those letters are other letters besides the first in the prompted verb (e.g., ``\textit{K}'', ``\textit{E}'', and ``\textit{A}'' in the context of \textit{break}). This suggests that the first letter component specifically captures first letters, and not other letters in prompted words; as such, ablating it still leaves a signal to the model to extract (non-first) letters. Further, it is possible that letter information is captured in a more general abstraction within the model which is activated by the first letter relation, but is not specific to it.

\begin{table*}
    \centering
    \setlength{\tabcolsep}{2.5pt}
    \small
    \begin{tabular}{llllllllllllllllll}
        \toprule
        \multicolumn{6}{c}{\textbf{Capital, $\frac{\text{China}}{\text{Nigeria}}$}} & \multicolumn{6}{c}{\textbf{Currency, $\frac{\text{China}}{\text{Nigeria}}$}} & \multicolumn{6}{c}{\textbf{Language, $\frac{\text{China}}{\text{Nigeria}}$}} \\
        \cmidrule[0.6pt](lr){1-6} \cmidrule[0.6pt](lr){7-12} \cmidrule[0.6pt](lr){13-18}
        \multicolumn{2}{c}{\textit{Original}} & \multicolumn{2}{c}{\textit{Ctry. Abl. }} & \multicolumn{2}{c}{\textit{Fact Abl.}} & \multicolumn{2}{c}{\textit{Original}} & \multicolumn{2}{c}{\textit{Ctry. Abl. }} & \multicolumn{2}{c}{\textit{Fact Abl.}} & \multicolumn{2}{c}{\textit{Original}} & \multicolumn{2}{c}{\textit{Ctry. Abl. }} & \multicolumn{2}{c}{\textit{Fact Abl.}} \\
        \cmidrule(lr){1-2}\cmidrule(lr){3-4}\cmidrule(lr){5-6}\cmidrule(lr){7-8}\cmidrule(lr){9-10}\cmidrule(lr){11-12}\cmidrule(lr){13-14}\cmidrule(lr){15-16}\cmidrule(lr){17-18}
        Beijing & \scriptsize{97.} & Madrid & \scriptsize{39.} & the & \scriptsize{12.} & Yuan & \scriptsize{80.} & Euro & \scriptsize{64.} & Yuan & \scriptsize{59.} & Mandarin & \scriptsize{59.} & Spanish & \scriptsize{49.} & Chinese & \scriptsize{24.} \\
        Be & \scriptsize{.38} & Warsaw & \scriptsize{10.} & China & \scriptsize{6.7} & Ren & \scriptsize{14.} & Lira & \scriptsize{20.} & Ren & \scriptsize{14.} & Chinese & \scriptsize{37.} & English & \scriptsize{22.} & China & \scriptsize{18.} \\
        Peking & \scriptsize{.35} & Rome & \scriptsize{9.5} & Beijing & \scriptsize{6.4} & RMB & \scriptsize{1.8} & Krone & \scriptsize{3.1} & Yen & \scriptsize{1.9} & English & \scriptsize{.69} & French & \scriptsize{6.7} & Mandarin & \scriptsize{11.} \\
        Shanghai & \scriptsize{.23} & Paris & \scriptsize{6.1} & Shanghai & \scriptsize{5.0} & Yen & \scriptsize{.98} & Franc & \scriptsize{2.4} & P & \scriptsize{1.7} & also & \scriptsize{.43} & German & \scriptsize{4.0} & also & \scriptsize{3.5} \\
        Xi & \scriptsize{.11} & Berlin & \scriptsize{5.0} & a & \scriptsize{2.7} & yuan & \scriptsize{.53} & Peso & \scriptsize{1.2} & Ba & \scriptsize{1.5} & Put & \scriptsize{.22} & Italian & \scriptsize{2.7} & a & \scriptsize{2.9} \\
        \cmidrule(lr){1-2}\cmidrule(lr){3-4}\cmidrule(lr){5-6}\cmidrule(lr){7-8}\cmidrule(lr){9-10}\cmidrule(lr){11-12}\cmidrule(lr){13-14}\cmidrule(lr){15-16}\cmidrule(lr){17-18}
        Abuja & \scriptsize{85.} & New & \scriptsize{7.7} & Lagos & \scriptsize{17.} & Naira & \scriptsize{93.} & Franc & \scriptsize{16.} & Naira & \scriptsize{62.} & English & \scriptsize{72.} & English & \scriptsize{50.} & Nigeria & \scriptsize{40.} \\
        Lagos & \scriptsize{11.} & Islamabad & \scriptsize{7.3} & the & \scriptsize{12.} & N & \scriptsize{2.9} & Euro & \scriptsize{9.1} & Dollar & \scriptsize{6.0} & Ha & \scriptsize{11.} & French & \scriptsize{31.} & English & \scriptsize{14.} \\
        ... & \scriptsize{.38} & Kathmandu & \scriptsize{6.0} & Nigeria & \scriptsize{7.1} & Nai & \scriptsize{2.5} & D & \scriptsize{7.4} & Niger & \scriptsize{4.3} & Yoruba & \scriptsize{4.7} & Spanish & \scriptsize{2.1} & Nigerian & \scriptsize{9.8} \\
        Nigeria & \scriptsize{.29} & Delhi & \scriptsize{5.7} & called & \scriptsize{5.1} & naira & \scriptsize{.36} & Pound & \scriptsize{5.1} & Currency & \scriptsize{2.4} & Igbo & \scriptsize{2.4} & Arabic & \scriptsize{2.0} & ... & \scriptsize{2.9} \\
        … & \scriptsize{.27} & Tehran & \scriptsize{4.9} & Abuja & \scriptsize{3.4} & K & \scriptsize{.23} & Krone & \scriptsize{4.8} & Nai & \scriptsize{1.4} & Nigerian & \scriptsize{1.6} & Dutch & \scriptsize{1.1} & also & \scriptsize{2.3} \\
        \bottomrule
    \end{tabular}
    \vspace{5pt}
    \caption{
    Top 5 output tokens and their likelihoods for China (top) and Nigeria (bottom) across all three relations, before and after ablating the relevant country or relation component in Gemma 2 2B. Underscore prefixes of tokens are omitted for space.}
    \label{tab:ablation_top_tokens countries}
\end{table*}

\begin{table*}
    \centering
    \setlength{\tabcolsep}{5pt}
    \small
    \begin{tabular}{llllllllllllllllll}
        \toprule
        \multicolumn{6}{c}{\textbf{Spanish, $\frac{\text{love}}{\text{red}}$}} & \multicolumn{6}{c}{\textbf{French, $\frac{\text{love}}{\text{red}}$}} & \multicolumn{6}{c}{\textbf{German, $\frac{\text{love}}{\text{red}}$}} \\
        \cmidrule[0.6pt](lr){1-6} \cmidrule[0.6pt](lr){7-12} \cmidrule[0.6pt](lr){13-18}
        \multicolumn{2}{c}{\textit{Original}} & \multicolumn{2}{c}{\textit{Word Abl. }} & \multicolumn{2}{c}{\textit{Lang. Abl.}} & \multicolumn{2}{c}{\textit{Original}} & \multicolumn{2}{c}{\textit{Word Abl. }} & \multicolumn{2}{c}{\textit{Lang. Abl.}} & \multicolumn{2}{c}{\textit{Original}} & \multicolumn{2}{c}{\textit{Word Abl. }} & \multicolumn{2}{c}{\textit{Lang. Abl.}} \\
        \cmidrule(lr){1-2}\cmidrule(lr){3-4}\cmidrule(lr){5-6}\cmidrule(lr){7-8}\cmidrule(lr){9-10}\cmidrule(lr){11-12}\cmidrule(lr){13-14}\cmidrule(lr){15-16}\cmidrule(lr){17-18}
        {Amor} & \scriptsize{41.} & {"} & \scriptsize{5.9} & {Love} & \scriptsize{42.} & {Amour} & \scriptsize{35.} & {Le} & \scriptsize{15.} & {Love} & \scriptsize{42.} & {Liebe} & \scriptsize{66.} & {Du} & \scriptsize{9.0} & {love} & \scriptsize{16.} \\
        {amor} & \scriptsize{19.} & {El} & \scriptsize{4.9} & {love} & \scriptsize{14.} & {amour} & \scriptsize{26.} & {La} & \scriptsize{6.4} & {I} & \scriptsize{11.} & {Lie} & \scriptsize{4.8} & {"} & \scriptsize{6.6} & {Liebe} & \scriptsize{13.} \\
        {Amo} & \scriptsize{7.0} & {La} & \scriptsize{4.4} & {I} & \scriptsize{13.} & {A} & \scriptsize{5.7} & {"} & \scriptsize{6.1} & {love} & \scriptsize{10.} & {Ich} & \scriptsize{4.4} & {Die} & \scriptsize{4.4} & {Love} & \scriptsize{9.6} \\
        {amo} & \scriptsize{5.1} & {Malo} & \scriptsize{3.2} & {"} & \scriptsize{5.3} & {J} & \scriptsize{3.8} & {le} & \scriptsize{3.7} & {"} & \scriptsize{5.8} & {"} & \scriptsize{3.4} & {das} & \scriptsize{4.1} & {I} & \scriptsize{8.7} \\
        {"} & \scriptsize{2.9} & {la} & \scriptsize{2.5} & {LOVE} & \scriptsize{2.7} & {"} & \scriptsize{2.7} & {la} & \scriptsize{3.3} & {LOVE} & \scriptsize{2.5} & {Herz} & \scriptsize{1.7} & {Der} & \scriptsize{3.9} & {"} & \scriptsize{6.8} \\\cmidrule(lr){1-2}\cmidrule(lr){3-4}\cmidrule(lr){5-6}\cmidrule(lr){7-8}\cmidrule(lr){9-10}\cmidrule(lr){11-12}\cmidrule(lr){13-14}\cmidrule(lr){15-16}\cmidrule(lr){17-18}
        {rojo} & \scriptsize{38.} & {La} & \scriptsize{3.5} & {Red} & \scriptsize{27.} & {Rouge} & \scriptsize{43.} & {La} & \scriptsize{4.2} & {Red} & \scriptsize{36.} & {Rot} & \scriptsize{51.} & {"} & \scriptsize{4.6} & {red} & \scriptsize{30.} \\
        {Rojo} & \scriptsize{34.} & {} & \scriptsize{3.4} & {red} & \scriptsize{15.} & {rouge} & \scriptsize{31.} & {"} & \scriptsize{3.4} & {red} & \scriptsize{19.} & {rot} & \scriptsize{28.} & {Sch} & \scriptsize{2.8} & {Red} & \scriptsize{17.} \\
        {rojo} & \scriptsize{2.1} & {"} & \scriptsize{3.2} & {"} & \scriptsize{8.1} & {Rou} & \scriptsize{4.7} & {} & \scriptsize{3.2} & {The} & \scriptsize{5.9} & {rote} & \scriptsize{5.8} & {ge} & \scriptsize{2.5} & {rojo} & \scriptsize{6.1} \\
        {Ro} & \scriptsize{2.0} & {El} & \scriptsize{2.8} & {The} & \scriptsize{5.1} & {ROU} & \scriptsize{2.6} & {Le} & \scriptsize{2.7} & {Rouge} & \scriptsize{5.6} & {Rote} & \scriptsize{3.5} & {} & \scriptsize{2.2} & {Rojo} & \scriptsize{4.3} \\
        {Roja} & \scriptsize{1.9} & {Bol} & \scriptsize{2.4} & {...} & \scriptsize{4.1} & {Le} & \scriptsize{2.2} & {En} & \scriptsize{2.6} & {"} & \scriptsize{5.5} & {ro} & \scriptsize{2.0} & {Gel} & \scriptsize{2.2} & {rouge} & \scriptsize{3.7} \\        \bottomrule
    \end{tabular}
    \vspace{5pt}
    \caption{
    Top 5 output tokens and their likelihoods for \textit{love} (top) and \textit{red} (bottom) across all three languages, before and after ablating the relevant word or language component. Underscore prefixes of tokens are omitted for space.}
    \label{tab:ablation_top_tokens translation}
\end{table*}

\begin{table*}
    \centering
    \setlength{\tabcolsep}{4.5pt}
    \small
    \begin{tabular}{llllllllllllllllll}
        \toprule
        \multicolumn{6}{c}{\textbf{Past Tense, $ \frac{\text{break}}{\text{like}}$}} & \multicolumn{6}{c}{\textbf{Capitalize, $ \frac{\text{break}}{\text{like}}$}} & \multicolumn{6}{c}{\textbf{1st Letter, $ \frac{\text{break}}{\text{like}}$}} \\
        \cmidrule[0.6pt](lr){1-6} \cmidrule[0.6pt](lr){7-12} \cmidrule[0.6pt](lr){13-18}
        \multicolumn{2}{c}{\textit{Original}} & \multicolumn{2}{c}{\textit{Verb Abl. }} & \multicolumn{2}{c}{\textit{Trans. Abl.}} & \multicolumn{2}{c}{\textit{Original}} & \multicolumn{2}{c}{\textit{Verb Abl. }} & \multicolumn{2}{c}{\textit{Trans. Abl.}} & \multicolumn{2}{c}{\textit{Original}} & \multicolumn{2}{c}{\textit{Verb Abl. }} & \multicolumn{2}{c}{\textit{Trans. Abl.}} \\
        \cmidrule(lr){1-2}\cmidrule(lr){3-4}\cmidrule(lr){5-6}\cmidrule(lr){7-8}\cmidrule(lr){9-10}\cmidrule(lr){11-12}\cmidrule(lr){13-14}\cmidrule(lr){15-16}\cmidrule(lr){17-18}
        {Broke} & \scriptsize{37.} & {-} & \scriptsize{9.9} & {Past} & \scriptsize{8.7} & {Break} & \scriptsize{22.} & {Capital} & \scriptsize{10.} & {break} & \scriptsize{20.} & {B} & \scriptsize{62.} & {T} & \scriptsize{40.} & {K} & \scriptsize{30.} \\
        {broke} & \scriptsize{27.} & {T} & \scriptsize{5.0} & {T} & \scriptsize{7.7} & {break} & \scriptsize{13.} & {The} & \scriptsize{7.6} & {Break} & \scriptsize{7.1} & {b} & \scriptsize{17.} & {C} & \scriptsize{11.} & {E} & \scriptsize{14.} \\
        {Broken} & \scriptsize{7.6} & {Was} & \scriptsize{4.6} & {(} & \scriptsize{7.4} & {BREAK} & \scriptsize{8.3} & {} & \scriptsize{4.2} & {} & \scriptsize{5.1} & {A} & \scriptsize{1.9} & {A} & \scriptsize{10.} & {k} & \scriptsize{13.} \\
        {BRO} & \scriptsize{6.2} & {To} & \scriptsize{4.5} & {t} & \scriptsize{5.6} & {Capital} & \scriptsize{5.7} & {I} & \scriptsize{3.6} & {capital} & \scriptsize{4.3} & {E} & \scriptsize{1.5} & {O} & \scriptsize{9.9} & {A} & \scriptsize{5.7} \\
        {Bre} & \scriptsize{2.8} & {O} & \scriptsize{4.3} & {E} & \scriptsize{5.1} & {The} & \scriptsize{5.0} & {Yes} & \scriptsize{2.3} & {The} & \scriptsize{3.3} & {Break} & \scriptsize{1.3} & {t} & \scriptsize{6.7} & {e} & \scriptsize{5.1} \\\cmidrule(lr){1-2}\cmidrule(lr){3-4}\cmidrule(lr){5-6}\cmidrule(lr){7-8}\cmidrule(lr){9-10}\cmidrule(lr){11-12}\cmidrule(lr){13-14}\cmidrule(lr){15-16}\cmidrule(lr){17-18}
        {liked} & \scriptsize{48.} & {was} & \scriptsize{8.7} & {like} & \scriptsize{8.4} & {Like} & \scriptsize{21.} & {Capital} & \scriptsize{5.6} & {like} & \scriptsize{12.} & {L} & \scriptsize{38.} & {E} & \scriptsize{48.} & {A} & \scriptsize{13.} \\
        {Liked} & \scriptsize{34.} & {have} & \scriptsize{7.6} & {Past} & \scriptsize{8.3} & {like} & \scriptsize{12.} & {The} & \scriptsize{4.7} & {} & \scriptsize{6.1} & {l} & \scriptsize{8.7} & {e} & \scriptsize{9.8} & {I} & \scriptsize{6.8} \\
        {-} & \scriptsize{1.1} & {Have} & \scriptsize{7.0} & {T} & \scriptsize{8.2} & {Capital} & \scriptsize{8.6} & {Yes} & \scriptsize{4.5} & {capital} & \scriptsize{5.6} & {A} & \scriptsize{6.2} & {H} & \scriptsize{9.5} & {a} & \scriptsize{6.7} \\
        {Loved} & \scriptsize{1.0} & {had} & \scriptsize{5.9} & {t} & \scriptsize{7.2} & {LIKE} & \scriptsize{4.3} & {No} & \scriptsize{4.1} & {the} & \scriptsize{4.6} & {I} & \scriptsize{4.1} & {F} & \scriptsize{6.6} & {E} & \scriptsize{4.6} \\
        {Liked} & \scriptsize{1.0} & {-} & \scriptsize{5.9} & {Like} & \scriptsize{7.2} & {The} & \scriptsize{4.0} & {} & \scriptsize{3.3} & {Like} & \scriptsize{3.8} & {"} & \scriptsize{3.2} & {T} & \scriptsize{3.8} & {"} & \scriptsize{4.2} \\\bottomrule
    \end{tabular}
    \vspace{5pt}
    \caption{
    Top 5 output tokens and their likelihoods for \textit{break} (top) and \textit{like} (bottom) across three verb transformations, before and after ablating the relevant verb or transformation component. Underscore prefixes of tokens are omitted for space.}
    \label{tab:ablation_top_tokens verbs}
\end{table*}

\subsection{Individual Concept and Relation Steering}\label{apx: additional examples ind steering}

Extending Table~\ref{tab:steering_top_tokens}, we provide three comprehensive sets of examples for steering individual concepts and relations: Table~\ref{tab:steering_top_tokens countries} (for country facts), Table~\ref{tab:steering_top_tokens translation} (for word translation), and Table~\ref{tab:ablation_top_tokens verbs} (for verb transformations). Observations from these extended examples largely resemble what we observed in Table~\ref{tab:steering_top_tokens}.

% Table~\ref{tab:steering_top_tokens verbs} provides an example for the verb transformations task, when we ablated the \textit{break} component and amplified the \textit{like} component, the model correctly responded with the past tense, capitalization, and first letter of the word \textit{like} despite being prompted to do so for \textit{break}. When the capitalize or first letter components are amplified, the correct token appears in the top 5 most likely tokens in all but one case (when steering from past tense to capitalize). When amplifying the past tense component, the model consistently outputs a distribution of nonsensical tokens. Besides that, other wrong answers are often still relevant, e.g., other capitalized words when steered toward the capitalize transformation, or other letters when steered toward the first letter transformation.

\begin{table*}
    \centering
    \setlength{\tabcolsep}{2.0pt}
    \small
    \begin{tabular}{llllllllllllllllll}
        \toprule
        \multicolumn{6}{c}{\textbf{Capital, CN}} & \multicolumn{6}{c}{\textbf{Currency, CN}} & \multicolumn{6}{c}{\textbf{Language, CN}} \\
        \cmidrule[0.7pt](lr){1-6} \cmidrule[0.7pt](lr){7-12} \cmidrule[0.7pt](lr){13-18}
        \multicolumn{2}{c}{\textit{$\cdot$, NG}} & \multicolumn{2}{c}{\textit{Currency, $\cdot$}} & \multicolumn{2}{c}{\textit{Language, $\cdot$}} & \multicolumn{2}{c}{$\cdot$, \textit{NG}} & \multicolumn{2}{c}{\textit{Capital, $\cdot$}} & \multicolumn{2}{c}{\textit{Language, $\cdot$}} & \multicolumn{2}{c}{\textit{$\cdot$, NG}} & \multicolumn{2}{c}{\textit{Capital, $\cdot$}} & \multicolumn{2}{c}{\textit{Currency, $\cdot$}} \\
        \cmidrule(lr){1-2}\cmidrule(lr){3-4}\cmidrule(lr){5-6}\cmidrule(lr){7-8}\cmidrule(lr){9-10}\cmidrule(lr){11-12}\cmidrule(lr){13-14}\cmidrule(lr){15-16}\cmidrule(lr){17-18}
        \hl{\_Abuja} & \scriptsize{87.} & \hl{\_Yuan} & \scriptsize{13.} & \hl{\_Mandarin} & \scriptsize{35.} & \hl{\_Naira} & \scriptsize{75.} & \hl{\_Beijing} & \scriptsize{97.} & \hl{\_Chinese} & \scriptsize{38.} & \hl{\_English} & \scriptsize{71.} & \hl{\_Beijing} & \scriptsize{96.} & \hl{\_Yuan} & \scriptsize{12.} \\
        {\_Nigeria} & \scriptsize{4.6} & \hl{\_yuan} & \scriptsize{11.} & \hl{\_Chinese} & \scriptsize{28.} & \hl{\_naira} & \scriptsize{8.3} & \hl{\_Be} & \scriptsize{.94} & \hl{\_Mandarin} & \scriptsize{38.} & {\_Yoruba} & \scriptsize{5.6} & \hl{\_Be} & \scriptsize{2.0}  & \hl{\_Ren} & \scriptsize{8.4} \\
        {\_Lagos} & \scriptsize{3.9} & \hl{\_RMB} & \scriptsize{11.} & {\_English} & \scriptsize{20.} & \hl{\_Nai} & \scriptsize{3.7} & \hl{\_BE} & \scriptsize{.57} & {\_English} & \scriptsize{20.} & {\_Ha} & \scriptsize{4.8} & \hl{\_Peking} & \scriptsize{.93}  & {\_China} & \scriptsize{6.1} \\
        {\_-} & \scriptsize{.58} & {\_China} & \scriptsize{7.0} & \hl{\_mandarin} & \scriptsize{2.4} & \hl{\_Nigeria} & \scriptsize{3.3} & \hl{\_Peking} & \scriptsize{.33} & \hl{\_Simplified} & \scriptsize{.92} & {\_Nigeria} & \scriptsize{4.4} & \hl{\_BE} & \scriptsize{.53} & {\_Chinese} & \scriptsize{5.9} \\
        {\_} & \scriptsize{.57} & \hl{\_Ren} & \scriptsize{4.9} & {\_Spanish} & \scriptsize{1.8} & \hl{\_Nigerian} & \scriptsize{2.4} & \hl{Beijing} & \scriptsize{.20} & {\_Spanish} & \scriptsize{.56} & {\_Igbo} & \scriptsize{3.4} & \hl{Beijing} & \scriptsize{.37} & \hl{\_RMB} & \scriptsize{5.9} \\
        \bottomrule
    \end{tabular}

    \caption{Top 5 output tokens and likelihoods for prompts about China's capital (columns 1–3), currency (columns 4–6), and language (columns 7–9), after ablating an in-prompt country or fact component (row 1 headings) and amplifying a target country (Nigeria) or fact component (row 2 headings). Highlighted tokens are or may begin correct answers for steered components. }
    \label{tab:steering_top_tokens countries}
\end{table*}

\begin{table*}
    \centering
    \setlength{\tabcolsep}{4.0pt}
    \small
    \begin{tabular}{llllllllllllllllll}
        \toprule
        \multicolumn{6}{c}{\textbf{Spanish, love}} & \multicolumn{6}{c}{\textbf{French, love}} & \multicolumn{6}{c}{\textbf{German, love}} \\
        \cmidrule[0.7pt](lr){1-6} \cmidrule[0.7pt](lr){7-12} \cmidrule[0.7pt](lr){13-18}
        \multicolumn{2}{c}{\textit{$\cdot$, red}} & \multicolumn{2}{c}{\textit{French, $\cdot$}} & \multicolumn{2}{c}{\textit{German, $\cdot$}} & \multicolumn{2}{c}{$\cdot$, \textit{red}} & \multicolumn{2}{c}{\textit{Spanish, $\cdot$}} & \multicolumn{2}{c}{\textit{German, $\cdot$}} & \multicolumn{2}{c}{\textit{$\cdot$, red}} & \multicolumn{2}{c}{\textit{Spanish, $\cdot$}} & \multicolumn{2}{c}{\textit{French, $\cdot$}} \\
        \cmidrule(lr){1-2}\cmidrule(lr){3-4}\cmidrule(lr){5-6}\cmidrule(lr){7-8}\cmidrule(lr){9-10}\cmidrule(lr){11-12}\cmidrule(lr){13-14}\cmidrule(lr){15-16}\cmidrule(lr){17-18}
        \hl{\_rojo} & \scriptsize{43.} & \hl{\_amour} & \scriptsize{14.}      & \hl{\_Liebe} & \scriptsize{71.}      & \hl{\_rouge} & \scriptsize{44.}  & \hl{\_amor} & \scriptsize{41.}   & \hl{\_Liebe} & \scriptsize{68.}      & \hl{\_rot} & \scriptsize{53.}  & \hl{\_amor} & \scriptsize{50.}     & \hl{\_amour} & \scriptsize{14.}  \\
        \hl{\_Rojo} & \scriptsize{8.7} & \hl{\_A} & \scriptsize{11.}      & {\_Ich} & \scriptsize{2.3}        & \hl{\_Rouge} & \scriptsize{30.}  & \hl{\_Amor} & \scriptsize{13.}     & {\_die} & \scriptsize{2.4}        & \hl{\_Rot} & \scriptsize{33.}  & {\_el} & \scriptsize{8.6}       & {\_"} & \scriptsize{9.7}  \\
        {\_Sang} & \scriptsize{5.7} & \hl{\_"} & \scriptsize{6.7}  & {\_die} & \scriptsize{1.5}         & {\_Sang} & \scriptsize{2.5}   & {\_el} & \scriptsize{6.9}   & {\_„} & \scriptsize{2.1}          & \hl{\_rote} & \scriptsize{3.6}  & {\_"} & \scriptsize{7.4}        & {\_l} & \scriptsize{7.5}  \\
        \hl{\_roja} & \scriptsize{2.8} & {\_l} & \scriptsize{6.5}      & {\_„} & \scriptsize{1.5}          & \hl{\_Rou} & \scriptsize{1.8}    & {\_"} & \scriptsize{6.1}      & \hl{\_lieben} & \scriptsize{1.7}     & \hl{\_Rote} & \scriptsize{2.0}  & \hl{\_Amor} & \scriptsize{3.7}        & \hl{\_A} & \scriptsize{6.7}  \\
        \hl{\_rojo} & \scriptsize{1.8} & \hl{\_Amour} & \scriptsize{5.3}       & {\_zu} & \scriptsize{1.4}        & {\_} & \scriptsize{1.2}       & \hl{\_A} & \scriptsize{3.1}      & {\_Ich} & \scriptsize{1.6}        & {\_Blut} & \scriptsize{1.2}  & \hl{\_a} & \scriptsize{2.7}       & {\_} & \scriptsize{6.5}  \\
        \bottomrule
    \end{tabular}

    \caption{Top 5 output tokens and likelihoods for prompts to translate the word \textit{love} into Spanish (columns 1–3), French (columns 4–6), and German (columns 7–9), after ablating an in-prompt word or language component (row 1 headings) and amplifying a target word (\textit{red}) or language component (row 2 headings). Highlighted tokens are or may begin correct answers for steered components.}
    \label{tab:steering_top_tokens translation}
\end{table*}

\begin{table*}
    \centering
    \setlength{\tabcolsep}{0.8pt}
    \small
    \begin{tabular}{llllllllllllllllll}
        \toprule
        \multicolumn{6}{c}{\textbf{Past Tense, understand}} & \multicolumn{6}{c}{\textbf{Capitalize, include}} & \multicolumn{6}{c}{\textbf{1st Letter, break}} \\
        \cmidrule[0.7pt](lr){1-6} \cmidrule[0.7pt](lr){7-12} \cmidrule[0.7pt](lr){13-18}
        \multicolumn{2}{c}{\textit{$\cdot$, break}} & \multicolumn{2}{c}{\textit{Cap., $\cdot$}} & \multicolumn{2}{c}{\textit{1st Lett., $\cdot$}} & \multicolumn{2}{c}{$\cdot$, \textit{understand}} & \multicolumn{2}{c}{\textit{Past Tense, $\cdot$}} & \multicolumn{2}{c}{\textit{1st Lett., $\cdot$}} & \multicolumn{2}{c}{\textit{$\cdot$, focus}} & \multicolumn{2}{c}{\textit{Past Tense, $\cdot$}} & \multicolumn{2}{c}{\textit{Cap., $\cdot$}} \\
        \cmidrule(lr){1-2}\cmidrule(lr){3-4}\cmidrule(lr){5-6}\cmidrule(lr){7-8}\cmidrule(lr){9-10}\cmidrule(lr){11-12}\cmidrule(lr){13-14}\cmidrule(lr){15-16}\cmidrule(lr){17-18}
        \hl{\_broke} & \scriptsize{28.} & \hl{\_Understand} & \scriptsize{6.7} & \hl{\_U} & \scriptsize{39.} & \hl{\_Understand} & \scriptsize{14.} & {$\otimes$} & \scriptsize{35.} & \hl{\_i} & \scriptsize{22.} & \hl{\_F} & \scriptsize{26.} & {$\otimes$} & \scriptsize{30.} & \hl{\_Break} & \scriptsize{21.}  \\
        \hl{\_Broke} & \scriptsize{21.} & {\_Know} & \scriptsize{6.4} & {\_understood} & \scriptsize{9.3} & {\_The} & \scriptsize{10.} & {\_surla} & \scriptsize{12.} & \hl{\_I} & \scriptsize{21.} & {\_C} & \scriptsize{18.} & {\_surla} & \scriptsize{16..} & {\_"} & \scriptsize{6.1}  \\
        \hl{\_BRO} & \scriptsize{8.5} & {\_Past} & \scriptsize{5.9} & {\_Understand} & \scriptsize{8.5} & {\_I} & \scriptsize{6.3} & {AddTagHelper} & \scriptsize{13.} & {\_include} & \scriptsize{21.} & {\_O} & \scriptsize{15.} & {AddTagHelper} & \scriptsize{13.} & {\_The} & \scriptsize{5.6}  \\
        {\_Bre} & \scriptsize{6.3} & {\_Answer} & \scriptsize{5.7} & \hl{\_u} & \scriptsize{3.6} & {\_understand} & \scriptsize{4.3} & {\_The\textit{s}e} & \scriptsize{9.7} & {\_Include} & \scriptsize{6.3} & \hl{\_f} & \scriptsize{4.4} & {\_The\textit{s}e} & \scriptsize{11.} & {\_``} & \scriptsize{5.5}  \\
        {\_brake} & \scriptsize{2.6} & {\_} & \scriptsize{4.5} & {\_United} & \scriptsize{3.2} & {\_} & \scriptsize{4.1} & {\_nahilalakip} & \scriptsize{6.9} & {\_In} & \scriptsize{1.9} & {\_o} & \scriptsize{3.3} & {\_nahilalakip} & \scriptsize{8.6} & {\_} & \scriptsize{3.7}  \\
        \bottomrule
    \end{tabular}

    \caption{Top 5 output tokens and likelihoods for prompts to get the past tense (columns 1–3), capitalized form (columns 4–6), and first letter (columns 7–9) of various verbs, after ablating an in-prompt verb or transformation component (row 1 headings) and amplifying a target verb (\textit{understand}) or transformation component (row 2 headings). Highlighted tokens are or may begin correct answers for steered components. ``\textit{s}'' refers to Old English long S.}
    \label{tab:steering_top_tokens verbs}
\end{table*}

\subsection{Composite Concept and Relation Steering}\label{apx: additional examples comp steering}

Extending Table~\ref{tab:composed_steering_top_tokens}, we provide two full sets of examples for steering concepts and relations in composition: Table~\ref{tab:composed_steering_top_tokens countries} (for country facts) and Table~\ref{tab:composed_steering_top_tokens translation} (for word translation). Observations from these extended examples largely resemble what we observed in Table~\ref{tab:composed_steering_top_tokens}. An unexpected exception is that the top next token when steering from the language of China to the capital of Nigeria is ``\textit{Lagos}'' (another major city in Nigeria) rather than \textit{Abuja}. We note that the full answer from the LLM contains the names of both cities, suggesting competition between these related tokens.

\begin{table*}
    \centering
    \setlength{\tabcolsep}{6pt}
    \small
    \begin{tabular}{llllllllllll}
        \toprule
        \multicolumn{4}{c}{\textbf{Capital, China}} & \multicolumn{4}{c}{\textbf{Currency, China}} & \multicolumn{4}{c}{\textbf{Language, China}} \\
        \cmidrule[0.6pt](lr){1-4} \cmidrule[0.6pt](lr){5-8} \cmidrule[0.6pt](lr){9-12}
        \multicolumn{2}{c}{\textit{Currency, Nigeria}} & \multicolumn{2}{c}{\textit{Language, Nigeria}} & \multicolumn{2}{c}{\textit{Capital, Nigeria}} & \multicolumn{2}{c}{\textit{Language, Nigeria}} & \multicolumn{2}{c}{\textit{Capital, Nigeria}} & \multicolumn{2}{c}{\textit{Currency, Nigeria}} \\
        \cmidrule(lr){1-2}\cmidrule(lr){3-4}\cmidrule(lr){5-6}\cmidrule(lr){7-8}\cmidrule(lr){9-10}\cmidrule(lr){11-12}
        \hl{\_Nigeria} & \scriptsize{43.} & \hl{\_English} & \scriptsize{75.} & \hl{\_Abuja} & \scriptsize{70.} & \hl{\_English} & \scriptsize{96.} & {\_Lagos} & \scriptsize{73.} & \hl{\_Naira} & \scriptsize{48.} \\
        \hl{\_Naira} & \scriptsize{30.} & {\_Yoruba} & \scriptsize{4.0} & {\_Lagos} & \scriptsize{25.} & {\_Yoruba} & \scriptsize{1.1} & \hl{\_Abuja} & \scriptsize{23.} & \hl{\_Nigeria} & \scriptsize{17.} \\
        \hl{\_naira} & \scriptsize{8.9} & {\_Igbo} & \scriptsize{3.6} & \_ & \scriptsize{3.3} & {\_French} & \scriptsize{.74} & {\_} & \scriptsize{2.3} & {\_Dollar} & \scriptsize{5.8} \\
        \hl{\_N} & \scriptsize{3.0} & {\_French} & \scriptsize{3.5} & {\_Nigeria} & \scriptsize{.34} & {\_Spanish} & \scriptsize{.60} & {\_Nigeria} & \scriptsize{.46} & {\_} & \scriptsize{4.5} \\
        {\_} & \scriptsize{2.8} & {\_Spanish} & \scriptsize{3.0} & {\_...} & \scriptsize{.32} & {\_Igbo} & \scriptsize{.54} & {\_...} & \scriptsize{.32} & \hl{\_naira} & \scriptsize{4.1} \\
        \bottomrule
    \end{tabular}

    \caption{Top 5 output tokens and their likelihoods for prompts about China's capital (columns 1–2), currency (3–4), and language (5–6), after ablating both in-prompt components and amplifying a target country-fact pair. Highlighted tokens are or may begin correct answers for steered components. }
    \label{tab:composed_steering_top_tokens countries}
\end{table*}

\begin{table*}
    \centering
    \setlength{\tabcolsep}{11pt}
    \small
    \begin{tabular}{llllllllllll}
        \toprule
        \multicolumn{4}{c}{\textbf{Spanish, love}} & \multicolumn{4}{c}{\textbf{French, love}} & \multicolumn{4}{c}{\textbf{German, love}} \\
        \cmidrule[0.6pt](lr){1-4} \cmidrule[0.6pt](lr){5-8} \cmidrule[0.6pt](lr){9-12}
        \multicolumn{2}{c}{\textit{French, red}} & \multicolumn{2}{c}{\textit{German, red}} & \multicolumn{2}{c}{\textit{Spanish, red}} & \multicolumn{2}{c}{\textit{German, red}} & \multicolumn{2}{c}{\textit{Spanish, red}} & \multicolumn{2}{c}{\textit{French, red}} \\
        \cmidrule(lr){1-2}\cmidrule(lr){3-4}\cmidrule(lr){5-6}\cmidrule(lr){7-8}\cmidrule(lr){9-10}\cmidrule(lr){11-12}
        \hl{\_rouge} & \scriptsize{47.}     & \hl{\_rot} & \scriptsize{31.}     & \hl{\_rojo} & \scriptsize{43.}    & \hl{\_rot} & \scriptsize{28.}     & \hl{\_rojo} & \scriptsize{53.}    & \hl{\_rouge} & \scriptsize{45.}  \\
        \hl{\_Rouge} & \scriptsize{8.2}     & \hl{\_Rot} & \scriptsize{26.}     & \hl{\_roja} & \scriptsize{6.5}         & \hl{\_Rot} & \scriptsize{20.}     & {\_el} & \scriptsize{7.2}         & {\_le} & \scriptsize{6.7}  \\
        {\_Sang} & \scriptsize{3.2}           & \hl{\_rote} & \scriptsize{11.}    & {\_el} & \scriptsize{6.2}    & \hl{\_rote} & \scriptsize{9.0}    & \hl{\_roja} & \scriptsize{5.6}    & \hl{\_Rouge} & \scriptsize{4.2}  \\
        {\_sang} & \scriptsize{1.9}         & {\_die} & \scriptsize{2.1}        & {\_El} & \scriptsize{3.2}         & {\_die} & \scriptsize{2.9}        & {\_"} & \scriptsize{3.1}          & {\_} & \scriptsize{3.7}  \\
        {\_} & \scriptsize{1.9}           & \hl{\_Rote} & \scriptsize{1.6}        & {\_"} & \scriptsize{2.5}         & \hl{\_roten} & \scriptsize{2.5}          & {\_El} & \scriptsize{2.2}         & {\_la} & \scriptsize{3.4}  \\\bottomrule
    \end{tabular}

    \caption{Top 5 output tokens and their likelihoods for prompts to translate \textit{love} into Spanish (columns 1–2), French (3–4), and German (5–6), after ablating both in-prompt components and amplifying a target word-language pair. Highlighted tokens are or may begin correct answers for steered components.}
    \label{tab:composed_steering_top_tokens translation}
\end{table*}

\section{Steering Results by Concept}\label{apx: additional tables}

Table~\ref{tab:all steering} summarized steering success rates by relation in each task. Here, Tables~\ref{tab:country_steering}-\ref{tab:words concept steering} break down concept steering success rate by concepts, while Tables~\ref{tab:relation_country_steering}-\ref{tab:words composite steering} break down composite steering success rates by concept-relation pairs.

\begin{table*}
    \centering
    \setlength{\tabcolsep}{6.6pt}
    \small
    \begin{tabular}{rccccccccccc}
        \toprule
        \textit{Country} & \textbf{CN} & \textbf{FR} & \textbf{DE} & \textbf{JP} & \textbf{NG} & \textbf{PL} & \textbf{RU} & \textbf{ES} & \textbf{UK} & \textbf{US} & \textbf{Average} \\
        \cmidrule(lr){1-1}\cmidrule(lr){2-11}\cmidrule(lr){12-12}
        \textit{Success Rate} & 1.00 & 1.00 & 1.00 & 1.00 & 1.00 & 0.78 & 1.00 & 0.78 & 1.00 & 1.00 & 0.96 \\
        \bottomrule
    \end{tabular}
    
    \caption{Steering success rate for each target country across all 27 prompts for other countries.}
    \label{tab:country_steering}
\end{table*}

% \begin{table*}
%     \centering
%     \setlength{\tabcolsep}{6.4pt}
%     \small
%     \begin{tabular}{rccccccccc}
%         \toprule
%         \textit{Animal} & \textbf{American crow} & \textbf{Bald eagle} & \textbf{Fire ant} & \textbf{Dojo loach} & \textbf{Goldfish} & \textbf{Gray wolf} & \textbf{Lion}  & \textbf{Average} \\
%         \cmidrule(lr){1-1}\cmidrule(lr){2-3}\cmidrule(lr){4-4}\cmidrule(lr){5-6}\cmidrule(lr){7-8}\cmidrule(lr){9-9}
%         \textit{Success Rate} & 1.00 & 0.39 & 0.52 & 0.39 & 0.44 & 0.72 & 1.00 & 0.53 \\
%         \bottomrule
%     \end{tabular}
    
%     \caption{Steering success rate for selected target animals across all 75 prompts for other animals, and overall average.}
%     \label{tab:animal_steering}
% \end{table*}

\begin{table*}
    \centering
    \setlength{\tabcolsep}{4.7pt}
    \footnotesize
    \begin{tabular}{rcccccccccccc}
        \toprule
        \textit{Word} &  \textbf{cat} & \textbf{dog} & \textbf{fish} & \textbf{you} & \textbf{beautiful} & \textbf{good} & \textbf{red} & \textbf{here} & \textbf{learn} & \textbf{love} & \textbf{run} & \textbf{Average} \\
        \cmidrule(lr){1-1}\cmidrule(lr){2-5}\cmidrule(lr){6-8}\cmidrule(lr){9-9}\cmidrule(lr){10-12}\cmidrule(lr){13-13}
        \textit{Success Rate} & 0.60 & 0.87 & 0.47 & 0.57 & 0.93 & 0.83 & 0.90 & 0.47 & 1.00 & 1.00 & 0.63 & 0.75 \\
        \bottomrule
    \end{tabular}
    
    \caption{Steering success rate for each target translated word across all 30 prompts for other words.}
    \label{tab:translation concept steering}
\end{table*}

\begin{table*}
    \centering
    \setlength{\tabcolsep}{4.7pt}
    \footnotesize
    \begin{tabular}{rccccccccc}
        \toprule
        \textit{Verb} & \textbf{break} & \textbf{focus} & \textbf{hide} & \textbf{include} & \textbf{like} & \textbf{possess} &  \textbf{sink} &  \textbf{understand} & \textbf{Average} \\
        \cmidrule(lr){1-1}\cmidrule(lr){2-9}\cmidrule(lr){10-10}
        \textit{Success Rate} & 0.66 & 0.34 & 0.69 & 0.63 & 0.83 & 0.06 & 0.20 & 0.46 & 0.48\\
        \bottomrule
    \end{tabular}
    
    \caption{Steering success rate for each target verb across all 35 prompts for other verbs.}
    \label{tab:words concept steering}
\end{table*}

\begin{table*}
    \centering
    \setlength{\tabcolsep}{7.5pt}    
    \small
    \begin{tabular}{rccccccccccc}
        \toprule
        \textit{Ctry. / Fact} & \textbf{CN} & \textbf{FR} & \textbf{DE} & \textbf{JP} & \textbf{NG} & \textbf{PL} & \textbf{RU} & \textbf{ES} & \textbf{UK} & \textbf{US} & \textbf{Average} \\
        \cmidrule(lr){1-1}\cmidrule(lr){2-11}\cmidrule(lr){12-12}
        \textit{Capital} & 1.00 & 1.00 & 1.00 & 1.00 & 1.00 & 1.00 & 1.00 & 1.00 & 1.00 & 1.00 & 1.00 \\
        \textit{Currency} & 1.00 & 0.94 & 1.00 & 0.22 & 0.61 & 0.50 & 0.50 & 0.94 & 1.00 & 1.00 & 0.77 \\
        \textit{Language} & 0.50 & 1.00 & 1.00 & 1.00 & 1.00 & 0.78 & 1.00 & 1.00 & 1.00 & 1.00 & 0.93 \\
        \cmidrule(lr){1-1}\cmidrule(lr){2-11}\cmidrule(lr){12-12}
        \textit{Average} & 0.83 & 0.98 & 1.00 & 0.74 & 0.87 & 0.76 & 0.83 & 0.98 & 1.00 & 1.00 & 0.90 \\
        \bottomrule
    \end{tabular}

    \caption{Composite steering success rates for each target country-fact pair across all 18 prompts that do not already query the target country or fact.}
    \label{tab:relation_country_steering}
\end{table*}

% \begin{table*}
%     \centering
%     \setlength{\tabcolsep}{7.5pt}    
%     \small
%     \begin{tabular}{rcccccccc}
%         \toprule
%         \textit{Animal / Cat.} & \textbf{American crow} & \textbf{Bald eagle} & \textbf{Fire ant} & \textbf{Dojo loach} & \textbf{Goldfish} & \textbf{Gray wolf} & \textbf{Lion}  & \textbf{Average} \\
%         \cmidrule(lr){1-1}\cmidrule(lr){2-3}\cmidrule(lr){4-4}\cmidrule(lr){5-6}\cmidrule(lr){7-8}\cmidrule(lr){9-9}
%         \textit{Class}  & 0.22 & 0.20 & 1.00 & 0.00 & 0.00 & 0.30 & 0.60 & 0.38 \\
%         \textit{Order}  & 0.04 & 0.04 & 0.00 & 0.00 & 0.00 & 0.00 & 0.00 & 0.02 \\
%         \textit{Family} & 0.02 & 0.00 & 0.06 & 0.00 & 0.00 & 0.04 & 0.04 & 0.08 \\
%         \cmidrule(lr){1-1}\cmidrule(lr){2-3}\cmidrule(lr){4-4}\cmidrule(lr){5-6}\cmidrule(lr){7-8}\cmidrule(lr){9-9}
%         \textit{Average} & 0.09 & 0.08 & 0.35 & 0.0 & 0.0 & 0.11 & 0.21 & 0.16 \\
%         \bottomrule
%     \end{tabular}

%     \caption{Composite steering success rates for selected target animal-category pairs across all 50 prompts that do not already query the target animal or category, and overall averages.}
%     \label{tab:taxonomy composite steering}
% \end{table*}

\begin{table*}
    \centering
    \setlength{\tabcolsep}{5pt}    
    \small
    \begin{tabular}{rcccccccccccc}
        \toprule
        \textit{Word / Lang.} &  \textbf{cat} & \textbf{dog} & \textbf{fish} & \textbf{you} & \textbf{beautiful} & \textbf{good} & \textbf{red} & \textbf{here} & \textbf{learn} & \textbf{love} & \textbf{run} & \textbf{Average} \\
        \cmidrule(lr){1-1}\cmidrule(lr){2-5}\cmidrule(lr){6-8}\cmidrule(lr){9-9}\cmidrule(lr){10-12}\cmidrule(lr){13-13}
        \textit{Spanish} & 0.55 & 0.65 & 0.10 & 0.50 & 1.00 & 0.20 & 0.75 & 0.55 & 0.70 & 1.00 & 0.25 & 0.57 \\
        \textit{French}  & 0.75 & 0.90 & 0.05 & 0.55 & 0.95 & 0.90 & 0.65 & 0.50 & 0.95 & 1.00 & 0.35 & 0.69 \\
        \textit{German}  & 0.55 & 0.45 & 0.15 & 0.50 & 0.80 & 1.00 & 0.75 & 0.60 & 1.00 & 1.00 & 0.55 & 0.67 \\\cmidrule(lr){1-1}\cmidrule(lr){2-5}\cmidrule(lr){6-8}\cmidrule(lr){9-9}\cmidrule(lr){10-12}\cmidrule(lr){13-13}
        \textit{Average} & 0.62 & 0.67 & 0.10 & 0.52 & 0.92 & 0.70 & 0.72 & 0.55 & 0.88 & 1.00 & 0.38 & 0.64 \\\bottomrule
    \end{tabular}

    \caption{Composite steering success rates for each target word-language pair across all 20 prompts that do not already query the target word or language.}
    \label{tab:translation composite steering}
\end{table*}

\begin{table*}
    \centering
    \setlength{\tabcolsep}{5pt}    
    \small
    \begin{tabular}{rccccccccc}
        \toprule
        \textit{Verb / Trans.} & \textbf{break} & \textbf{focus} & \textbf{hide} & \textbf{include} & \textbf{like} & \textbf{possess} &  \textbf{sink} &  \textbf{understand} & \textbf{Average} \\\cmidrule(lr){1-1}\cmidrule(lr){2-9}\cmidrule(lr){10-10}
        \textit{Synonym} & 0.00 & 0.00 & 0.00 & 0.00 & 0.14 & 0.00 & 0.00 & 0.00 & 0.02 \\
        \textit{Antonym} & 0.00 & 0.00 & 0.00 & 0.00 & 0.00 & 0.00 & 0.00 & 0.00 & 0.00 \\
        \textit{Past Tense} & 0.00 & 0.00 & 0.00 & 0.00 & 0.00 & 0.00 & 0.00 & 0.00 & 0.00 \\
        \textit{Capitalize}  & 0.00 & 0.00 & 0.93 & 0.14 & 0.18 & 0.00 & 0.00 & 0.04 & 0.16 \\
        \textit{First Letter} & 1.00 & 0.00 & 0.89 & 0.93 & 0.71 & 1.00 & 1.00 & 0.75 & 0.79 \\\cmidrule(lr){1-1}\cmidrule(lr){2-9}\cmidrule(lr){10-10}
        \textit{Average} & 0.20 & 0.00 & 0.36 & 0.21 & 0.21 & 0.20 & 0.20 & 0.16 & 0.19 \\
        \bottomrule
    \end{tabular}

    \caption{Composite steering success rates for each target verb-transformation pair across all 10 prompts that do not already query the target verb or transformation.}
    \label{tab:words composite steering}
\end{table*}

\section{Steering Failure Analysis}\label{apx:steering failure analysis}

\begin{figure*}
    \centering
    \setlength{\columnsep}{0pt}
    \begin{multicols}{3}
        \includegraphics[width=1.0\linewidth]{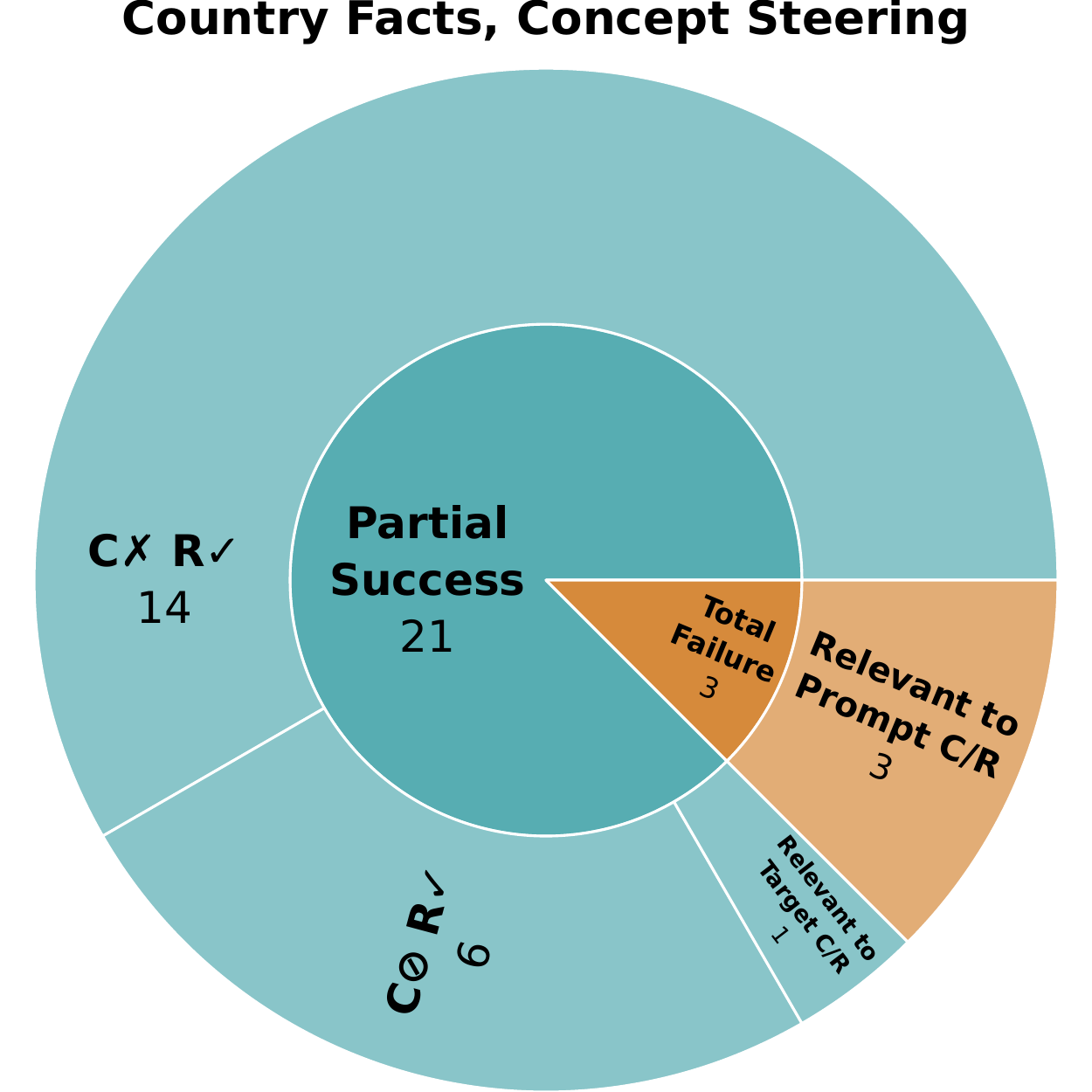}

        \vspace{5pt}

        \includegraphics[width=1.0\linewidth]{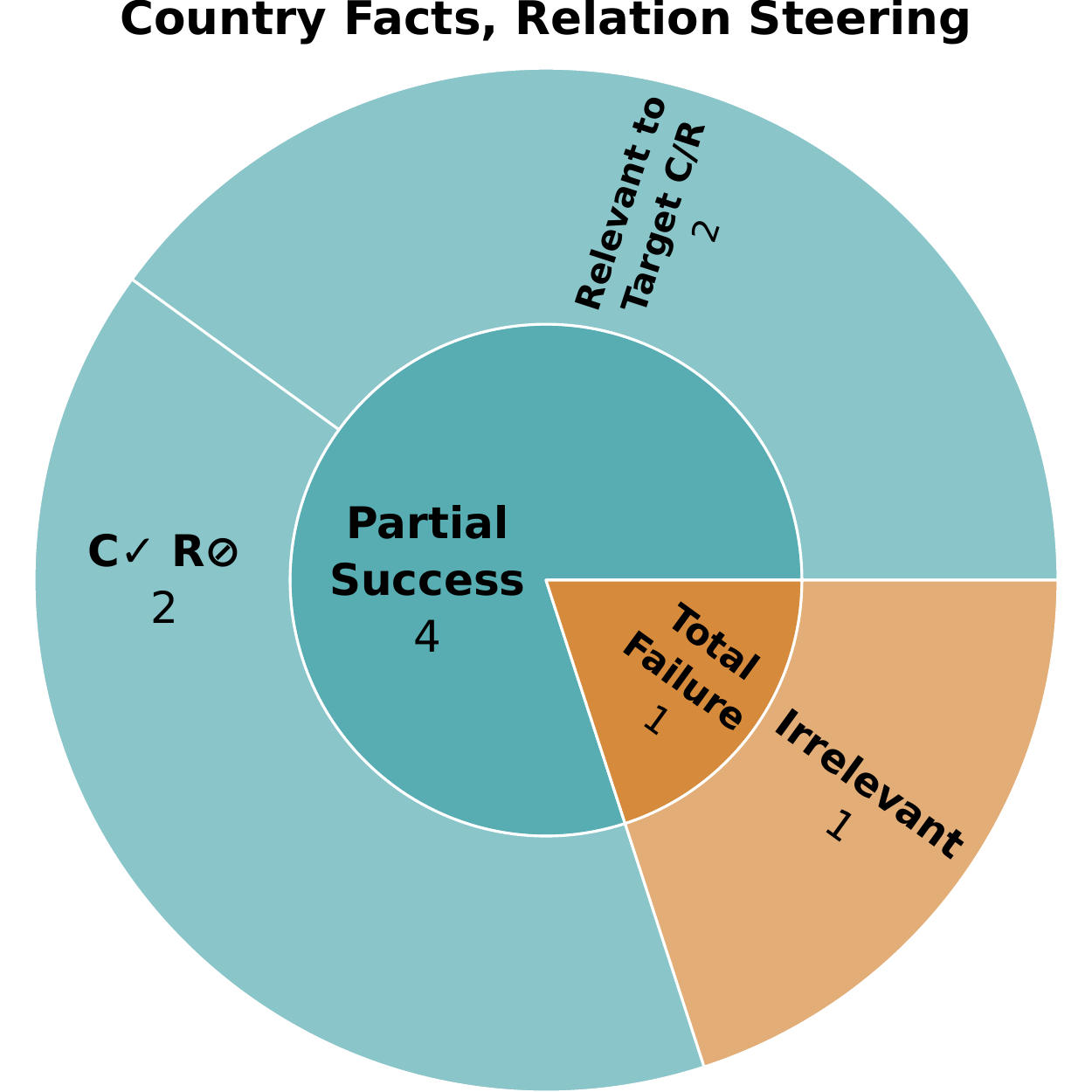}

        \vspace{5pt}

        \includegraphics[width=1.0\linewidth]{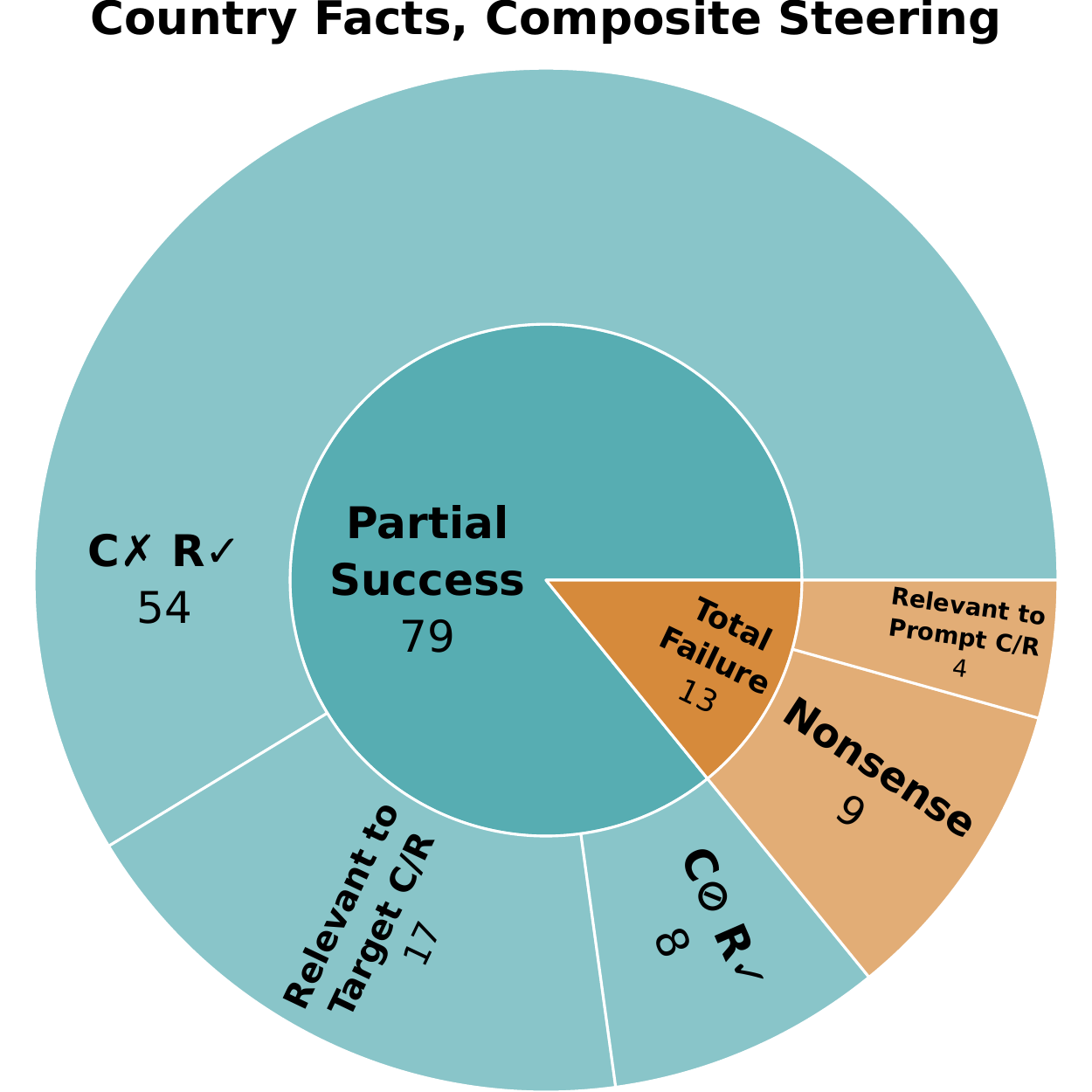}

        \includegraphics[width=1.0\linewidth]{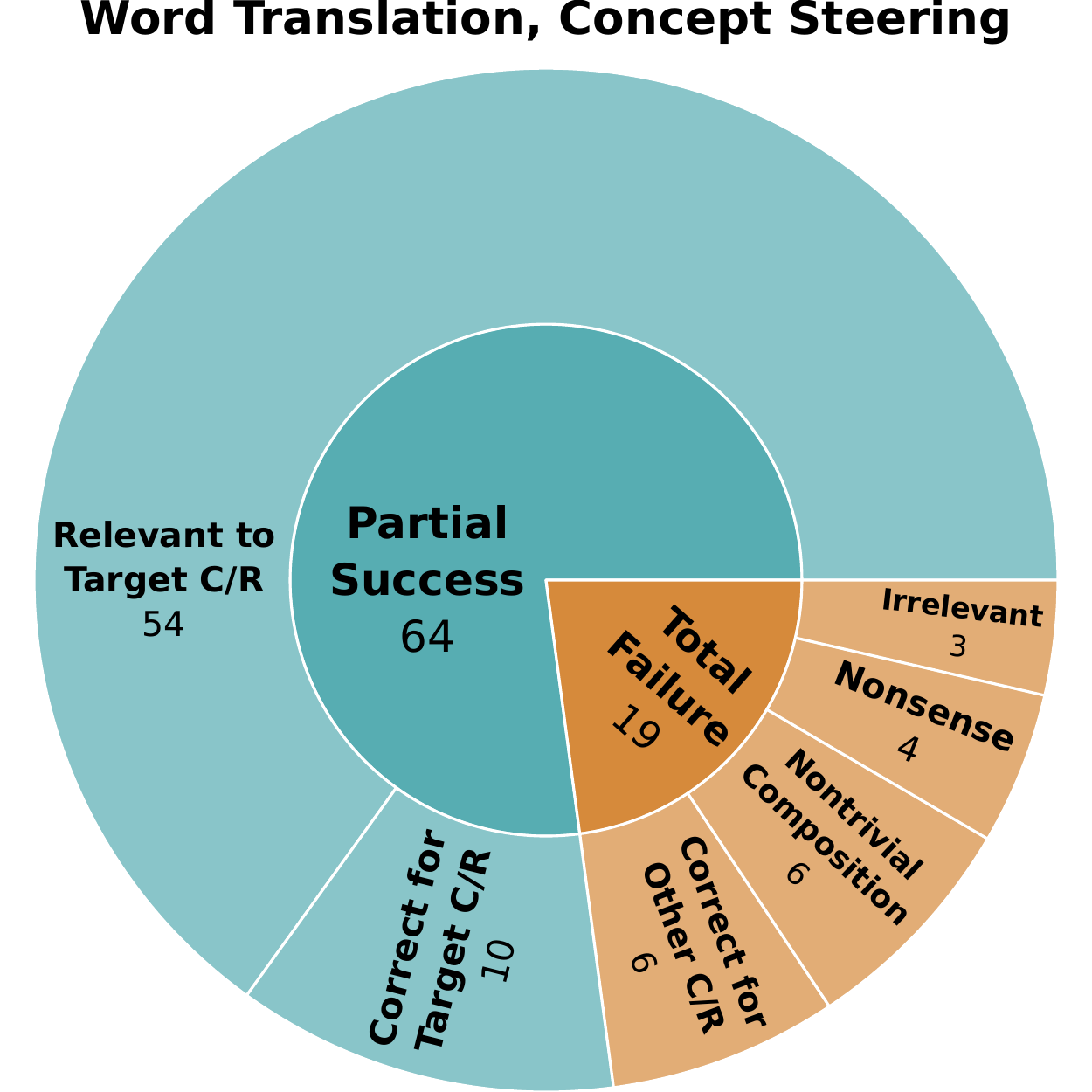}

        \vspace{5pt}

        \includegraphics[width=1.0\linewidth]{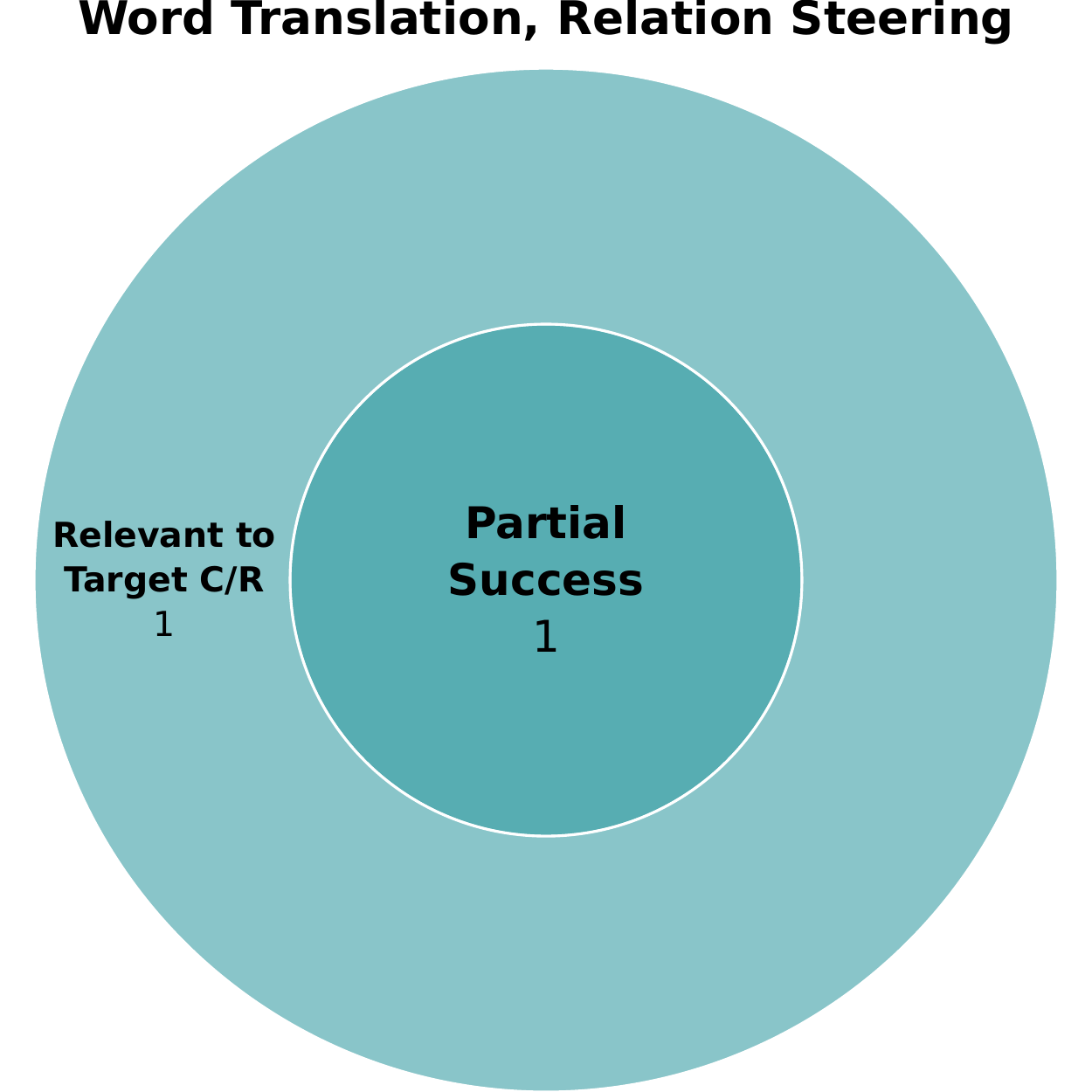}

        \vspace{5pt}

        \includegraphics[width=1.0\linewidth]{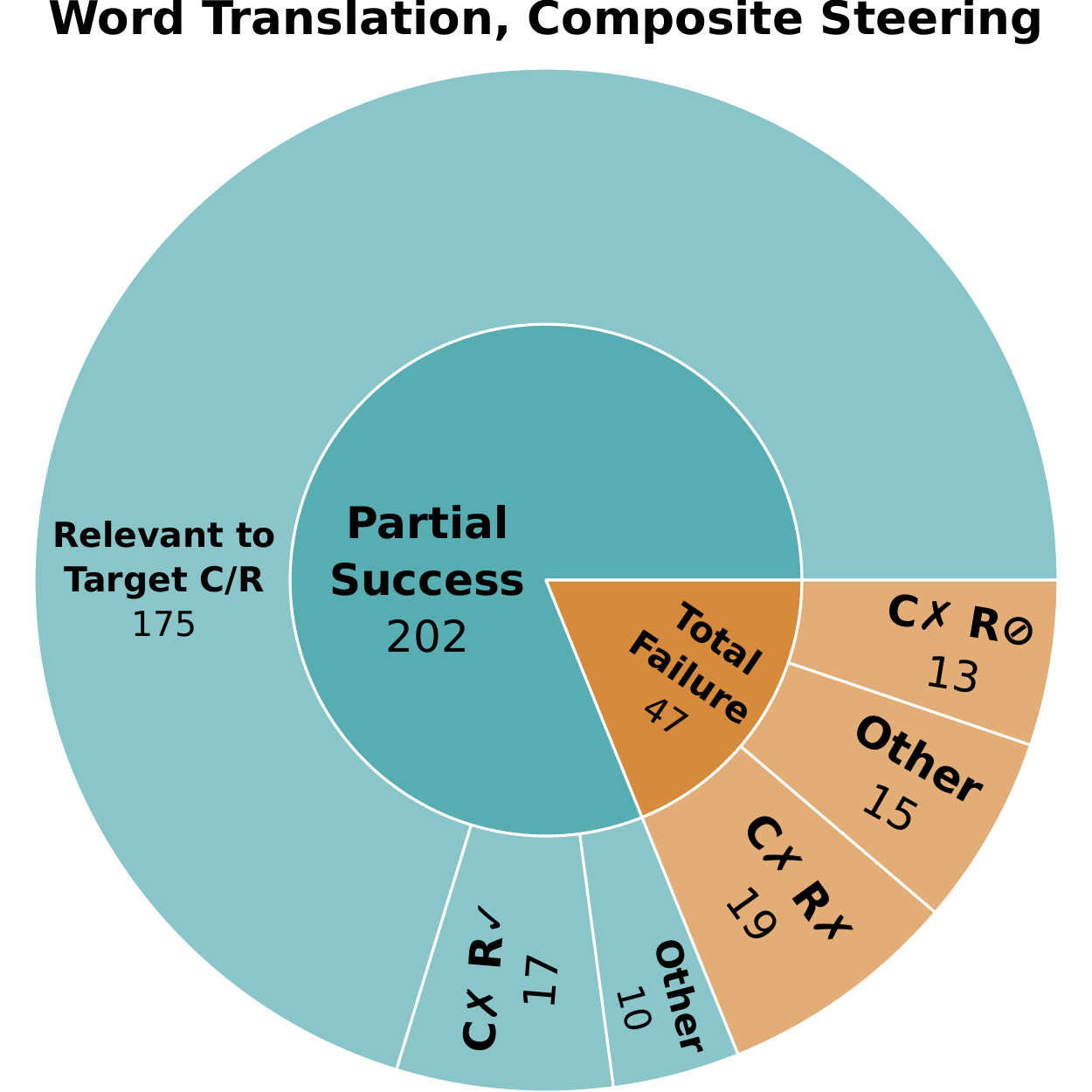}        
        
        \includegraphics[width=1.0\linewidth]{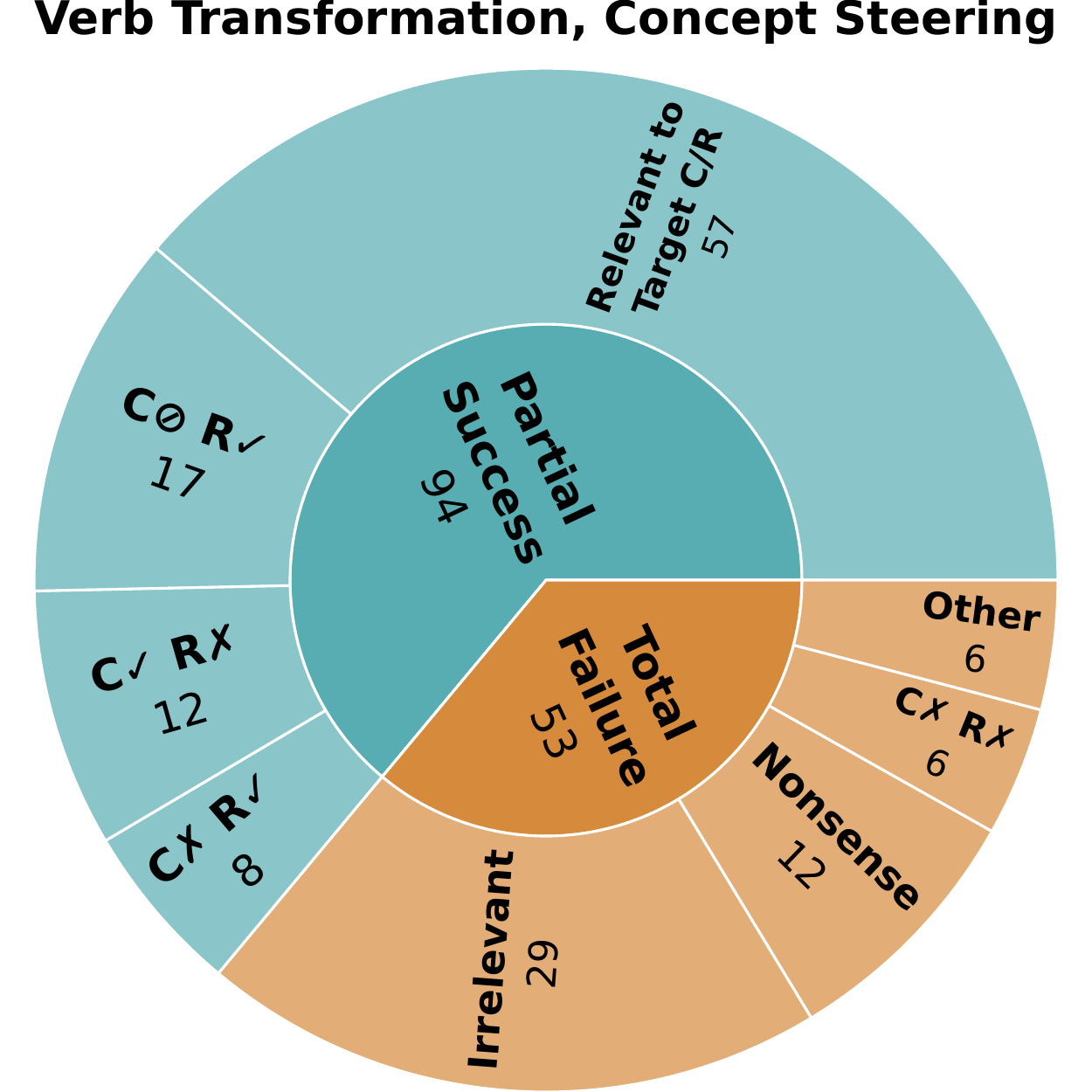}

        \vspace{5pt}

        \includegraphics[width=1.0\linewidth]{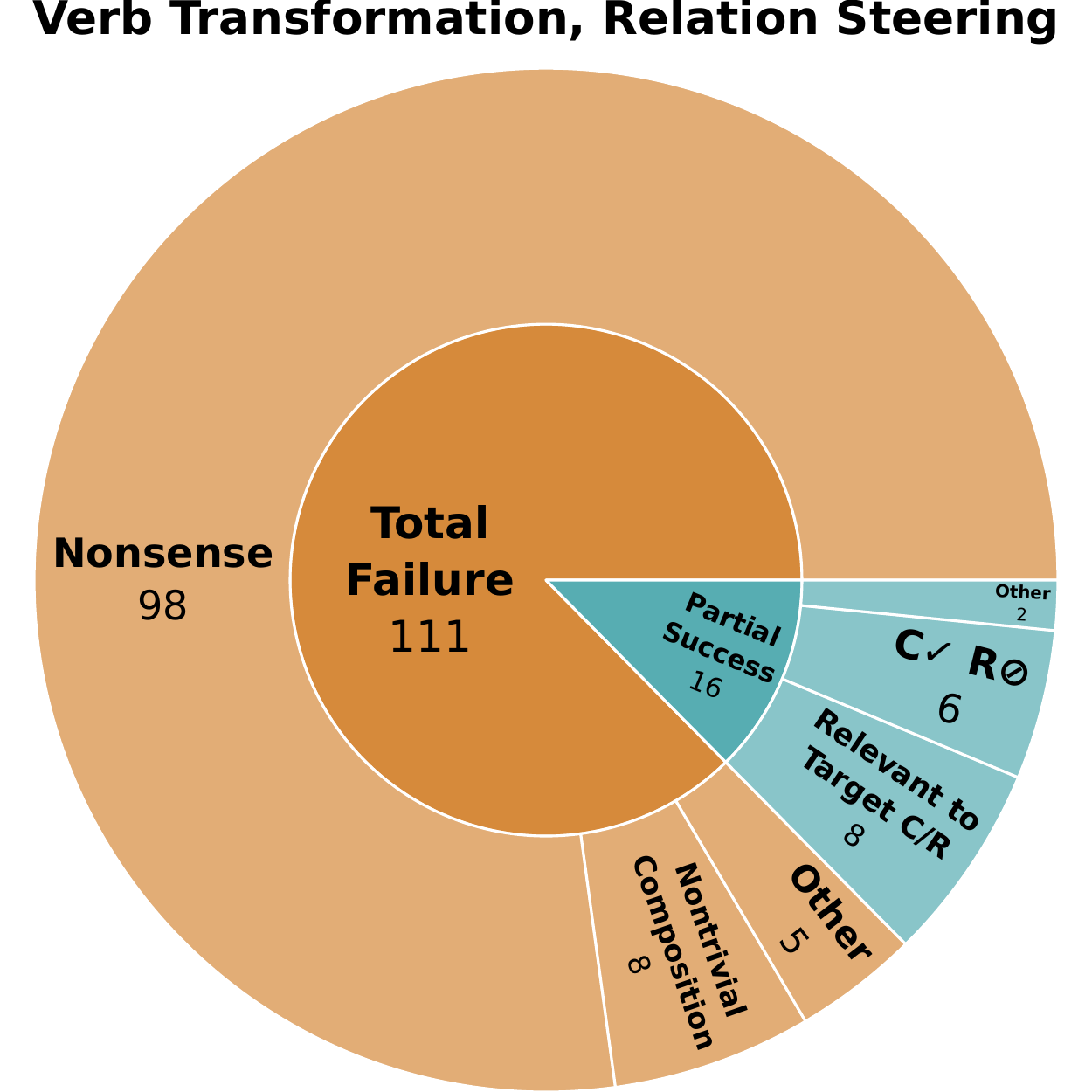}

        \vspace{5pt}

        \includegraphics[width=1.0\linewidth]{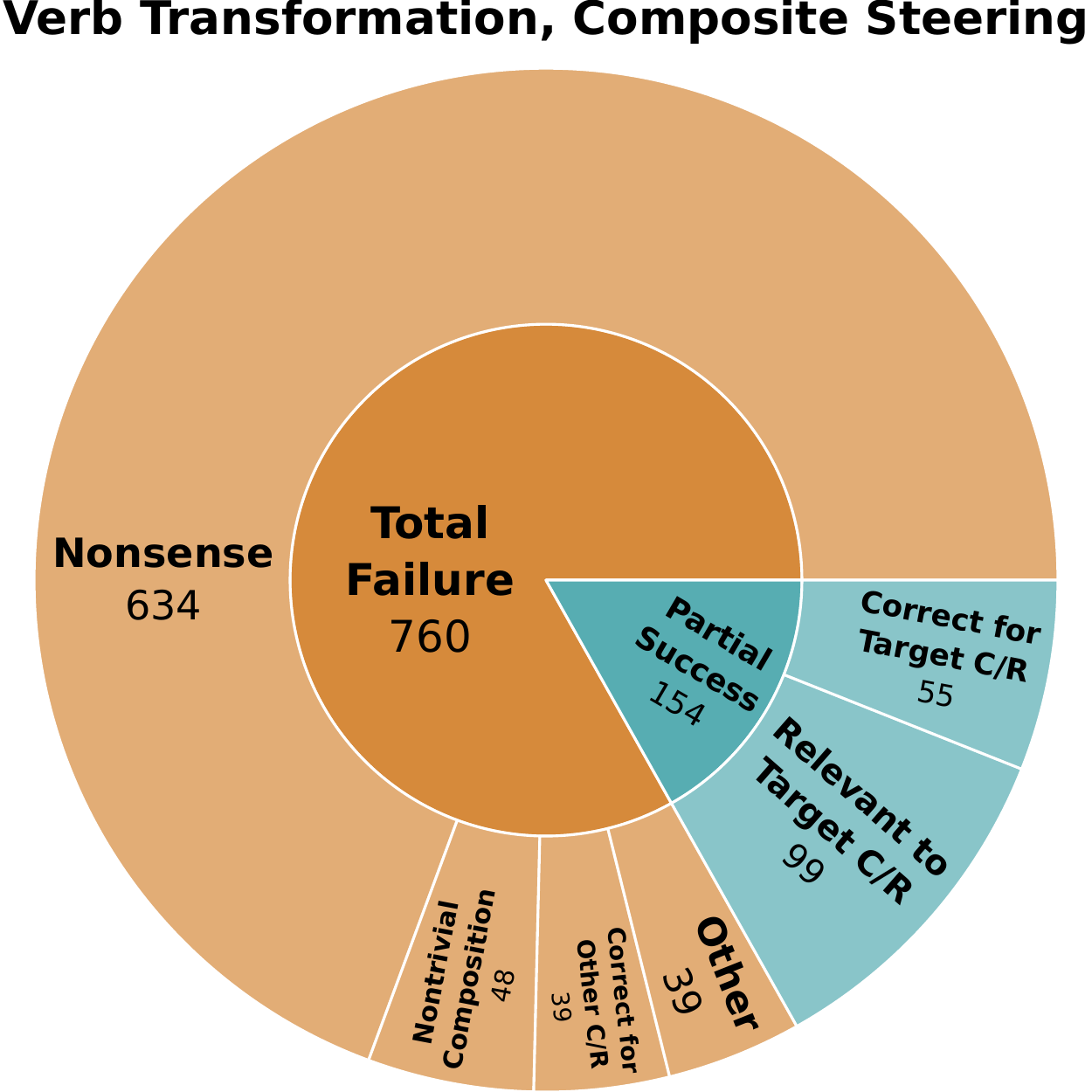}                
    \end{multicols}

    \vspace{-5pt}
    
    \caption{Coarse and fine categorization of steering failures for all tasks and steering types. For coarse categories, \textit{Partial Success} refers to steered model outputs that were apparently related or close to the expected answer for the target concept-relation pair, while \textit{Total Failure} describes outputs that were not, including both counterfactually incorrect and incoherent outputs.
    In the fine categories, \textit{C} and \textit{R} refer to concept and relation. \cmark{} indicates a steered model output that was a correct answer for the target concept or relation, $\bm{\oslash}$ for the in-prompt concept or relation, and \xmark{} for other concepts or relations; such cases are sparser in composite steering on the verb transformations task, and are thus condensed into broader categories. \textit{Other} refers to fine categories that make up less than 4\% of total failures, which are condensed for readability. While \textit{Irrelevant} refers to fluent outputs that were not clear responses to their corresponding prompts, \textit{Nonsense} refers to outputs that were not clearly fluent text, including odd tokens (e.g., ``AddTagHelper'', ``$\otimes$$\otimes$$\otimes$$\otimes$$\otimes$''), repeated tokens (e.g., ``few few few few few''), or otherwise unintelligible sequences of tokens (e.g., ``1. The'').}
    \label{fig:steering_failure_analysis}
\end{figure*}

In cases where steering fails, it is practically useful to understand the space of behaviors that may arise in LLMs. Figure~\ref{fig:steering_failure_analysis} visualizes a categorization of steering failures quantified in Section~\ref{sec:component_steering}.
% While we summarize key qualitative insights here, we visualize the categorization of failures across all tasks and steering types in Appendix~\ref{apx:steering failure analysis}.\footnote{We also further discuss interesting but rare failure cases where in-prompt and target concept/relation information composes in nontrivial ways, raising questions for future work toward understanding the distribution of knowledge in LLMs.}

\paragraph{Common failure cases.}

In the country facts task, most failure cases were still partly successful, reinforcing the accuracy of our component selection in this task. 
Specifically, most incorrect counterfactual responses were correct answers for the target fact but not country (e.g., a capital city for the wrong country, often the prompted country).
Additionally, many were at least relevant to target concepts or relations, e.g., names and non-capital cities of target countries.
In word translation, most failures were similarly relevant to the target language and/or word. Usually, incorrect outputs were one or more non-target words in the target language. However, in rare cases the generated words were also nearly correct, e.g., steering from the French translation of \textit{love} to \textit{run} yielded ``à pied'' (meaning ``on foot''). This suggests that for these two tasks, the concept and relation components we selected were generally fairly relevant to their corresponding contexts, even if they yielded some incorrect answers.

In verb transformation, however, most relation and composite steering failures invoked unintelligible or prompt-irrelevant model outputs. This again suggests that the components for this task are less reliable or specific than others, thus manipulating them can drastically impact general model performance.
Notably, when steering from other relations to capitalization, the model often generated (incorrect) sequences of capitalized tokens, e.g., "The Word, The Word." This suggests that the capitalization component primarily represents stylization of generated outputs rather than word-specific application. 
In concept steering, which was more effective, most failures were partial successes, suggesting that the components for verbs are more viable, and still compose predictably with more complex transformation operations. Many cases yielded relevant responses for target relations and concepts, such as the target verb in the wrong form (e.g., not capitalized or in past tense), other letters besides the first letter of the target verb, and capitalized sequences of irrelevant tokens.

\paragraph{Nontrivial composition.}

We additionally observed an interesting nontrivial composition of in-prompt and target concepts and relations in the word translation and verb transformation tasks.

In word translation, when steering from translating the word \textit{fish} to other words across languages, the model often output translations of words related to water (e.g., \textit{ocean}, \textit{sea}, \textit{river}, \textit{boat}). This may suggest that the word \textit{fish} is captured not only in our \textit{fish} component, but also by other activated aquatic knowledge. When the \textit{fish} component is ablated, this other knowledge may then integrate with target word knowledge in complex ways to yield responses that bridge the gap. For example, when steering from translating \textit{fish} to \textit{here} in both Spanish and German, the model generated Spanish/German words for \textit{sea}, perhaps connecting \textit{fish} to the location aspect of \textit{here}. Meanwhile, when steering from translating \textit{fish} to \textit{run} in Spanish, the model generated a Spanish word for \textit{river} (capable of ``running''). 
While only anecdotal evidence, future work may consider further exploring how lexical knowledge is distributed across LLMs and synthesized during text generation.

In verb transformation, we again see nontrivial compositions of in-prompt and target relations, specifically when steering from other relations to capitalization. Here, steered outputs often consisted of capitalized tokens referring to the prompted relation, e.g., the model generated ``The Past Tense Of The Word'' when steering from past tense, ``The Opposite of'' when steering from antonym, and ``The First Letter Of The'' when steering from first letter. As mentioned in Section~\ref{sec:component_steering}, this provides more evidence that the capitalization relation primarily represents the stylizing of generated text rather than the capability to manipulate it in a context-specific manner. Further, the intermingling of in-prompt and target information suggests that information about prompted relations is very widely distributed across many network layers and feature directions, even though the components for relations like past tense, antonym, and first letter are relatively large.

\section{Supplementary Results on Component Organization}\label{apx:component_organization}

\begin{figure*}
    \centering
    \includegraphics[width=0.9\linewidth]{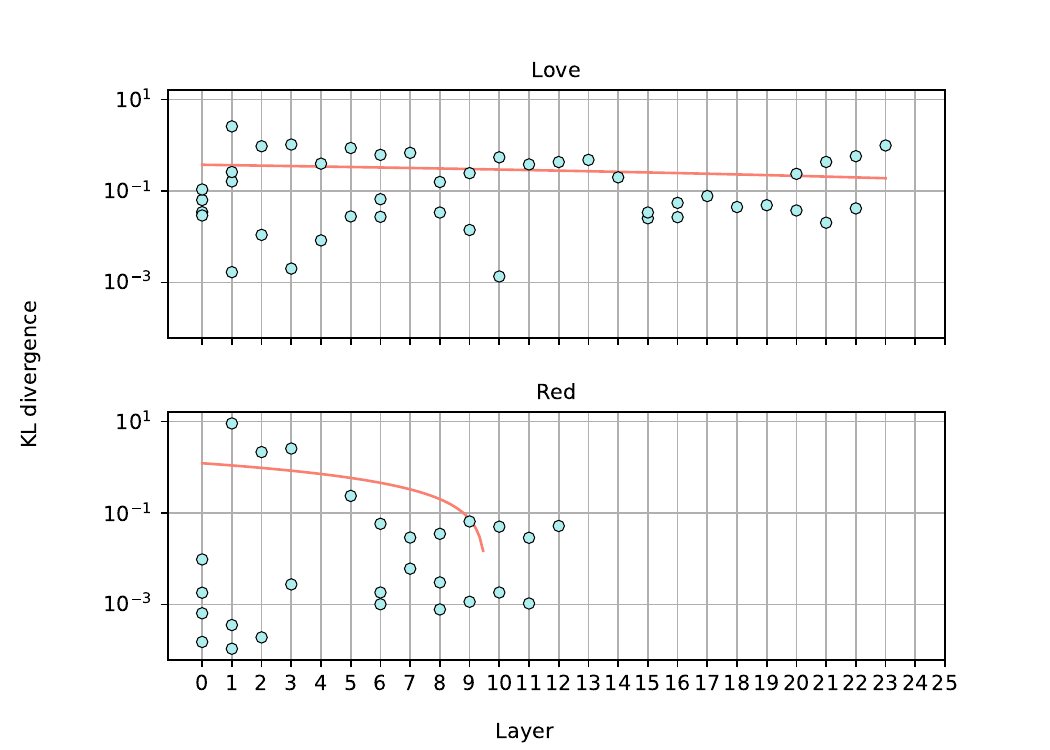}
    \vspace{-10pt}
    \caption{\revision{KL divergence between pre- and post-ablation output token distributions for each node in the \textit{love} and \textit{red} components from the word translation task, plotted by layer. Linear regression lines plotted in red.}}

    \vspace{-10pt}
    
    \label{fig:translation_concept_node_importance}
\end{figure*}

\begin{figure*}
    \centering
    \includegraphics[width=0.9\linewidth]{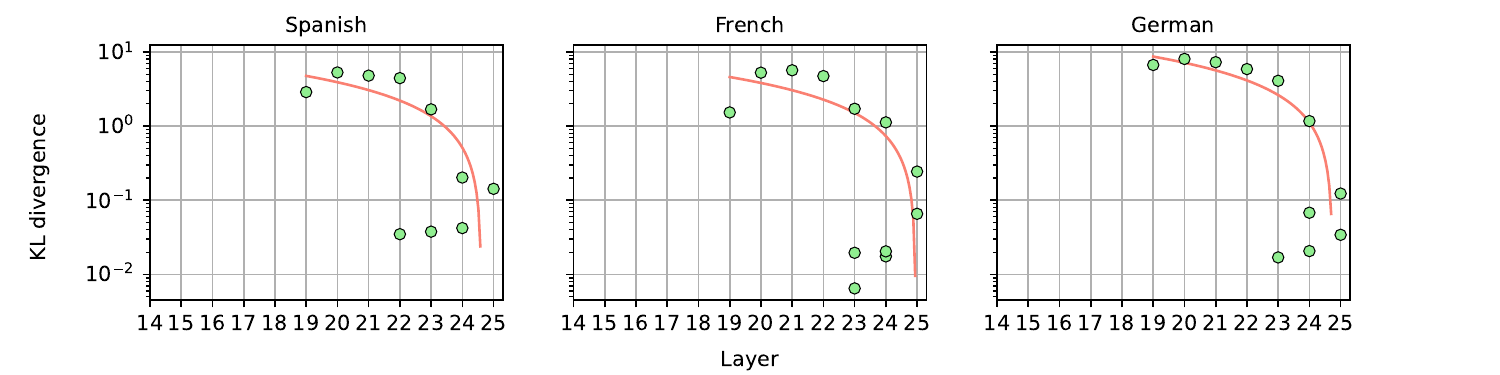}
    \vspace{-10pt}
    \caption{\revision{KL divergence between pre- and post-ablation output token distributions for each node in the translation target language components, plotted by layer. Linear regression lines plotted in red.}}

    \vspace{-10pt}
    
    \label{fig:translation_relation_node_importance}
\end{figure*}

\begin{figure*}
    \centering
    \includegraphics[width=0.9\linewidth]{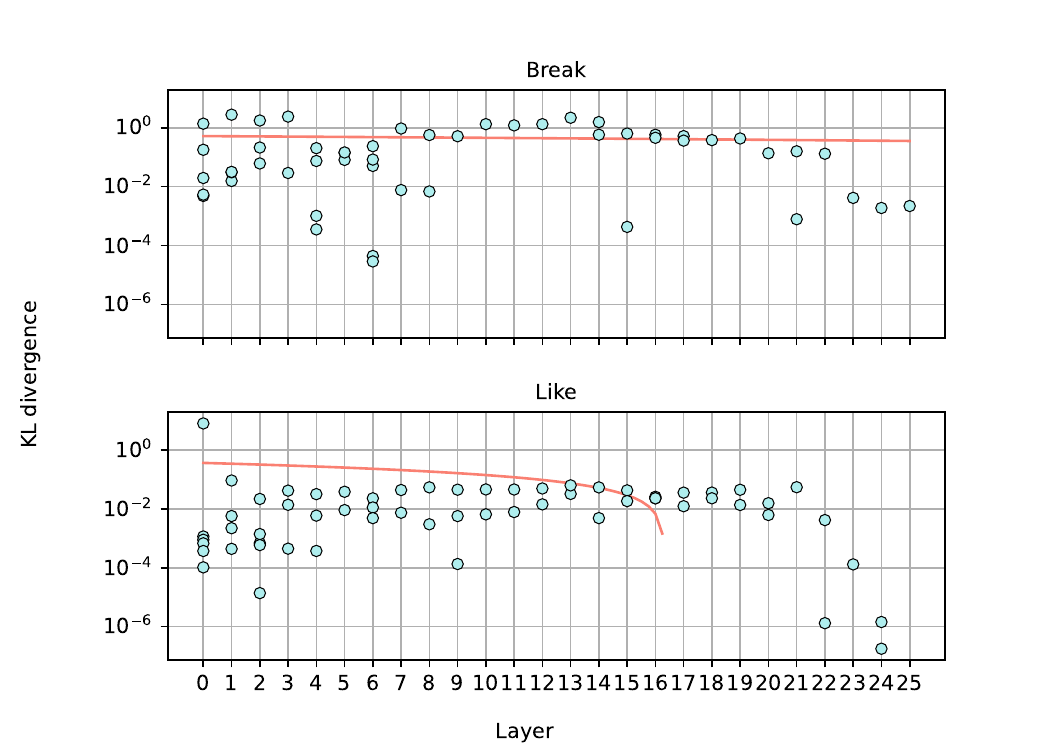}
    \vspace{-10pt}
    \caption{\revision{KL divergence between pre- and post-ablation output token distributions for each node in the \textit{break} and \textit{like} components from the verb transformation task, plotted by layer. Linear regression lines plotted in red.}}

    \vspace{-10pt}
    
    \label{fig:verbs_concept_node_importance}
\end{figure*}

\begin{figure*}
    \centering
    \includegraphics[width=0.9\linewidth]{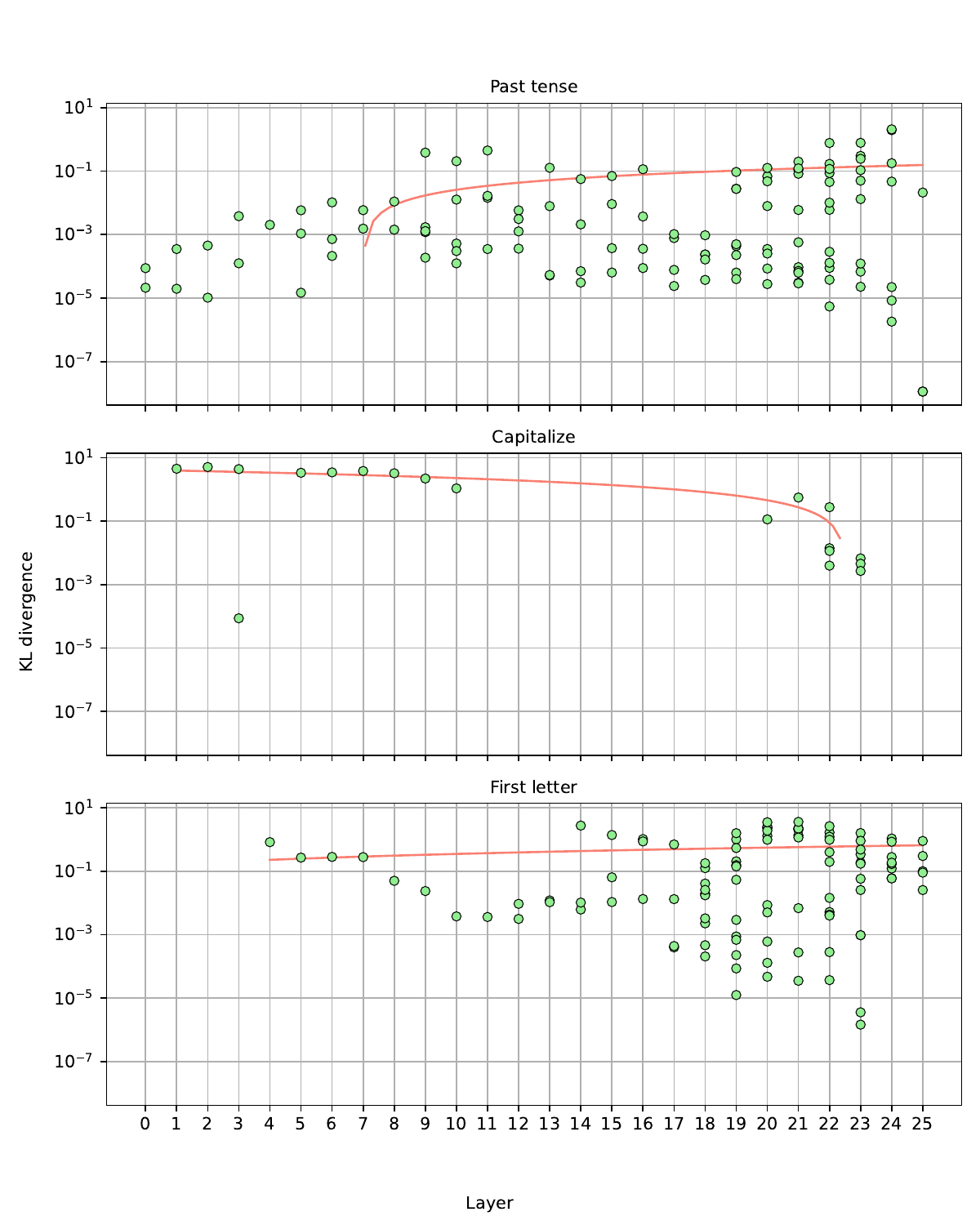}
    \vspace{-10pt}
    \caption{\revision{KL divergence between pre- and post-ablation output token distributions for each node in the verb transformation components, plotted by layer. Linear regression lines plotted in red.}}

    \vspace{-10pt}
    
    \label{fig:verbs_relation_node_importance}
\end{figure*}

\revision{Extending Figure~\ref{fig:country_node_importance} from Section~\ref{sec:component_organization}, we provide additional KL divergences by node and layer in Figure~\ref{fig:translation_concept_node_importance} (for translated words), Figure~\ref{fig:translation_relation_node_importance} (for target translation languages), Figure~\ref{fig:verbs_concept_node_importance} (for verbs to transform), and Figure~\ref{fig:verbs_relation_node_importance} (for verb transformations).}

\section{Supplementary Results on Taxonomical Reasoning}\label{apx:taxonomy}

Beyond the tasks in the paper, we additionally attempted to apply our method to a dataset for the taxonomic class, order, and family of 26 animals. However, we excluded these results from the paper after discovering that Gemma 2 2B could not perform this task perfectly off-the-shelf, thus steering based on extracted components also performed poorly.

\begin{figure}[htb!]
    \centering
    \includegraphics[width=0.7\linewidth]{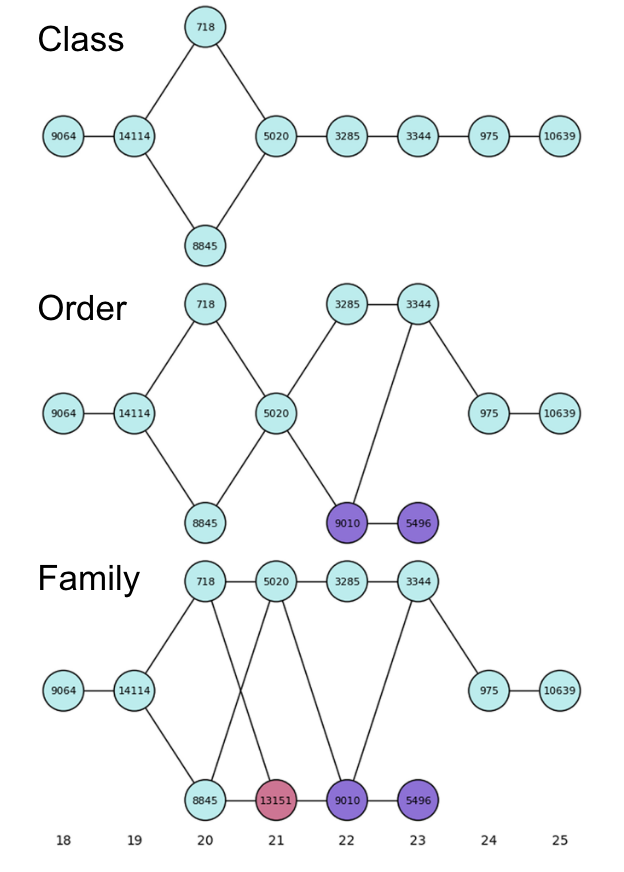}
    \caption{Species components for Angelfish as extracted from class, order, and family tasks. Nodes present in all three are shown in blue, nodes present in order and family are shown in purple, and nodes present in just family are shown in red.}
    \label{fig:subgraph_hierarchy}
\end{figure}

Interestingly, however, causally impactful components for animal species exhibited a very strong hierarchical structure across the three taxonomy tasks. Generally, the component for a given species at the class level was a subgraph of the component for the same species at the order level, and the order level component a subgraph of the family level component. (\ref{fig:subgraph_hierarchy}) This can be phrased as a general observation that as we move from more general to more specific categories, components increase in size by adding features.

We were able to reconstruct a reasonably accurate taxonomic hierarchy of the animals in our dataset by applying agglomerative clustering on the Jaccard distance matrix between family components of different species. The taxonomy is very accurate at class level, with some notable mistakes at order and family level. This coheres with previous results on LLMs' general performance on taxonomy tasks \cite{sun2024large}.

In future work, we intend to explore simpler tasks based on hierarchical knowledge.

% \textcolor{red}{TODO: cite https://arxiv.org/pdf/2406.11131 and evaluate if mistakes in hierarchy correspond to mistakes in question answering}.

\section{Transcoder Analysis}\label{apx:transcoders}

Transcoders \cite{dunefsky2024transcoders} are a major development in interpretability research. They operate by reconstructing the multilayer perceptron layer outputs from the inputs to the MLP layer. 

To investigate whether our approach could be used to identify components for transcoders, we extend our translation task to the Gemma Scope 2 2B transcoders \cite{lieberum2024gemmascopeopensparse} and focus on language components for French, German, and Spanish. The model's outputs were quite robust to ablating top-activated features, so we were only able to identify causal components for \textit{dog}, \textit{run}, \textit{beautiful}, \textit{learn}, \textit{love}, \textit{red}, and \textit{good}. In initial experiments, we found that for some prompts, we could ablate essentially all of the top-activated features without changing the model's output, so we attribute this poor performance at least in part to relevant information being carried in the error term.

We found that a correlation threshold of 0.7 was more suitable for identifying causal components in transcoders, by testing decreasing values in increments of 0.5 and inspecting the results. We then ran our algorithm as described, with the exception of skipping the density pruning step. The lower threshold caused us to identify very large components with quite a bit of overlap. We assigned only the nodes unique to each language to the final components used for ablation and steering. 

We conducted a relation steering experiment, ablating the component identified for the original language and steering the component identified for another. Testing integer values up to 20, we found that 13 produced the best accuracy. Our final accuracy was 27\%.

This performance is not as good as our method achieves with SAEs, and comes with quite a few limitations. However, through manual inspection we found that the same causal features that \citeauthor{circuit-tracer} identify in their language intervention demo notebook are present in our identified components. While our method may need refining and adaptation to work at scale with transcoders, based on these results we feel confident that feature coactivation can help identify relevant conceptual components in transcoders. 

\section{License Information}\label{apx:licenses}

We provide available licenses and terms of use for key artifacts employed in this work, including relevant links:
\begin{itemize}
    \item \textbf{Hugging Face Transformers} 
    \begin{itemize}
        \item License: Apache 2.0 (\url{https://github.com/huggingface/transformers/blob/main/LICENSE})
        \item Terms of Service: \url{https://huggingface.co/terms-of-service}
    \end{itemize}
    \item \textbf{Gemma 2 2B}
    \begin{itemize}
        \item License: Apache 2.0 (\url{https://github.com/google-deepmind/gemma/blob/main/LICENSE})
        \item Terms of Use: \url{https://ai.google.dev/gemma/terms}
    \end{itemize}
    \item \textbf{Gemma 2 9B}
    \begin{itemize}
        \item License: Apache 2.0 (\url{https://github.com/google-deepmind/gemma/blob/main/LICENSE})
        \item Terms of Use: \url{https://ai.google.dev/gemma/terms}
    \end{itemize}
    \item \textbf{Gemma Scope}
    \begin{itemize}
        \item License: Apache 2.0 (\url{https://huggingface.co/google/gemma-scope-2b-pt-res/blob/main/LICENSE})
        \item Terms of Use: \url{https://ai.google.dev/gemma/terms}
    \end{itemize}    
    \item \textbf{Transformer Lens} 
    \begin{itemize}
        \item License: MIT (\url{https://github.com/TransformerLensOrg/TransformerLens/blob/main/LICENSE})
    \end{itemize}
    \item \textbf{SAE Lens} 
    \begin{itemize}
        \item License: MIT (\url{https://github.com/jbloomAus/SAELens/blob/main/LICENSE})
    \end{itemize}
    \item \textbf{NetworkX}
    \begin{itemize}
        \item License: 3-clause BSD (\url{https://github.com/networkx/networkx/blob/main/LICENSE.txt})
    \end{itemize}
    \item \textbf{Neuronpedia API} 
    \begin{itemize}
        \item License: MIT (\url{https://github.com/hijohnnylin/neuronpedia/blob/main/LICENSE})
    \end{itemize}

\end{itemize}

We have verified that this work acts in accordance with all available licenses and terms of use.

\end{document}